\documentclass[10pt,journal,compsoc]{IEEEtran}
\usepackage{rotating}
\usepackage{tikz}
\usepackage{bbding}
\usepackage{pifont}
\usepackage{wasysym}
\usepackage{amssymb}
\usepackage{subcaption}
\usepackage{caption}
\usepackage{graphicx}
\usepackage{verbatim}

\usepackage{array} 

\newcolumntype{L}[1]{>{\raggedright\let\newline\\\arraybackslash\hspace{0pt}}m{#1}} 
\newcolumntype{C}[1]{>{\centering\let\newline\\\arraybackslash\hspace{0pt}}m{#1}} 
\newcolumntype{R}[1]{>{\raggedleft\let\newline\\\arraybackslash\hspace{0pt}}m{#1}}

\usepackage{xcolor}

\ifCLASSOPTIONcompsoc
  \usepackage[nocompress]{cite}
\else
  \usepackage{cite}
\fi
\ifCLASSINFOpdf
\fi
\usepackage{pdfpages}

\usepackage[switch]{lineno}

\begin{document}



\title{Machine Learning-Aided Operations and Communications of Unmanned Aerial Vehicles: A Contemporary Survey}


%
%
%
%

\author{Harrison Kurunathan, Hailong Huang, Kai Li, Wei Ni, and Ekram Hossain, \emph{Fellow}, \emph{IEEE}
\IEEEcompsocitemizethanks{\IEEEcompsocthanksitem Harrison Kurunathan and Kai Li are with CISTER - Research Centre in Real-Time and Embedded Computing Systems, Porto, Portugal (emails: \{hhkur,Kai\}@isep.ipp.pt).
\IEEEcompsocthanksitem H. Huang is with the Department of Aeronautical and Aviation Engineering, The Hong Kong Polytechnic University, Hong Kong (email: hailong.huang@polyu.edu.hk).
\IEEEcompsocthanksitem W. Ni is with the Commonwealth Scientific and Industrial Research Organization (CSIRO), Sydney, Australia (email: wei.ni@csiro.au).
\IEEEcompsocthanksitem E. Hossain is with the University of Manitoba, Canada (email: Ekram.Hossain@umanitoba.ca).
}
}

%
%

\markboth{Submitted to the xxx, May 2022}%
{Shell \MakeLowercase{\textit{et al.}}: Bare Demo of IEEEtran.cls for Computer Society Journals}
%



\IEEEtitleabstractindextext{%
\begin{abstract}

Over the past decade, Unmanned Aerial Vehicles (UAVs) have provided pervasive, efficient, and cost-effective solutions for data collection and communications. Their excellent mobility, flexibility, and fast deployment enable UAVs to be extensively utilized in agriculture, medical, rescue missions, smart cities, and intelligent transportation systems. 
Machine learning (ML) has been increasingly demonstrating its capability of improving the automation and operation precision of UAVs and many UAV-assisted applications, such as communications, sensing, and data collection. 
The ongoing amalgamation of UAV and ML techniques is creating a significant synergy and empowering UAVs with unprecedented intelligence and autonomy.  
This survey aims to provide a timely and comprehensive overview of ML techniques used in UAV operations and communications and identify the potential growth areas and research gaps.  
We emphasize the four key components of UAV operations and communications to which ML can significantly contribute, namely, perception and feature extraction, feature interpretation and regeneration, trajectory and mission planning, and aerodynamic control and operation. 
We classify the latest popular ML tools based on their applications to the four components and conduct gap analyses. 
This survey also takes a step forward by pointing out significant challenges in the upcoming realm of ML-aided automated UAV operations and communications. 
It is revealed that different ML techniques dominate the applications to the four key modules of UAV operations and communications. While there is an increasing trend of cross-module designs, little effort has been devoted to an end-to-end ML framework, from perception and feature extraction to aerodynamic control and operation. 
It is also unveiled that the reliability and trust of ML in UAV operations and applications require significant attention before full automation of UAVs and potential cooperation between UAVs and humans come to fruition.

\end{abstract}

\begin{IEEEkeywords}
Unmanned Aerial Vehicle (UAV), UAV-aided communications, UAV operations, Artificial Intelligence (AI), Machine Learning (ML)
\end{IEEEkeywords}}

\maketitle

\IEEEdisplaynontitleabstractindextext

%
\IEEEpeerreviewmaketitle

\section*{Acronyms}
AI    \hspace{0.3 in} Artificial Intelligence                 \\
ARD   \hspace{0.2 in} Automatic Relevant Detection            \\
AoI   \hspace{0.25 in} Age of Information                      \\
CNN   \hspace{0.2 in} Convolutional Neural Network            \\
CPS   \hspace{0.2 in} Cyber Physical Systems                  \\
CPU   \hspace{0.2 in} Central Processing Unit                 \\
DDPG  \hspace{0.2 in} Deep Deterministic Policy Gradient      \\
DNN   \hspace{0.2 in} Deep Neural Network                     \\
DRL   \hspace{0.2 in} Deep Reinforcement Learning             \\
DL    \hspace{0.3 in} Deep Learning                           \\
DQN   \hspace{0.2 in} Deep Q Network                          \\
ESN   \hspace{0.2 in} Echo State Network                      \\
GA    \hspace{0.3 in} Genetic Algorithm                       \\
GAN   \hspace{0.2 in} Generative Adversarial Network          \\
GMM   \hspace{0.2 in} Gaussian Mixture Modeling               \\
GPU   \hspace{0.2 in} Graphics Processing Unit                \\
GWR   \hspace{0.2 in} Geographically Weighted Regression      \\
IoT   \hspace{0.2 in} Internet of Things                      \\
ITS   \hspace{0.2 in} Intelligent Transport System            \\
LM    \hspace{0.3 in} Levenberg Marquardt                     \\
LoS   \hspace{0.2 in} Line-of-Sight                           \\
LR    \hspace{0.3 in} Linear Regression                       \\
LSTM  \hspace{0.2 in} Long Short Term Memory                  \\
LTE   \hspace{0.2 in} Long Term Evolution                     \\
MEC   \hspace{0.2 in} Mobile Edge Computing                   \\
ML    \hspace{0.2 in} Machine Learning                        \\
MLP   \hspace{0.2 in} Multi Layer Perceptron                  \\
MPC   \hspace{0.2 in} Model Predictive Control                \\
NIR   \hspace{0.2 in} Near Infra Red                          \\
OSL   \hspace{0.2 in} Optimal Strategy Library                \\
PID   \hspace{0.2 in} Proportional Integrative Derivative     \\
QoS   \hspace{0.2 in} Quality of Service                      \\
RNN   \hspace{0.2 in} Recurrent Neural Network                \\
RL    \hspace{0.2 in} Reinforcement Learning                  \\
RSS   \hspace{0.2 in} Received Signal Strength                \\
R-CNN \hspace{0.2 in} Recurrent Convolutional Neural Network  \\
SfM   \hspace{0.2 in} Structure from Motion                   \\
SINR  \hspace{0.2 in} Signal to Noise plus Interference ratio \\
SNN   \hspace{0.2 in} Spiking Neural Network                  \\
SNR   \hspace{0.2 in} Signal-to-Noise Ratio                   \\
STDP  \hspace{0.2 in} Spike Time Dependant Plasticity         \\
SVM   \hspace{0.2 in} State Vector Machine                    \\
UAV   \hspace{0.2 in} Unmanned Aerial Vehicle                 \\
VTOL  \hspace{0.2 in} Vertical Take Off and Landing           \\
WPT   \hspace{0.2 in} Wireless Power transfer                 \\


\section{Introduction}\label{sec:introduction}
\begin{figure*}[h!]
\centering
\includegraphics [width= 7.2 in]{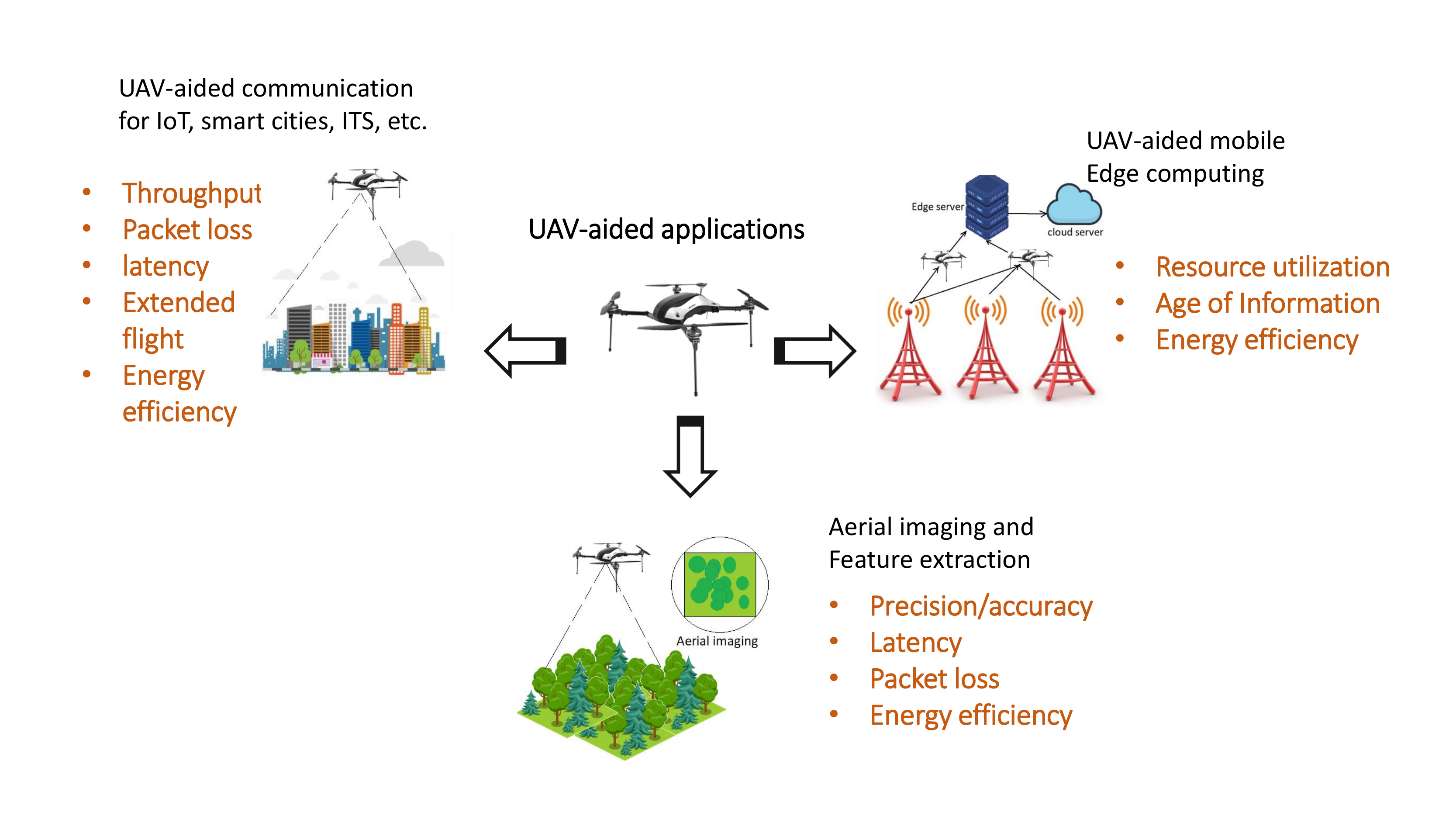}
\caption{The paradigms of UAV-assisted IoT that demand variable quality-of-services ranging from guaranteed reliability, minimal latency, the freshness of information, and energy efficiency.}
\label{f1_overview}
\end{figure*}

With their excellent mobility, versatility, and ability to cover wide and harsh environments, unmanned aerial vehicles (UAVs) have been increasingly proliferating with extensive applications, as shown in Fig.~\ref{f1_overview}. The Global UAV Market has been projected to grow at a cumulative rate of 19.9\%, and generate a revenue of \$55.649 billion from 2020 to 2027 \cite{market2015global}. In the past, UAVs were primarily used for surveillance and reconnaissance in military applications~\cite{XinYuanTCOM2019,XinYuanTIFS2020}. With the new trends in aerial photography and monitoring over the past decade~\cite{ShuyanTIFS2021}, UAVs have started to enable many civil and commercial application domains. 
For example, UAVs have been increasingly implemented in several monitoring domains, such as marine \cite{ma2020uav,islam2020bumar}, traffic \cite{HailongTASE2021,alioua2020uavs}, goods delivery~\cite{BinLiuTITS2021}, public safety~\cite{HailongTVT2021,HailongINDIN2020}, and agriculture \cite{uddin2018uav}. UAVs are also extensively considered to extend the connectivity and coverage of terrestrial communications systems, for example, mobile cellular systems. They can serve as aerial cellular base stations~\cite{mozaffari2017wireless} or mobile repeaters and transponders~\cite{zeng2016throughput,LukasACMCPS2021} with radio transceivers to offer connectivity and data services to users on the ground \cite{li2018uav}, or deliver confidential messages~\cite{SavkinWCL2020}.

In the emerging application of mobile edge computing~(MEC), UAVs are considered to play the role of mobile computing servers, providing a cost-effective alternative to expensive physical computing infrastructure \cite{narang2017uav}. This is attributed to the fact that commercially available UAVs are increasingly computationally capable and equipped with compact central processing unit (CPU) and graphics processing unit (GPU) modules~\cite{Chao_Sun_MEC_2021}. For instance, \textcolor{black}{UAVs can act as micro base stations (BSs) and provide edge computing resources, by dynamically moving over remote locations, where data coverage is required and computing resources need to be provided on-demand~\cite{jeong2017mobile, lu2019toward}.} UAVs act as mobile computing servers that allow computationally restrictive ground devices, e.g., Internet-of-Things (IoT) devices, to offload their computationally intensive applications. The optimization of the UAV trajectory and radio transmission power of the ground devices contributes to the maximization of performance of the offloaded computing applications~\cite{qian2019user,Chao_Sun_MEC_2021}. By dispatching and placing the UAVs at carefully selected locations, it is possible to increase the throughput, coverage, and spectral efficiency.

\subsection{Objectives of UAV Operations and Communications}

UAVs have also been increasingly employed for data collection or forwarding in remote, human-unfriendly environments, where conventional terrestrial communication infrastructures relying on persistent power supplies are unavailable or unreliable~\cite{KaiLiTMC2016}. In particular, one or multiple UAVs hover over a geographical entity to capture and process images or data from the ground sensors~\cite{alzahrani2020uav,XinYuanTVT2020}. 
Each UAV can adjust its flight trajectory and altitude to physically approach a ground node, enjoying an excellent line-of-sight (LoS) radio connection to improve the data rate and save the energy of the ground node~\cite{SunChaoWCSP2020}.

The typical objectives of UAV-assisted data collection and communications include, but are not limited to,
\begin{itemize}
    
    \vspace{2 mm}
    \item \textbf{Richness of collected data:} The richness of collected data can be interpreted as system throughput; or in other words, the capability of the UAVs to collect data. By exploiting the richness of data, data-driven approaches have been increasingly  adopted to optimize the design, operation, and maintenance of cyber-physical systems (CPS), such as smart city~\cite{cui2021big} and intelligent transport systems~\cite{QimeiIOTJ2019,YingzhuSpringer2021}. The value of data is increasingly recognized~\cite{jordan2015machine}. It is important to minimize the data loss during the collection and processing of data, thereby retaining useful information and avoiding rare features from being overlooked~\cite{samir2019uav}.
    
    \vspace{2 mm}
    \item \textbf{Freshness and timeliness of collected data:} While freshness apparently adds value to data~\cite{liu2020uav}, timely collection of data is crucial for systems relying on UAVs to deliver data, such as systems deployed in remote, human-unfriendly environments. This is due to the finite batteries and buffers of devices in those systems~\cite{cao2019energy}. Many IoT devices deployed in remote, human-unfriendly environments are (re)charged by renewable energies scavenged from ambient sources, such as polar power and wind, which are unreliable and can experience unexpected shortages~\cite{ShuyanXiaojingTGCN2020,XiaojingTWC2016}. Delayed data collection would not only cost the freshness of the data but lead to buffer overflows and subsequently data losses~\cite{mukherjee2021igridedgedrone}.
    
    \vspace{2 mm}
    \item \textbf{Representativeness of collected data:} Apart from the richness and freshness, the representativeness of data is critical to many data analytics and modeling activities~\cite{XinchenJSAC2019}. The representativeness refers to the resemblance of collected data to the entire dataset~\cite{XinchenJSAC2019}. It is important to avoid the well-known overfitting problem during data analytics and modeling~\cite{guo2020partially}.
    
    \vspace{2 mm}
    \item \textbf{Reliability and dependability:} Reliability is often measured by the outage probability of transmissions, and heavily depends on the propagation channels~\cite{GordonCOMST2019}. Reliability can have a strong impact on the throughput and freshness of delay-sensitive data, while data is sensitive to delays in many UAV-assisted communications and data collection scenarios. This is because the UAVs can quickly fly out of a node's communication coverage and the node must withhold its transmission until the next opportunity when the UAVs approach or pass by. 
\end{itemize}

\subsection{ML for UAV Operations}

\begin{figure*}[h!]
\centering
\includegraphics [width= 6.6 in]{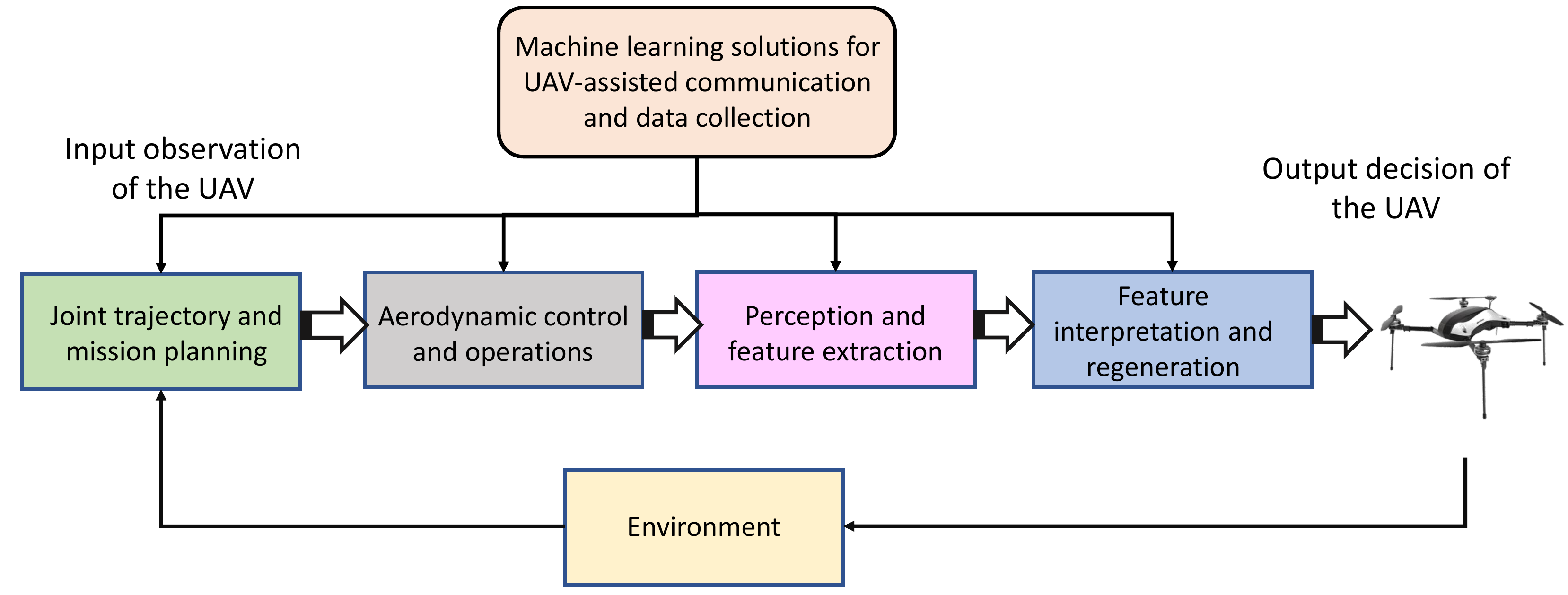}
\caption{Four key components of UAV control and operations,  namely, UAV perception and feature extraction, feature interpretation and regeneration, trajectory and mission planning, and aerodynamic control and operation,  to which ML techniques can considerably contribute.}
\label{f_MLinUAV}
\end{figure*}

Machine Learning (ML) enables systems to learn from data, creates data-driven solutions, and has been increasingly applied online in distributed settings~\cite{ShuyanCOMST2021}. By associating UAVs with ML, it is possible to add functionalities like processing images for classification and segmentation, deciding the trajectories of the drone, caching, scheduling, and monitoring \cite{pajares2015overview}. With ML helping around controlled mobility, trajectories, and adjustable altitudes, UAVs have become a suitable candidate for enabling various IoT paradigms that need an additional layer of artificial intelligence (AI). Capabilities such as feature extraction and prediction add a layer of AI to the existing UAV-enabled monitoring applications, as illustrated in Fig.~\ref{f_MLinUAV}. 

The integration of ML into UAV platforms provides many opening opportunities and new methods in various domains, such as real-time monitoring, data collection and processing, and prediction in the computer/wireless networks, smart cities, military, agriculture, and mining. It is reported in~\cite{khan2019unmanned}  that the UAV and ML research for smart cities and military use was accelerated by 40\% in 2019. It is also revealed that the UAV and ML research in other sectors, such as agriculture, maritime monitoring, and infrastructure monitoring, displays constant growth. The incorporation of ML in these UAV-assisted applications can help improve the quality of many applications and services. 
ML has been used in the resource management and trajectory optimization of UAVs~\cite{alsenwi2020uav,lu2020power,li2020energy,hu2018uav}, e.g., to save the overall energy usage and extend the cruise time. ML-based stochastic computational offloading practices~\cite{zhang2018stochastic} have been used in the literature to improve the resource allocation of mobile edge computing. ML has also been employed for joint optimization of UAV's flight path, radio emission power, and cached contents, striking a balance between energy efficiency and latency~\cite{xu2018overcoming,chen2017caching,chai2020online}. ML frameworks, such as those developed in \cite{abd2019deep} and \cite{wu2021uav}, have been used to optimize the UAV's flight route and coordinate the ground sensors' data transmissions to maintain the freshness of data; or in other words, minimizing the Age of Information (AoI).
In Table \ref{table1}, we present prominent performance indicators of UAV-aided applications and their potential ML solutions. In what follows, we provide a brief discussion on the ML solutions to meet these performance requirements. To this end, it is of prominent importance and urgency to develop a taxonomic analysis of ML applications to enhance UAV operations and communications.

\begin{table}[]
\centering
\caption{Performance metrics and ML solutions}
\begin{tabular}{ll}
\hline \\
\multicolumn{1}{c}{Performance indicators} & ML solutions                                                            \\
\hline\\
Age of Information              & \begin{tabular}[c]{@{}l@{}}Trajectory planning, \\ node selection\end{tabular}     \\
Communication reliability       & \begin{tabular}[c]{@{}l@{}}Backhaul, fronthaul\\ caching\end{tabular}               \\
Energy efficiency               & QoS-based cruise control                                                              \\
Resource utilization            & Computational offloading                                                               \\
Image resolution                & Feature extraction                                                                      \\
Accuracy                        & Prediction and classification                                                      \\
Packet loss                     & QoS-based data collection                                                         \\
Security                        & Adversarial learning                                          \\
\hline                         
\end{tabular}
\label{table1}
\end{table}

\subsection{Key Components of UAV Operations and Communications}

To achieve the objectives of UAV operations and  the QoS requirement for UAV-assisted communications, it is imperative to holistically design and optimize the trajectories and the data collection schedules of UAVs. This would require the UAVs to understand the operating environment and data demand, comply with the aerodynamics and energy availability of the UAVs, and allow the UAVs to dynamically adjust their speed, heading, elevation, and acceleration adapting to the changing environment and demand.  Consequently, the UAV operations and communications involve the following four major components:
\begin{itemize}
   
    \item Joint trajectory and mission planning, including multi-UAV cooperation;
    \item Aerodynamic control and operation, i.e., refine the online operations of individual UAVs;
     \item Perception and feature extraction, i.e., understand the environment;
    \item Feature interpretation and regeneration, i.e., digitize, interpret and model the environment.
\end{itemize}
This procedure is shown in Fig.~\ref{f_MLinUAV}.

\begin{figure*}[h!]
\centering
\includegraphics [width=7in]{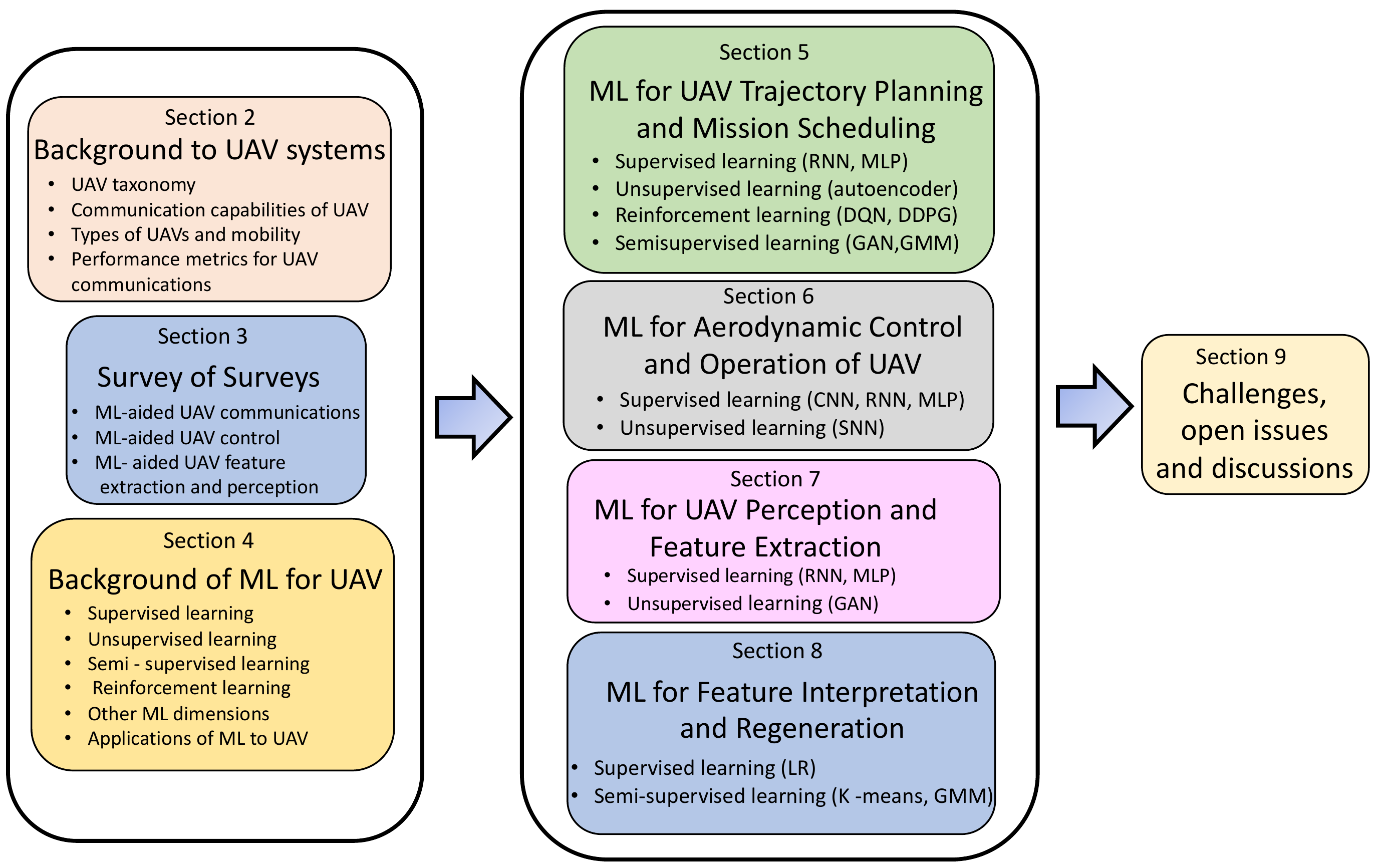}
\caption {The structure of this survey.}
\label{fig: organisation}
\end{figure*}

\subsection{Contributions of This Survey}
 
This survey presents a comprehensive overview of ML techniques specially designed for UAV applications, as well as their advantages and drawbacks under different UAV-aided operations and communications. Specifically, the survey categorizes the UAV-compatible ML techniques holistically based on the performance metrics that they are designed against, and the applications that they embrace in support of feature extraction, environment modeling, UAV control, and data collection of UAV operations, as depicted in Fig.~\ref{f_MLinUAV}. The gaps and opportunities of existing studies are identified. 

The key contributions of the survey are as follows:
\begin{itemize}
    \item We provide an in-depth review of all existing surveys and technical research related to ML-assisted UAV applications and operations;  
    \item We provide a comprehensive application analysis of critical ML techniques and discuss their advantages and drawbacks in regards to their applications to  feature extraction, environment modeling, UAV control, and data collection of UAV operations;
    \item We provide a detailed outlook of significant challenges and open issues in applications of ML to UAV-assisted communications and operations.
\end{itemize}
Important findings include, but not limited to the following: 
\begin{itemize}
    \item Novel ML techniques with augmented functionalities and a combination of different ML techniques have been introduced over the last decade to meet the performance demands of UAV-aided applications.
    
    \item While convolutional neural network (CNN) is predominately applied to UAV image processing, low-power ML techniques, such as spiking neural network (SNN), have started to demonstrate their applicability, especially in online operations. 
    
    \item Deep reinforcement learning (DRL) techniques with continuous action spaces, such as policy-based deep deterministic policy gradient (DDPG), are increasingly demonstrating their potential for online UAV flight control and communication scheduling. They can concatenate with other DL modules, such as recursive neural network (RNN), to enhance feature extraction and accelerate exploration and exploitation. 
    
    \item While there is a significant effort in creating ML modules to support feature extraction, environment modeling, planning and scheduling, and control and operation, little to no effort has been devoted to creating an ML-based end-to-end control solution from feature extraction to planning, control, and operation. 
    
    \item Little consideration has been given to quantifying the reliability and trustworthiness of ML modules in UAV operations and applications. 
\end{itemize}

Every ML technique is unique and supports one or more performance metrics demanded by different UAV-aided applications. Generally, supporting one specific performance or function can result in undesired trade-offs on other aspects. Layers of AI on these ML-enabled applications can help alleviate the trade-offs. It is important to provide a balanced view of two major features, namely, the state-of-the-art ML algorithms, and the application domains supported by the ML algorithms, as is done in this survey. {\em This is the key differentiator of this survey from the existing reviews which are typically featured with particular UAV applications of ML techniques}, as discussed in Section~\ref{sec. gap of survey}.

It is worth mentioning that this survey reviews over 300 recent research papers on the latest specially-designed, UAV-compatible, ML techniques in support of UAV operations and communications, including but not limited to, SNN, R-CNN, double looped RNN, multi-agent DRL, and double DQN. 

\textcolor{black}{This survey provides a combined overview of the supervised, unsupervised, and reinforcement learning architectures. We endeavor to encapsulate a broader umbrella of applications pertaining specifically to the UAV-aided operations in terms of trajectory,  mission planning, aerodynamic control (mobility control), feature extraction, and perception. We also provide a detailed discussion on some of the prominent challenges and open issues in this field.}

\subsection{Organization of This Survey}
The rest of this survey is organized as follows. 
In Section \ref{sec: performance vs ML}, we discuss the ML tools for UAV operations and communications. From Section \ref{sec: feature extraction} till Section \ref{sec: control and operation}, we segregate the ML techniques in their respective application domains. Specifically, we cover trajectory and mission planning in Section \ref{sec: planning and scheduling},  aerodynamic control and operation in Section \ref{sec: control and operation}, UAV perception and feature extraction in Section \ref{sec: feature extraction}, and feature interpretation and regeneration in Section \ref{sec: feature interpretation}. Section \ref{sec: open issues} outlines the open scopes and remaining challenges of ML's applications to UAV operations and applications, followed by concluding remarks in Section \ref{sec: conclusion}. In Fig. \ref{fig: organisation}, we provide the detailed organization of this survey.

\section{Background to UAV Systems}

\subsection{UAV Taxonomy}
\textcolor{black}{UAV systems can be categorized into two types, the single UAV systems, and the multi UAV systems \cite{vergouw2016drone, gupta2013review}. In a single-UAV system, the entire mission will rely on a single UAV. In a multi-UAV scenario, UAVs in a swarm can facilitate the mission. More UAVs can cover a larger geographical area within a shorter time than their single UAV counterparts. Multi-UAV systems also have the capability to process tasks in parallel, speeding up the mission completion time. Single-UAV systems will have to maintain constant communication with the ground infrastructures or the operator. In a multi-UAV system, one specific coordinating UAV can communicate with the ground and forward the messages to other UAVs \cite{galkin2019uavs}.
}

\begin{figure*}[h!]
\centering
\includegraphics [width= 6 in]{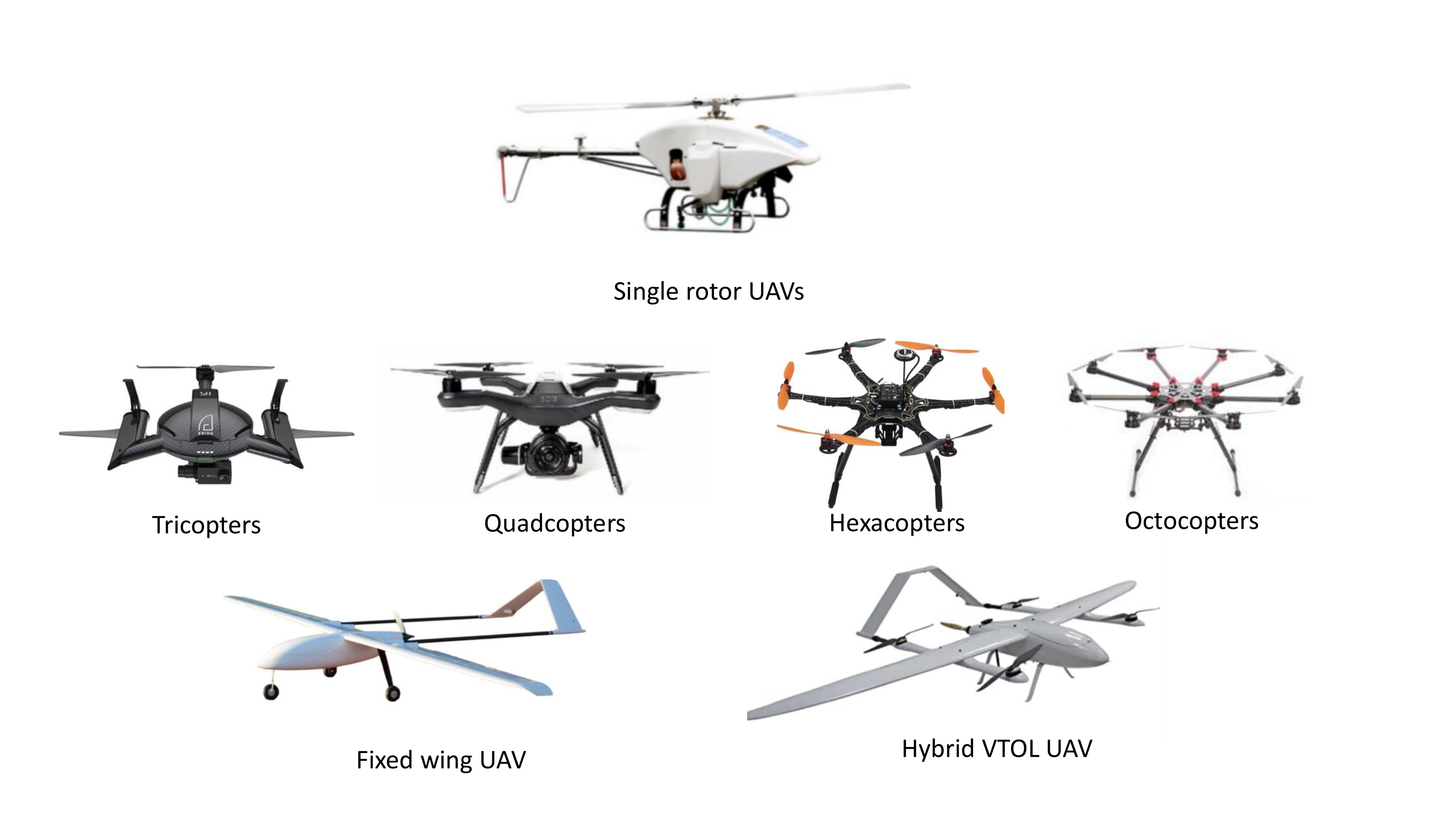}
\caption {Different types of UAV based on number of rotors and take-off methods.}
\label{mlpstruct}
\end{figure*}

\subsection{Communication Capabilities of UAV}
\textcolor{black}{In accordance with the guidelines from the International Telecommunication Union (ITU) \cite{marcus2014spectrum, zeng2020uav}, a UAV must be able to communicate in three different ways. Firstly, communication for UAV command and control must be possible. This includes the telemetry report (e.g., the flight status from the UAV to the ground pilot), signaling from the ground to UAVs (e.g., attitude control data), and updates from the ground for autonomous UAVs. Attitude control is the process of controlling the orientation of the UAV in accordance with an inertial frame of reference or other entities along the UAV trajectory. Secondly, communication for air traffic control (ATC) relay ensures the safety of traditional manned aircraft from any UAVs in the same fly zone. This communication is critical in the presence of higher densities of air traffic. Finally, communication aiding collision avoidance supports sensing and avoiding any obstacle with a sufficient safety distance in the cruise line.
}

\subsection{Types of UAVs and Mobility} 
\textcolor{black}{UAVs can be classified by the flight system adhered or by the rotors in the flight system. Some of the common types of UAVs include Vertical TakeOff and Landing (VTOL) UAVs, fixed-wing UAVs, and hybrid VTOL UAVs. 
}

\textcolor{black}{VTOL UAVs can be further classified based on the number of rotors: single-rotor UAVs \cite{carholt2016design} and multi-rotor UAVs \cite{finger2017review}. Single-rotor UAVs have a similar structure to helicopters in design. This type of UAV comprises  of one single rotor that acts as the wing helping in elevation and a tail rotor for controlling the direction and stability. It can be very energy-efficient as it has to support only one rotor, and this can improve the cruise time of the UAV. However, they are less stable than their multi-rotor counterparts. Multirotor UAVs have more than one motor. They can be tricopters (three rotors), quadcopters (four rotors), hexacopters (six rotors), and octocopters (eight rotors), among others. They provide great control over the position and framing on air, thus making them suitable for many applications such as photography. These UAVs support multiple degrees of axes and rotate on their own axis. 
One disadvantage of this type of UAV lies in the limitation in flight time  \cite{suprapto2017design}. 
}

\textcolor{black}{Fixed-wing UAVs \cite{cory2008experiments} have rigid wings extending toward two sides of the body of the UAVs and resemble an airplane. Unlike  rotor-based UAVs, they utilize the energy to move forward. They are utilized to cruise over long distances covering a larger geographic area for hours. Landing is tougher for these UAVs than it is for their rotor-based counterparts.} 

\textcolor{black}{Hybrid VTOL UAVs \cite{ozdemir2014design} combine the characteristics of fixed-wing and rotor-based designs. These VTOL UAVs have rotors attached to the fixed wings. The rotor blades of the UAVs create a vertical thrust like a large propeller. This thrust enables the UAV to take off and land vertically, and to hover. The wings allow the the UAV to glide and enable pitch (i.e., rotation around the front-to-back axis), roll (i.e., rotation around the side-to-side axis ), and yaw (rotation around the vertical axis) motions. These UAVs are more maneuverable than fixed-wing UAVs, and can fly longer than typical multi-rotor UAVs.
}

\subsection{Performance Metrics for UAV Communications}

The following performance metrics are commonly used in UAV-assisted communications systems. These performance metrics play a major role in defining the actions taken by the ML strategies.

\textbf{Spectral and Energy Efficiency: }
Effective throughput or spectral efficiency is one of the most commonly desired features in UAV-assisted communications, alongside energy efficiency~\cite{zeng2018cellular}. However, the spectral and energy efficiencies are direct trade-offs of each other. Throughput ensures reliable communication and energy consumption in UAVs can be considered as one of the key factors for the execution of missions as it can have a direct impact on other QoS, such as latency and the safety of the UAV itself. The depletion of the UAV battery can be due to several factors, such as weather, the speed of the UAV, the power consumption for trajectory alignment, and maneuvering. There is a demand for models that optimize the energy consumption of UAVs while transmitting or receiving information and maintain an efficient throughput. 

\textbf{Age of Information (AoI): }
\textcolor{black}{AoI quantifies how fresh or new the data is at the UAV. AoI measures the time lapse since the latest reception of the packet update at an information recipient \cite{lyu2021optimizing}. The freshness of data can help in providing optimized control and better reliability. The trajectory of the flight plays a key role in determining the freshness of the information collected. However, the UAV and the ground network have to compromise regarding power efficiency and data transmission latency to obtain fresh data. This can be alleviated by careful trajectory planning and sensor selection during data collection.}

\textbf{Reliability: } Communication reliability is typically measured by the signal-to-noise ratio (SNR)~\cite{yang2020performance}, signal-to-noise-plus-interference ratio (SINR)~\cite{geraci2018understanding}, or outage probability~\cite{abualhaol2006outage}. Different from other radio systems where reliability is primarily subject to the fading conditions of wireless channels and the mobility of the transmitter or receiver, the reliability of UAV-assisted communication systems can deteriorate because of jitters due to the inherent random wind gusts. This can result in an angle ambiguity, where the information beams between the UAV-mounted base station and the user equipment can be misaligned. This gives rise to a need for techniques that can predict the angles between the UAV and the user equipment so that the UAV and the user equipment can prepare the transmit and receive beams in advance. However, there can be some trade-offs in terms of delay and energy efficiency to establish reliable transmission.



\textbf{Resource Utilization: }
Due to the limited resources (i.e., limited bandwidth, limited energy of each node) in the system, there is a demand to efficiently allocate the limited resources to improve the total data transmitting rate. This raises a need for learning techniques that can understand the environment and allocate the resources accordingly. In order to efficiently utilize the bandwidth, resource allocation techniques are needed to dynamically select caching contents and allocate transmission channels through learning \cite{liu2020artificial}.

\textcolor{black}{
\section{Survey of Surveys}\label{sec. gap of survey}}
Some earlier studies review more general applications of UAVs to communications, networking, and data collection, where the UAVs are controlled in more traditional ways, such as control theory or conventional optimization. The challenges in UAV communication have been discussed in several works such as \cite{gupta2015survey, krishna2017review, lu2018wireless, 8675384} and \cite{popescu2019survey}. Some of the heavily featured challenges include  cyber-security threats, demand for energy efficiency, stable communication, and monitoring. The survey in \cite{8918497} provides a tutorial overview on UAV communications with an emphasis on integrating UAVs into fifth-generation (5G) communications and future cellular networks. A few general surveys \cite{hayat2016survey} and \cite{aicha2020survey} discuss experimental results from UAV-aided applications, multi-UAV projects, testbeds, and simulation environments. There have been some surveys pertaining to the area of UAV channel modeling. For instance, the studies \cite{khawaja2019survey} and \cite{khuwaja2018survey} focus on air-to-ground channel measurement campaigns and modeling practices, and provide a survey of channel measurement methodologies and characterization efforts. The survey in \cite{yan2019comprehensive} presents a comprehensive view of UAV communications pertaining to aeronautical channel modeling in line with the specific aeronautical characteristics and scenarios. The authors in this work provide a design guideline for managing the link budget of UAV communications with respect to the link losses and channel fading effects. The authors of \cite{oubbati2020softwarization} emphasize software-defined networking (SDN)-enabled UAV-assisted systems in UAV-assisted cellular communications, monitoring, and routing; and reveal that the usage of ML approaches helps alleviate many challenges in these networks.

{\color{black}
The surveys in \cite{baltruvsaitis2018multimodal, buskirk2018introduction}, and \cite{mahdavinejad2018machine} showcase how ML aims to have the ability to build models that can process and relate information. Novel ML techniques, such as representation learning, deep learning, distributed and parallel learning, transfer learning, active learning, and kernel-based learning, have enabled big data processing \cite{qiu2016survey}. UAVs present us with a wide variety of applications that can benefit from the capabilities of machine learning. Several surveys, e.g.,~\cite{das2017survey, boutaba2018comprehensive, kim2021review}, showcase the prowess of ML in several application domains such as speech recognition, IoT, computer vision, bio-surveillance, robotic control, and empirical experiments \cite{khalid2014survey, dara2018feature}. A brief summary of notable ML surveys is provided in Table \ref{table2}. 
}

\noindent \\

\begin{table}[h!]

\caption{General surveys on ML-aided applications}
\centering

\begin{tabular}{l|l|l}

\begin{tabular}[c]{@{}l@{}} \textbf{Paper} \\ \textbf{and year}\end{tabular} &
  \begin{tabular}[c]{@{}l@{}} \textbf{Applications}\\ \textbf{covered}\end{tabular} &
  \begin{tabular}[c]{@{}l@{}}\textbf{ML techniques}\\ \textbf{covered}\end{tabular} \\
  \hline
\begin{tabular}[c]{@{}l@{}}\cite{baltruvsaitis2018multimodal}\\ 2018\end{tabular} &
  Multimodal ML &
  \begin{tabular}[c]{@{}l@{}}RNN\\ CNN\end{tabular} \\
  
  \hline

\begin{tabular}[c]{@{}l@{}}\cite{buskirk2018introduction}\\ 2018\end{tabular} &
  \begin{tabular}[c]{@{}l@{}}Predictive modeling \\ perception\end{tabular} &
  SVM \\
  
  \hline

\begin{tabular}[c]{@{}l@{}}\cite{mahdavinejad2018machine}\\ 2018\end{tabular} &
  \begin{tabular}[c]{@{}l@{}}IoT,\\ Data analysis\end{tabular} &
  \begin{tabular}[c]{@{}l@{}}K means\\ Distributed ML\end{tabular} \\

\hline

\begin{tabular}[c]{@{}l@{}}\cite{qiu2016survey}\\ 2016\end{tabular} &
  \begin{tabular}[c]{@{}l@{}}Big data\\ Processing\end{tabular} &
  \begin{tabular}[c]{@{}l@{}}Transfer learning\\ Meta learning\\ DNN\end{tabular}\\
\hline

\begin{tabular}[c]{@{}l@{}}\cite{das2017survey}\\ 2017\end{tabular} &
  \begin{tabular}[c]{@{}l@{}}Speech recognition\\ computer vision, \\ perception\end{tabular} &
  \begin{tabular}[c]{@{}l@{}}SVM\\ ANN\\ CNN\end{tabular} \\
  
  \hline

\begin{tabular}[c]{@{}l@{}}\cite{boutaba2018comprehensive}\\ 2018\end{tabular} &
  \begin{tabular}[c]{@{}l@{}}Traffic prediction\\ QoS management\end{tabular} &
  \begin{tabular}[c]{@{}l@{}}ANN\\ CNN\\ DNN\end{tabular} \\
  
\hline
\begin{tabular}[c]{@{}l@{}}\cite{kim2021review}\\ 2021\end{tabular} &
  Robotics &
  \begin{tabular}[c]{@{}l@{}}Supervised learning\\ Unsupervised learning\end{tabular} \\  
\hline  

\begin{tabular}[c]{@{}l@{}}\cite{khalid2014survey}\\ 2018\end{tabular} &
  Feature extraction &
  DNN \\

\hline
\begin{tabular}[c]{@{}l@{}}\cite{dara2018feature}\\ 2018\end{tabular} &
  Feature extraction &
  ANN \\

\end{tabular}
\label{table2}
\end{table}

\subsection{ML-aided UAV Communications}
Several papers have been devoted to reviewing the status quo of ML's applications to UAV communications and control with a particular emphasis on channel modeling, security, resource management, path planning, control, and navigation. Existing surveys, such as \cite{bithas2019survey, carrio2017review} and \cite{lahmeri2021artificial}, provide a detailed survey of several ML frameworks that have been deployed for UAV-assisted communications. They target at functional aspects, including channel modeling, positioning, and security. The surveys presented in \cite{challita2019machine} and \cite{alsamhi2019survey} expose some security challenges in interference suppression, hand-off assistance, cyber-physical risks, and identity and information authentication. 

\textcolor{black}{Technologies that are used to realize UAV-aided communications, such as AI, ML, deep reinforcement learning (DRL), mobile edge computing (MEC), and software-defined networks (SDN), have been surveyed in \cite{ullah2020uavs}, where joint optimization problems for enhancing UAV system efficiency have been reviewed. The survey in \cite{ben2022uav} focuses on UAV-centric ML solutions to UAV-aided communication, which elaborates on the various roles that UAVs can play in the context of collaboration, cooperation, and changing network dynamics. The domain of UAV-enabled mobile edge computing and the need for UAVs’ cooperation in 5G/6G networks are also covered in~\cite{ben2022uav}.}

\subsection{ML-aided UAV Control}
Recent ML-aided UAV communication papers predominantly focus on DRL due to its prowess in supporting online path planning and navigation. The surveys presented in \cite{azar2021drone} and \cite{zhao2018survey} aggregate reinforcement learning techniques under three application domains, namely, path planning, navigation, and control, and provide interesting simulation results on the average reward of several DRL techniques for hovering, landing, random way-point, and target-following tasks. Another survey \cite{luong2019applications} overviews the applications of DRL techniques from a network perspective, such as network access control, smart caching, mobile edge computing, cyber security, physical connectivity, resource management, and information collection. There are also survey works, such as \cite{wang2019survey} and \cite{shakeri2019design}, that pinpoint fundamental design challenges of multi-UAV systems and UAV-aided cyber-physical systems (CPS). They explore interesting aspects of UAV applications, such as target monitoring and tracking, auto-piloting and navigation, and image processing by employing ML techniques. 

\subsection{ML-aided UAV Feature Extraction and Perception}
The survey in \cite{hamylton2020evaluating} showcases techniques for mapping island vegetation from UAV images. In particular, they evaluate the ML approach using CNN to leverage spatial information from the UAV images within the architecture of the learning framework. Thanks to the aerial monitoring capabilities of the UAVs, they are extensively used in surveillance applications. In line with traffic management, parking lot management, and facilitating rescue operations in disaster zones and rugged terrains, detection of on-ground vehicles is becoming a vital spot of UAV applications. \textcolor{black}{The survey in \cite{srivastava2021survey} presents a survey of deep learning techniques for performing on-ground vehicle detection from aerial imagery captured using UAVs, where the approaches taken for improving the accuracy and alleviating the computation overhead and their optimization objective are summarized and discussed.}

\textcolor{black}{Table \ref{table3} lists and categorizes the existing surveys on UAVs based on their applications and performance indicators covered. Table \ref{table4} covers the existing surveys of ML applications to UAV-aided communications and operations, but does not focus on how ML could aid the applications covered. The surveys in Table \ref{table5} predominantly cover a specific usage of certain ML techniques for specific UAV applications. In this survey, we aim to provide a broader view of the applications of ML in the context of UAV operations by offering reasoning, advantages and drawbacks towards the usage of specific ML techniques for an application.}

\begin{table*}[]
\centering
\caption{Existing surveys on the application of UAVs}
\begin{tabular}{llll}
\centering
\textbf{Paper} & \textbf{Short description} & \textbf{Performance metrics} & \textbf{Application covered}   \\ \hline

\begin{tabular}[c]{@{}l@{}}\cite{gupta2015survey} \\ \textbf{2015}\end{tabular}  & \begin{tabular}[c]{@{}l@{}}Survey on the issues \\ that hinder the stability of UAV network\end{tabular}                                     & \begin{tabular}[c]{@{}l@{}}Reliability, mobility, \\ energy consumption\end{tabular}                          & Communication                                                                   \\ \hline

\begin{tabular}[c]{@{}l@{}}\cite{hayat2016survey} \\ \textbf{2016}\end{tabular} 
   & \begin{tabular}[c]{@{}l@{}}Survey on applications\\  of UAVs for civil applications\end{tabular} & \begin{tabular}[c]{@{}l@{}}Connectivity, adaptability, \\ safety, security, scalability\end{tabular} & \begin{tabular}[c]{@{}l@{}}Communication,\\ aerial imaging\end{tabular}                                                               \\ \hline

\begin{tabular}[c]{@{}l@{}}\cite{krishna2017review} \\ \textbf{2017}\end{tabular}  & \begin{tabular}[c]{@{}l@{}}A review on cybersecurity vulnerabilities \\ for unmanned aerial vehicles\end{tabular}                            & Security                                                                                                      & \begin{tabular}[c]{@{}l@{}}Control-based \\ applications\end{tabular}                                                                              \\ \hline

\begin{tabular}[c]{@{}l@{}}\cite{khuwaja2018survey} \\ \textbf{2018}\end{tabular} & \begin{tabular}[c]{@{}l@{}}A survey of channel modeling for \\ UAV communications\end{tabular}                                               & Packet loss, reliability                                                                                      & \begin{tabular}[c]{@{}l@{}}Communication\\ Data collection\end{tabular}        
                                                                  \\ \hline

\begin{tabular}[c]{@{}l@{}}\cite{lu2018wireless} \\ \textbf{2018}\end{tabular}  & \begin{tabular}[c]{@{}l@{}}A survey of wireless charging \\ techniques for UAVs\end{tabular}                                                 & Energy efficiency                                                                                             & \begin{tabular}[c]{@{}l@{}}Imaging\\ Communication\end{tabular}              
                                                                  \\ \hline

\begin{tabular}[c]{@{}l@{}}\cite{khawaja2019survey} \\ \textbf{2019}\end{tabular}  & \begin{tabular}[c]{@{}l@{}}A survey of air-to-ground propagation \\ channel modeling for unmanned aerial vehicles\end{tabular}               & Packet loss, throughput                                                                                       & \begin{tabular}[c]{@{}l@{}}Communication,\\ data collection\end{tabular}    
                                                                   \\ \hline

\begin{tabular}[c]{@{}l@{}}\cite{popescu2019survey} \\ \textbf{2019}\end{tabular} & \begin{tabular}[c]{@{}l@{}}Comparison of multiple theoretical\\  and applied contributions in UAV-WSN\end{tabular} & Reliability, energy efficiency                                                                                & \begin{tabular}[c]{@{}l@{}}Imaging,\\ communication \end{tabular}    
                                                                   \\ \hline

\begin{tabular}[c]{@{}l@{}}\cite{8675384} \\ \textbf{2019}\end{tabular} & \begin{tabular}[c]{@{}l@{}}Survey on UAV cellular communications,\\  regulations and security challenges\end{tabular} & Security & Communications                                    \\ \hline   

\begin{tabular}[c]{@{}l@{}}\cite{aicha2020survey} \\ \textbf{2020}\end{tabular}  & Survey of UAVs communication networks                   & Security, mobility, reliability   & \begin{tabular}[c]{@{}l@{}}Data collection,\\ monitoring, tracking\end{tabular}  
\\ \hline

\begin{tabular}[c]{@{}l@{}} [35] \\ \textbf{2020}\end{tabular} & \begin{tabular}[c]{@{}l@{}}Survey on SDN oriented \\  UAV-assisted systems.\end{tabular} & throughput, computational delay & \begin{tabular}[c]{@{}l@{}} communications  \\ monitoring\end{tabular}        \\ \hline

\end{tabular}
\label{table3}
\end{table*}

\begin{table*}[]
\centering
\caption{Existing surveys on the applications of ML to UAV-assisted communications}
\begin{tabular}{lllll}
\centering
 \begin{tabular}[c]{@{}l@{}} \textbf{Paper}  \\ \textbf{and year} \end{tabular}  & \textbf{Short Description} &\textbf{Performance metrics} & \textbf{Application covered} & \textbf{ML Techniques} \\ \hline
 
\begin{tabular}[c]{@{}l@{}} \cite{wang2019survey}  \\ \textbf{2017} \end{tabular}   & \begin{tabular}[c]{@{}l@{}}Review of UAV networks  \\ from the CPS perspective \end{tabular}  & \begin{tabular}[c]{@{}l@{}}Resource allocation, packet loss\\ reliability, energy efficiency\end{tabular}   & \begin{tabular}[c]{@{}l@{}}Communication, control, \\ computation \end{tabular}  & Reinforcement learning \\ \hline 

\\
 \begin{tabular}[c]{@{}l@{}} \cite{zhao2018survey} \\ \textbf{2018} \end{tabular}   & \begin{tabular}[c]{@{}l@{}} Survey of computational  
\\ intelligence algorithms in \\  UAV path planning \end{tabular}  & \begin{tabular}[c]{@{}l@{}}Trajectory planning 
\\ and optimizing trajectory \end{tabular}   & UAV control
  & \begin{tabular}[c]{@{}l@{}} Supervised learning,
 \\unsupervised learning\end{tabular} \\ \hline 
 \\
 
 \\

\begin{tabular}[c]{@{}l@{}}\cite{bithas2019survey}  \\ \textbf{2019} \end{tabular}   & \begin{tabular}[c]{@{}l@{}} Survey on ML techniques \\  for UAV communications\end{tabular}  & \begin{tabular}[c]{@{}l@{}}Resource allocation, packet loss\\ security\end{tabular}   & Communications, control  &                                                                \begin{tabular}[c]{@{}l@{}}Supervised,  unsupervised ML \\ deep learning techniques\end{tabular}      \\ 
\hline  

\\
 \begin{tabular}[c]{@{}l@{}} \cite{luong2019applications}  \\ \textbf{2019} \end{tabular}   & \begin{tabular}[c]{@{}l@{}} Overview of DRL \\  from a network and  \\  communication aspect\end{tabular}  & \begin{tabular}[c]{@{}l@{}}Resource allocation, packet loss\\ reliability, energy efficiency\end{tabular}   & \begin{tabular}[c]{@{}l@{}}Path planning, control, \\ feature extraction \end{tabular}  & \begin{tabular}[c]{@{}l@{}}Deep reinforcement \\ learning \end{tabular}  \\
\hline  

\\
 \begin{tabular}[c]{@{}l@{}}  \cite{challita2019machine}  \\ \textbf{2019} \end{tabular}  & \begin{tabular}[c]{@{}l@{}} A survey on the security \\ challenges in \\ UAV-assisted WSN\end{tabular}  & \begin{tabular}[c]{@{}l@{}} Interference management\\ security\end{tabular}   & Communication  & \begin{tabular}[c]{@{}l@{}} CNN, RNN \\ DRL\end{tabular} \\ \hline  

\\
 \begin{tabular}[c]{@{}l@{}}  \cite{alsamhi2019survey}  \\ \textbf{2019} \end{tabular}  & \begin{tabular}[c]{@{}l@{}}Outline on the usage of \\ ML in UAV -assisted \\ robotic applications\end{tabular}  & \begin{tabular}[c]{@{}l@{}}Connectivity, delay\\ security\end{tabular}   &  Control applications  & Deep learning techniques \\ \hline  

\\
 \begin{tabular}[c]{@{}l@{}} \cite{shakeri2019design}  \\ \textbf{2019} \end{tabular}  & \begin{tabular}[c]{@{}l@{}} Survey on the  design \\ challenges of  multi-UAV \\ for IoT applications\end{tabular}  & \begin{tabular}[c]{@{}l@{}}Resource allocation, packet loss\\ reliability, energy efficiency\end{tabular}   & \begin{tabular}[c]{@{}l@{}}Path planning, control, \\ feature extraction \end{tabular}  & \begin{tabular}[c]{@{}l@{}} Unsupervised learning, \\reinforcement learning\end{tabular} \\ \hline

 \\
 \begin{tabular}[c]{@{}l@{}} \cite{carrio2017review}  \\ \textbf{2019} \end{tabular}   & \begin{tabular}[c]{@{}l@{}}Applications of AI\\  deep reinforcement  in \\ UAV-based networks\end{tabular}  & \begin{tabular}[c]{@{}l@{}}Resource allocation, packet loss\\ reliability, energy efficiency\end{tabular}   & \begin{tabular}[c]{@{}l@{}}Path planning, control \\ feature extraction \end{tabular}  & \begin{tabular}[c]{@{}l@{}} unsupervised learning, \\reinforcement learning\end{tabular} \\ \hline  
 
 \\
 \begin{tabular}[c]{@{}l@{}} \cite{azar2021drone}  \\ \textbf{2021} \end{tabular}   & \begin{tabular}[c]{@{}l@{}}Survey on drone \\ DRL techniques\end{tabular}  & \begin{tabular}[c]{@{}l@{}}Resource allocation, packet loss,\\ reliability, energy efficiency\end{tabular}   & \begin{tabular}[c]{@{}l@{}}Communication, control, \\ caching \end{tabular}                                                                & \begin{tabular}[c]{@{}l@{}}Deep reinforcement\\ learning techniques\end{tabular}     \\ 
\hline  
\\

 \begin{tabular}[c]{@{}l@{}} \cite{lahmeri2021artificial}  \\ \textbf{2021} \end{tabular}   & \begin{tabular}[c]{@{}l@{}} Applications of AI\\ using DRL  in\\ UAV-assisted networks\end{tabular}  & \begin{tabular}[c]{@{}l@{}}Resource allocation, packet loss,\\ reliability, energy efficiency\end{tabular}   & \begin{tabular}[c]{@{}l@{}}Communication, control, \\ imaging\end{tabular}  & \begin{tabular}[c]{@{}l@{}} Supervised, unsupervised, \\reinforcement learning, federated learning\end{tabular}      \\  
\hline

\end{tabular}
\label{table4}
\end{table*}

\section{Background of ML for UAV applications}\label{sec: performance vs ML}

ML algorithms can be classified based on their usage of datasets to extract data and how the models are trained. It is necessary and efficient to organize the ML algorithms with respect to learning methods when one needs to consider the significance of the training data and choose the classification rule that provides a greater level of accuracy. Some of the common ML techniques which have been used for UAV operations and communications are described below. 

\subsection{Supervised Learning}
The usage of labeled datasets defines supervised learning for training algorithms to classify data or predict outcomes. The weights are adjusted following the input data that is fed into the model. Supervised learning helps in classifying or segmenting data accurately. Supervised learning uses a training set to teach models and obtain the desired output. Every training dataset comprises of inputs from the environment trained over a training period to get optimal output. With the help of a loss function, the model's accuracy is adjusted until the desired outcome is obtained.
\\

\textbf{Convolutional Neural Network (CNN)}: CNN is a commonly used ML technique for feature extraction that allows for generating high-dimensional feature images from the raw sensor data acquired for the IoT system. Due to its support for feature extraction functionalities, CNN is commonly used for the classification of images, image, and video recognition, analyzing medical images, computer vision, and language processing. In a CNN, images are represented in the form of matrices, then they are multiplied with each other to obtain an output from which the image features are extracted.  

The basic CNN architecture comprises of three layers: a convolution layer, a pooling layer, and a connection layer~\cite{o2015introduction}. The convolution layer detects various features from input images. A primary convolution is carried out between the input image and a filter. The output of this convolution is the feature map that provides us with information about the image, such as the edges and corners~\cite{xu2018vision}. As convolution can incur high computational costs, pooling is done to decrease the size of the convoluted feature map. In order to achieve this, the connections between the layers and independently operating on each feature map are reduced. The input images from the convolution and pooling layers are flattened and fed to the connected layer. This flattened vector then goes through one or more connected layers to have a classification in the images.
\\

\textbf{Recurrent Neural Network (RNN)}: 
The main advantage of RNN is that it can model a data sequence, generally in time series, by assuming that one sample is dependent on the previous one. The recurrent structure of these networks  results in a long time for training. In general, the training time for RNNs is much longer than those of feed-forward networks.  Also, implementation of RNNs can be challenging since they require calibrating the previous outputs and the current inputs into a state change function per node.
\\

\textbf{Multilayer Perceptron (MLP)}: A perceptron is an algorithm under supervised learning. It is very useful for classifying linearly separable datasets. It is usually used in binary classifiers in which it is decided whether an input belongs to a specific class. A perceptron can be considered as a single-layer neural network with input values, weights and bias, net sum, and an activation function. An MLP can classify datasets that are not linearly separable. As shown in Fig. 4, a traditional MLP comprises of an input layer, an output layer, and multiple hidden layers between them. The inputs move forward through the MLP by taking the multiplication of the input with the weights existing between the hidden and the input layers. An MLP uses an activation function at every layer. The output that is defined at every step is pushed forward using the activation function. The MLP uses backpropagation for training along with its multiple layers.
\\

\begin{figure}[h!]
\centering
\includegraphics [width= 3.4 in]{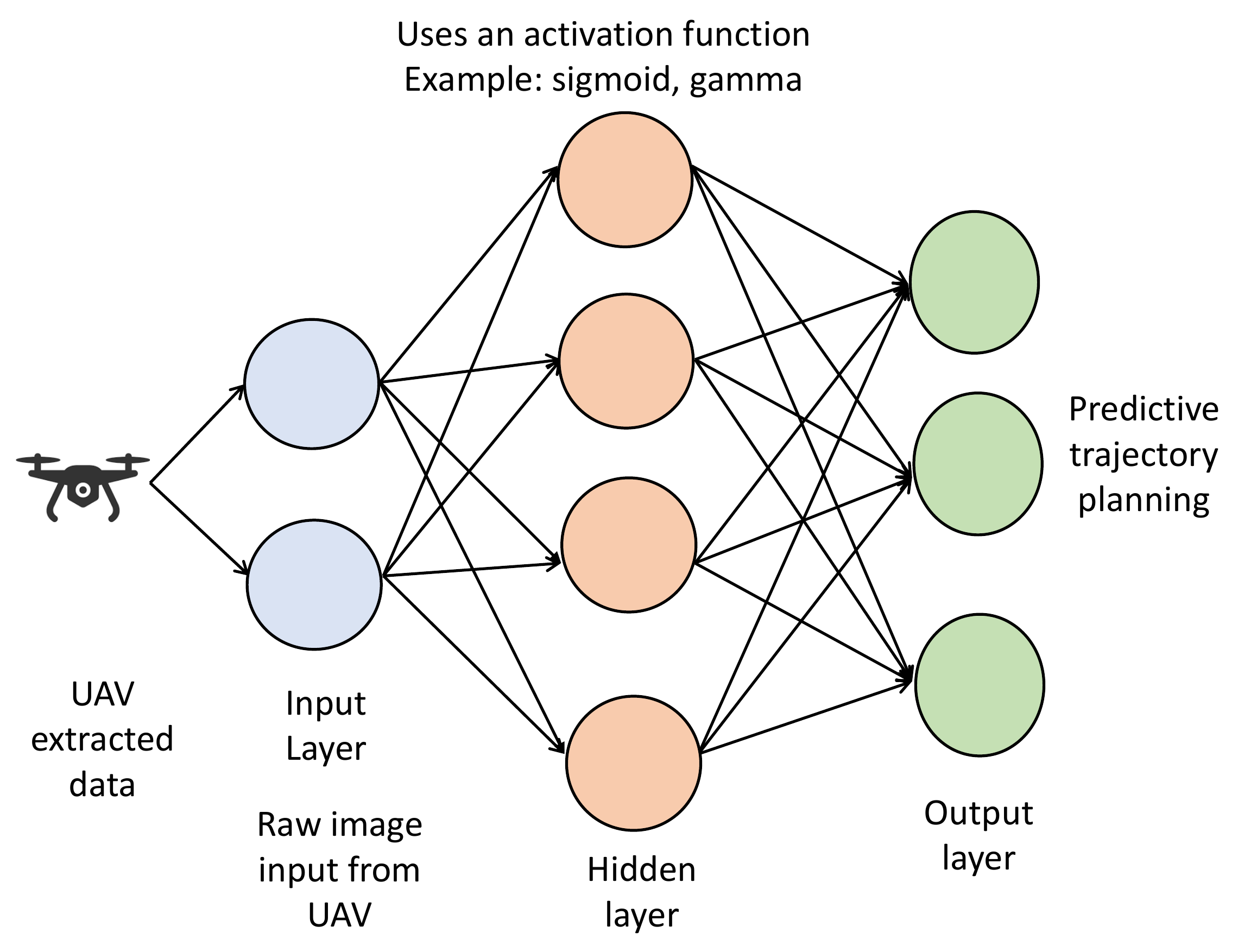}
\caption{An illustration of the layers of an MLP with an activation function in the hidden layers that can vary based on the inputs.}
\label{mlpstruct}
\end{figure}

\textbf{Linear regression}: The idea of linear regression revolves around finding the linear relationship between the dependent and independent variables. The key objective of linear regression models is to find the best fit linear line and the optimal values of intercept and coefficients, such that the error is minimized~\cite{niu2019estimating}. The regression error can be defined as the difference between the actual and predicted values, and the goal is to reduce this difference. The training is accomplished to obtain the best fit, where the error between predicted and actual values is minimized~\cite{zhang2020estimation}. 

\subsection{Unsupervised Learning}
The usage of unlabelled datasets defines unsupervised learning for training algorithms to classify data. The respective algorithm employed helps to discover patterns to solve clustering or association problems from unlabelled data. Unsupervised learning is usually used when some common properties within a dataset are not apparent.
\\

\textbf{Generative Adversarial Network (GAN)}: In a GAN, two neural networks compete with each other in the form of a zero-sum game, where the gain of an agent is the loss of the other~\cite{yang2019attack}. As shown in Figure \ref{GANstruct}, the basic structure of a GAN comprises  a real dataset, random fake data is provided to the discriminator to be analyzed and to meet the condition \cite{zhang2021distributed}. GAN is typically used to discover and learn regularities or patterns in input data, where a generative network and a discriminative network are trained simultaneously in an adversarial manner~\cite{goodfellow2020generative}. 
\\

\begin{figure}[h!]
\includegraphics [width=3.5in]{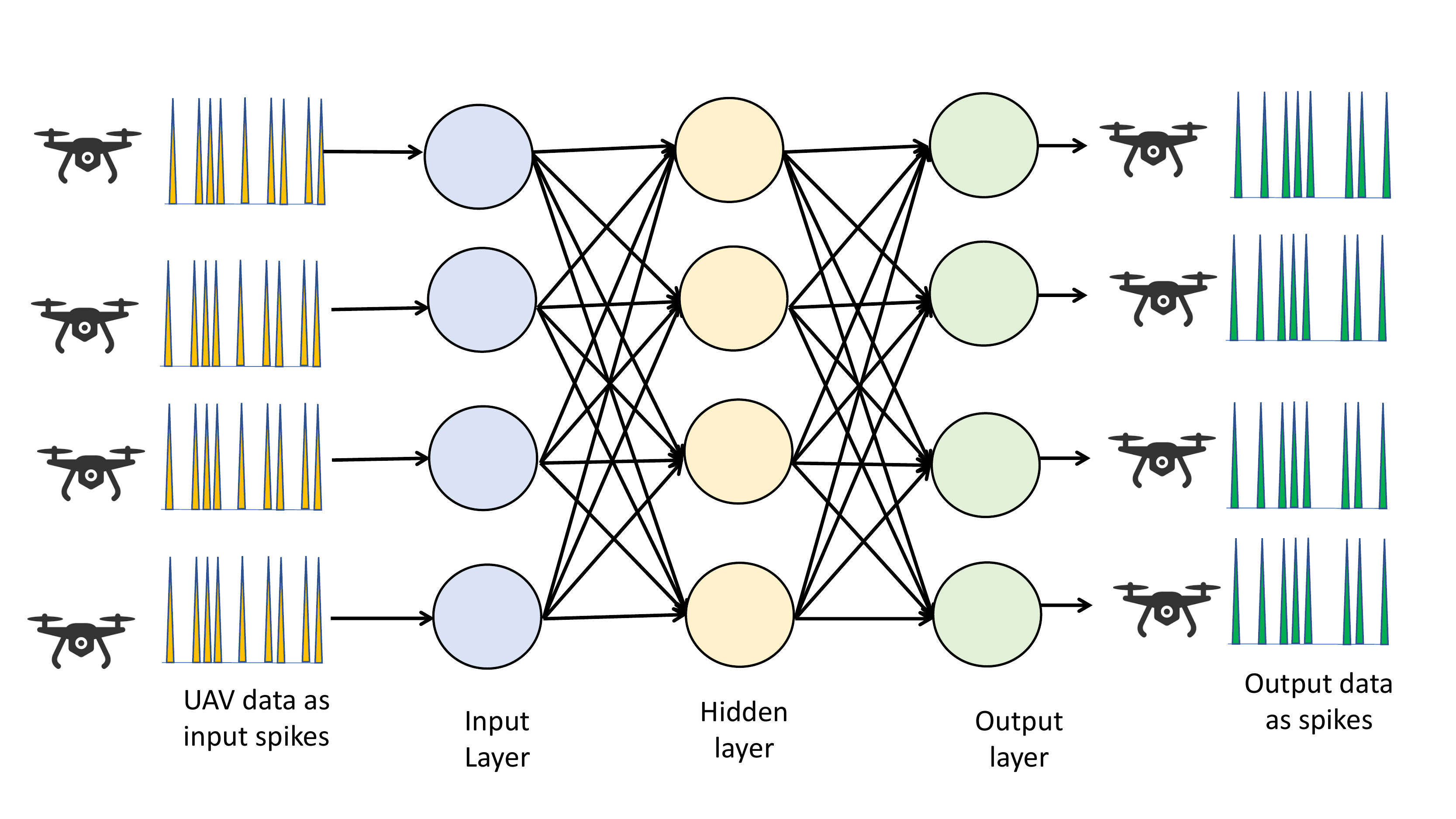}
\caption{An illustration of SNN providing the UAV output in the form of data spikes which are discrete events along the timeline.}
\label{snnstruct}
\end{figure}

\textbf{Spiking Neural Network (SNN)}: Unlike the MLP models that periodically propagate information, spiking neural networks (SNNs) use the concept of spikes, which are discrete events along a timeline (Fig. \ref{snnstruct}). The spikes can be determined by differential equations that represent various biological processes, the most critical part of which is the membrane potential of a neuron~\cite{ros2006event}. A spike occurs when the neuron reaches its potential. The SNNs have the capability of processing spatio-temporal data. The neurons are locally connected and process a large quantity of input data. Spikes allow temporal data to be processed without the extra complexity, as opposed to RNNs~\cite{he2020comparing}.
\\

\textbf{Autoencoder}: \textcolor{black}{Autoencoders are composed of an encoder network and a decoder network that aim to minimize the training error between input data and reconstruction of the input data.} To achieve the equality of target values and the input data, the encoder network aims to transform the input signal into a low-dimensional code, while the decoder network is used to reconstruct the data from the code. 

\subsection{Semi-supervised Learning}
Semi-supervised learning combines small quantities of labeled data and large quantities of unlabeled data during training. When unlabelled data are utilized with a small amount of labeled data, the overall learning accuracy can be significantly improved. Semi-supervised learning combines the features of clustering and classification algorithms. Clustering is an unsupervised ML technique that groups data based on their similarities. This clustered data is labeled and used in the training of a supervised ML model for classification. 
\\

\textbf{K-means clustering}: It takes a number $K$ as the input, groups samples one after another into the nearest cluster, and determines the centroids of the clusters repeatedly. Originally designed to be unsupervised learning, K-means clustering has been increasingly used as a semi-supervised learning technique to improve its clustering accuracy with the assistance of some training data. In the event of convergence, the input data is grouped towards a centroid and does not change position with further training.  K-means clustering has been used to process images. For example, it is used to label pixels based on color information for rice yield estimation in~\cite{REZA2019109}. The area of the rice grains is then calculated from the clustered images. K-means clustering is easy to implement because it classifies a given dataset according to the distance information. However, since only the distance information is used for clustering, it has a high probability of producing unbalanced clusters. Specifically, the number of elements in a cluster depends on the distribution of the elements. When the elements are non-uniformly distributed, some clusters could be large while others are small. 
\\

\textbf{Gaussian Mixture Modeling (GMM):} It is also a widely used method for clustering~\cite{zhang2018machine}. GMM is a probabilistic method, which distinguishes it from the deterministic counterpart K-means. As opposed to the rigid association policy in K-means, GMM computes clusters using Gaussian distributions. In this sense, each sample point is associated to a cluster with a probability.
Just like K-means clustering, GMM was originally designed to be unsupervised learning but has been increasingly used as semi-supervised learning in the presence of small amounts of training data.

\subsection{Reinforcement Learning (RL)}
Under this learning setting, an intelligent entity referred to as ``agent'', learns an optimal or near-optimal policy that maximizes the "reward function". The reinforcement signal provided by the other users in the systems also accumulates from the immediate rewards. Over the timeline, the agent receives the current state and reward. Based on the respective reward and state, an immediate action is chosen from the set of actions. Following this, the environment changes to a new state, and the reward associated with the transition is obtained. In different system states, the actions are taken with the objective to maximize  the cumulative reward. 
\\

\textbf{Single-agent and multi-agent Deep Reinforcement learning (DRL)}: 
DRL incorporates deep learning into RL, where the agent implements a deep neural network (DNN) to approximate the Q-value for evaluating its action-value function, as opposed to looking up a Q-table (as done in RL). Employing a single agent, DRL is also known as single-agent DRL. It has the potential to assist the agent in making sequential decisions or actions based on unstructured input data sampled from a much larger state space, as compared to RL. 
Single-agent DRL includes a Deep Q Network (DQN) that can solve learning problems containing a large discrete state or action space~\cite{emami2021deepQ}. It also includes a Deep Deterministic Policy Gradient (DDPG) algorithm that enables DRL to operate under continuous state and action spaces by taking an actor-critic architecture~\cite{li2020onboardDDPG}. Other single-agent DRL algorithms include Proximal Policy Optimization (PPO)~\cite{Toan2021BigComp}, and variants of DQN and DDPG, such as double DQN~\cite{DDQN2016AAAI} and Twin Delayed DDPG~\cite{Dankwa2019TD3}.

Multi-agent DRL is the extension of single-agent DRL in support of multiple agents, where multiple agents train  a DRL model collaboratively for fast action exploration and rapid convergence. Some examples of multi-agent DRL are multi-agent DQN~\cite{Yousef2021TVT} and multi-agent DDPG~\cite{DBLP:journals/twc/WangLNBG21}. 

\subsection{Other ML Techniques}
{\color{black}In addition to the conventional ML techniques, there are other ML algorithms and techniques such as spatially decentralized learning structure (e.g., federated learning~\cite{yang2019federated}, distributed learning~\cite{chen2021towards}), temporally transferable learning mode (e.g., transfer learning~\cite{bonatti2020learning}, and meta-learning~\cite{hu2020meta}).}
\\


\begin{figure*}[h!]
\centering
\includegraphics [width= 6.7 in]{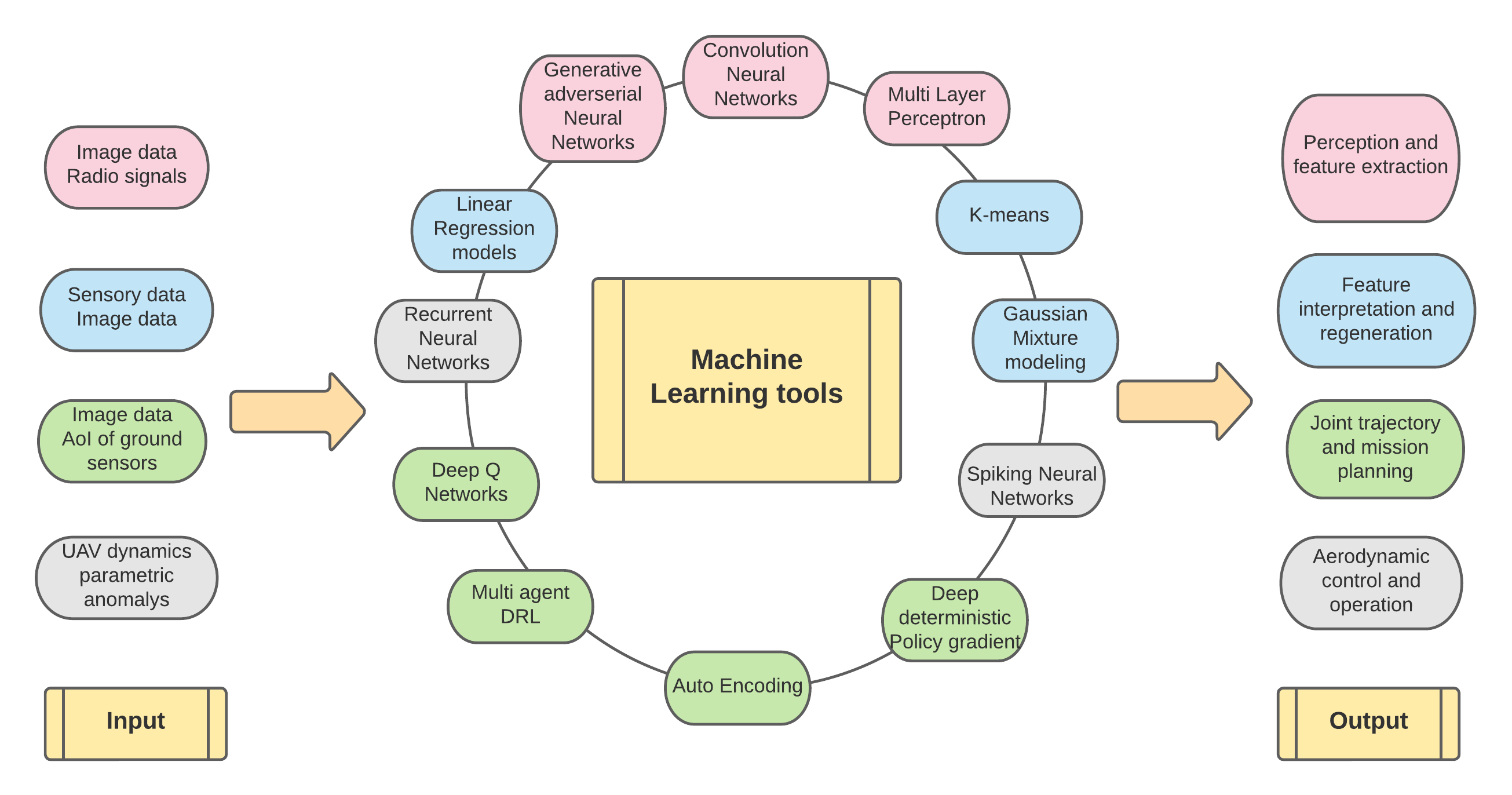}
\caption {Different ML techniques surveyed in this paper and their suitable application domains.}
\label{mltools}
\end{figure*}

\textcolor{black}{
\textbf{Distributed learning:}
In general, in a distributed ML technique, centralized data can be distributed among the worker devices (or nodes) for learning  \cite{chen2021towards}. With distributed ML, a data-parallel approach can be used where different nodes use different sets of data, while in a model-parallel approach, the same set of data is used by different nodes  to learn a global model.  Distributed ML techniques aid in making informed decisions and analysis from large amounts of data~\cite{peteiro2013survey}.}\\

\textcolor{black}{
\textbf{Federated learning:}
Federated learning \cite{yang2019federated} is a decentralized  ML technique where the algorithm is trained across servers holding local data samples or multiple decentralized edge devices. The training can eventually benefit from the data set across multiple servers than a centralized local server. Federated learning models follow a three-step process; initially, a subset of the learned updates are aggregated, then this updated data is used to form a consensus change and finally it is updated into the shared model. This process is done in a loop to improve the learning for every processing time. Meta-learning can be devised by personalizing federated learning methods to edge users. Federated learning has been enabling IoT applications such as autonomous vehicles \cite{du2020federated, elbir2020federated}, Industry 4.0, \cite{qu2020blockchained} and digital health \cite{rieke2020future}. } 
\\

\textbf{Transfer learning:}
It is a novel ML technique that uses the results obtained from one model and re-utilizes the solution for a different but related problem. Transfer learning can utilize both labeled and unlabeled data to train a model. Transfer learning can choose the developing model approach when a large amount of data is available to train the model for the initial result grab. On the other hand, it can also use the pre-trained model, where already trained datasets can be reused to reach a solution for the problem. This method has been predominantly utilized for classification approaches due to its capability of utilizing already classified information, thus reducing the strain on processing which is common in classic methods like CNN. In the context of ML-aided UAV applications, transfer learning methods have been used to learn the perception-action policies from a simulated environment and then use that knowledge to control an autonomous drone~\cite{bonatti2020learning}.
\\

\textbf{Meta-learning:}
Meta-learning learns from the outcomes of other learning techniques and is analogous to transfer learning in which learning algorithms are applied to metadata. Meta-learning necessitates the use of other learning methods trained on the metadata~\cite{hu2020meta}. By learning the metadata, these models are used to improve the existing models by learning the algorithm itself. Metadata includes the  characteristics of the learning problem, characteristics  of the underlying algorithm properties, and even the patterns derived from the learning experiments. Meta-learning has been used for trajectory planning by learning the dynamic networking environments~\cite{hu2020meta,jiang2021attention}. Meta-learning and transfer learning are very new concepts and they have a limited number of researches compared to the classic ML techniques.
\\


\subsection{Applications of ML to UAV Systems}
Traditionally, ML has been used in several application domains, such as speech recognition, autonomous vehicles, image classification, and wireless communications. ML enables  IoT communications, where many devices can autonomously decide to activate and transmit at the same time by learning the activities on the air interface. Over the past decade, several ML algorithms have been widely used to meet the challenges and demands that an IoT system can impose. Data-driven ML algorithms can make appropriate sequential decisions, adapting to the changing environments. In a UAV-aided communications scenario
, ML can easily monitor and learn the changes in radio fading channels, traffic patterns, user context, and device positions and take appropriate decisions to provide QoS for communications.
\\
\\
{\color{black}
\textbf{Trajectory Planning and Mission Scheduling:} 
This is a common practice in UAV-assisted applications to decrease communication latency by following the data buffer occupancy at the UAV to pre-select way-points and alleviate data traffic congestion~\cite{ebrahimi2020autonomous}. Trajectory planning becomes vital for extending the cruise time by deploying strategic charging points to satisfy the UAV's need for energy supply. It is found that trajectories and the communication schedules of the UAV have a direct impact on the network throughput of the users on the ground. There are algorithms~\cite{hu2020meta} to relax the restriction of the UAV’s trajectory in regards to energy efficiency, including the starting point, destination, the maximum and minimum speeds, and the maximum acceleration and deceleration of a UAV. The broadly adopted goal of trajectory planning is to allow UAVs to dynamically change their heading and speed and adjust their displacement or distance to effectively deliver data traffic to devices on the ground.

Backhaul refers to the communication links between the base stations  and the core network. Fronthaul, on the other hand, connects the base stations to the remote radio units. Caching is used in radio access networks to reduce communication latency. As aerial base stations, UAVs enhance the performance of cache-enabled networks by exploiting the backhaul and fronthaul links efficiently using ML-based prediction techniques. For instance, the work in \cite{chen2017caching} uses user-centric information, such as the statistical distribution of requests for contents and the historical patterns of users' mobility, for UAV deployment and smart caching.  Individual user behaviors are classified into distinctive patterns by developing a conceptor-based echo-state network (ESN) method, a class of RNN, on a cloud platform. Effective smart caching policies are created for the UAVs by improving the overall prediction accuracy  through ML-based techniques.}
The availability of computational resources is always a hindrance for UAV communication as it lacks adequate processing power. Researchers have used point cloud processing and trajectory planning to navigate through unknown environments. Due to the nature of the unfamiliar environment, computations can be expensive and will deplete UAV batteries. The authors of \cite{ji2021economy} propose an autonomous navigation system that uses the cloud. In their proposed system, the UAV transmits the point cloud using a cellular network to a cloud that plans the trajectories. The UAV velocity and trajectory are optimized online through learning techniques to handle the communication more precisely.

\textbf{Aerodynamic Control and Operation of UAV:}
 With the lack of consistent energy supply, UAVs have a time constraint for their operation. ML frameworks \cite{abd2019deep,wu2021uav} have been used in optimizing the UAV's flight path, as well as the schedule of signaling to update the UAV with the states of the ground nodes. Using efficient control, the weighted sum AoI can be effectively minimized. Methods like these enable a decent trade-off between QoS metrics such as  energy efficiency and reliability in UAV-assisted caching and edge computing domains. When several UAVs with different trajectories fly over the geographical area and have a stable connection with the sensors on the ground, there is a need for the fairness of the computing load at the UAVs and to reduce the overall energy consumption. ML techniques~\cite{wang2020multi} that model the UAVs’ trajectories help optimize the offloading decisions and, therefore, meet the QoS requirements such as  energy efficiency.

{\color{black}
\textbf{Perception and Feature Extraction in UAV-based IoT:}
Feature extraction is one of the fundamental topics of computer vision. With the advent of UAV-based IoT applications, imagery feature extraction has become one of its core concepts. UAVs can be equipped with many visual sensors, such as cameras, hyperspectral cameras, Lidar, and radar to measure the environment~\cite{son2017brief}. The captured data provides machine perception and environmental information, which can help the UAVs to model the environment for their design and optimization of the mission execution, routing, and collision avoidance~\cite{pajares2015overview}.

\textbf{Feature Interpretation and Regeneration in UAV-based IoT:}
Interpretation is the action of understanding the extracted features  and predicting possible outcomes. Feature interpretation has been widely used for processing the sensory data and taking specific actions such as adjustment of a flight route. Interpretation of features can help in environmental modeling and even optimizing trajectory planning.
One of the examples that a UAV interprets and regenerates its surrounding environment is the recent simultaneous localization and mapping (SLAM) techniques, where discrete cloud points indicating reflections of obstacles and objects can be regenerated to be continuous and differentiable surfaces, e.g., by using GMM, to facilitate trajectory planning. 
}

In Table \ref{table5}, we provide a brief summary of popular ML tools and their applications to the four important modules of UAV operations and communications, i.e., trajectory and mission planning (see Section \ref{sec: planning and scheduling}), aerodynamic control and operation (see Section \ref{sec: control and operation}), UAV perception and feature extraction (see Section \ref{sec: feature extraction}), and feature interpretation and regeneration (see Section \ref{sec: feature interpretation}). 
More details are provided in the following sections.

\begin{table*}[]
\caption{ML-assisted UAV operations and applications, where every ``\checkmark'' represents five research publications in a domain to show the popularity of different ML tools in specific applications. Featuring research works are provided in the table.}

\begin{tabular}{l|llll}
\textbf{ML Techniques \& Remarks} &
  \textbf{\begin{tabular}[c]{@{}l@{}}Trajectory \& \\ Mission Planning\end{tabular}} &
  \textbf{\begin{tabular}[c]{@{}l@{}}Aerodynamic Control \\ \& Operation\end{tabular}} &
  \textbf{\begin{tabular}[c]{@{}l@{}}Perception \&\\  Feature Extraction\end{tabular}} &
  \textbf{\begin{tabular}[c]{@{}l@{}}Feature Interpretation\\  \& Regeneration\end{tabular}} 
  \\
  \hline \\

\begin{tabular}[c]{@{}l@{}} \textbf{Autoencoders}\\ Autoencoders can reconstruct\\ images and videos by training\\ making them apt for\\ prediction and control applications\end{tabular} & \begin{tabular}[c]{@{}l@{}}\checkmark \checkmark\\ e.g., \cite{mesquita2019fully}, \cite{kwak2019autoencoder, dai2018unsupervised}\\ \cite{sarkar2018sequential}\end{tabular}
   &
   &
   & 
   \\ \\

\begin{tabular}[c]{@{}l@{}} \textbf{Deep Q-Network (DQN)}\\ The action space of DQN\\ has to be discrete\\ hence it is primarily used in\\ offline control applications\end{tabular} & \begin{tabular}[c]{@{}l@{}}\checkmark \checkmark \checkmark\\ e.g., \cite{abedin2020data}, \cite{wang2020priority},\\ \cite{li2020online}, \cite{li2020joint}, \cite{zhang2021trajectory}\end{tabular} 
   &
   &
  &
   \\ \\

\begin{tabular}[c]{@{}l@{}} \textbf{Deep Deterministic Policy Gradient (DDPG)}\\ The action space of DDPG is continuous\\ hence it is majorly used in\\ online control applications\end{tabular} & \begin{tabular}[c]{@{}l@{}}\checkmark \checkmark \checkmark\\ e.g., \cite{kurunathan2021deep, li2021continuous},\\ \cite{li2020onboardDDPG,samir2020age,wang2020deep},\\ \cite{peng2020ddpg}\end{tabular}
   &
   &
   &
   \\ \\
   
\begin{tabular}[c]{@{}l@{}} \textbf{Convolution Neural Network (CNN)}\\ With its strong support of\\ feature extraction functionalities\\ CNN is extensively used in\\ image classification based applications\end{tabular} &

  & \begin{tabular}[c]{@{}l@{}}\checkmark\\ e.g., \cite{bojarski2016end,padhy2018deep}\end{tabular}
   & \begin{tabular}[c]{@{}l@{}}\checkmark \checkmark \checkmark\\ e.g., \cite{amorim2019semi},  \cite{ghorbanzadeh2019uav}\\ \cite{kyrkou2018dronet,amorim2019semi,xu2017car}\end{tabular}
   &
    
   \\ \\

\begin{tabular}[c]{@{}l@{}} \textbf{Recurrent Neural Network (RNN)}\\ The sequential problem\\ solving structure of RNN makes it\\ more suitable for\\ control-oriented applications\end{tabular} 
  & 
  \begin{tabular}[c]{@{}l@{}}\checkmark\\ e.g., \cite{xiao2019trajectory}, \cite{pugach2017nonlinear}\end{tabular} 
  & 
  \begin{tabular}[c]{@{}l@{}}\checkmark \checkmark\\ e.g., \cite{challita2019machine,lei2020time},\\\cite{wang2019data}\end{tabular}
  & 
  \begin{tabular}[c]{@{}l@{}}\checkmark\\ e.g., \cite{rahnemoonfar2018flooded}\end{tabular} 
   & 
    \\ \\

\begin{tabular}[c]{@{}l@{}} \textbf{Multi-Layer Perceptron (MLP)}\\ The adaptability of MLP networks\\ to be trained online and\\ offline makes them suitable for\\control-oriented applications\end{tabular} &
  
   & 
   
  \begin{tabular}[c]{@{}l@{}}\checkmark\\e.g., \cite{Gunchenko_harvesting_2017,annepu2020unmanned}\end{tabular} & 
  
  \begin{tabular}[c]{@{}l@{}}\checkmark\\ e.g., \cite{mansouri2017remaining,Gomez-Avila_EKF_2020}\\
  \cite{Hernandez-Barragan_Tracking_2021, Rahnemoonfar_Flood_2021}\end{tabular} &

   \\ \\

\begin{tabular}[c]{@{}l@{}} \textbf{Spiking Neural Network (SNN)}\\ The SNN can process\\ spatio-temporal data\\ making them apt for control\\ and classification applications\end{tabular} &
   & \begin{tabular}[c]{@{}l@{}}\checkmark \checkmark\\ e.g., \cite{salt2019parameter, salt2017obstacle} \\ \cite{stagsted2020towards, qiu2020evolving}\end{tabular}
   &
   & 
  
  \\ \\
  
\begin{tabular}[c]{@{}l@{}} \textbf{Generative Adversarial Network (GAN)}\\ GAN discovers and learns regularities\\ or patterns in input data making\\ them apt for image and communication\\-based applications\end{tabular} &
   &
   & \begin{tabular}[c]{@{}l@{}}\checkmark \checkmark\\ e.g., \cite{becker2021generating,wang2018effective}\\ \cite{kerdegari2019smart}\end{tabular}
   &
   \\ \\

\begin{tabular}[c]{@{}l@{}}\textbf{K-means}\\ K means can cluster data\\ efficiently making them apt\\ for classification and\\ decision based applications\end{tabular} & & &
  &
  \begin{tabular}[c]{@{}l@{}}\checkmark \checkmark\\ e.g.,\cite{kmeans1, kmeans2}\\ \cite{kmeans5}\end{tabular} 
   \\ \\

\begin{tabular}[c]{@{}l@{}} \textbf{Linear Regression (LR)}\\ LR models obtains optimal values\\ of intercept and coefficients\\ making them suitable for\\ classification applications\end{tabular} & & &
 &
  \begin{tabular}[c]{@{}l@{}}\checkmark \checkmark\\  e.g., \cite{LR1, LR2}\\ \cite{LR5}\end{tabular} 
   \\ \\

\begin{tabular}[c]{@{}l@{}} \textbf{Gaussian Mixture Model (GMM)}\\ GMM models 2D complex, static obstacles,\\ making them apt trajectory planning\\ and control applications\end{tabular} & \begin{tabular}[c]{@{}l@{}}\checkmark\\ e.g., \cite{Yao_GMM_2017},  \cite{newaz2016fast}\end{tabular}
   &
   &
   & \begin{tabular}[c]{@{}l@{}}\checkmark \checkmark \checkmark\\e.g., \cite{Lin_GMM_2014,Yao_GMM_2017}\\ \cite{Mok_GMM_2017,Qiao_GMM_2015}\end{tabular} 
   \\ \\
 
\end{tabular}
\label{table5}
\end{table*}

\section{ML for UAV Trajectory Planning and Mission Scheduling}\label{sec: planning and scheduling}

\textcolor{black}{In recent years, ML techniques have been used to perfect the control of flight patterns to improve service quality. Trajectories are planned so that data is collected from the nodes maintaining the freshness of data. Supervised learning is utilized to minimize the training errors  for efficient trajectory planning. The prediction and classification capabilities of reinforcement learning strategies, such as Deep Q-Network (DQN) and Deep Deterministic Policy Gradient (DDPG), can be exploited in different environments (e.g., environments with discrete and  continuous action spaces). Reinforcement learning has been widely applied for trajectory planning and task scheduling of UAVs.} 

{\color{black}
\subsection{Supervised Learning-based UAV Communications}

The prediction of a UAV's trajectory is studied when UAVs are used to provide communication services in a smart city, e.g., in~\cite{xiao2019trajectory}. The accurate position information of the UAV is crucial in this application because it has a strong impact on the beamforming performed by the associated base station. The authors of~\cite{xiao2019trajectory} present an RNN-based arrival angle predictor for position prediction with a series of data processing procedures. Simulation results justify that the developed approach is able to learn and train the angle data and apply it to high-speed moving UAVs. 

\begin{figure*}[h!]
\centering
\includegraphics [width=6.5in]{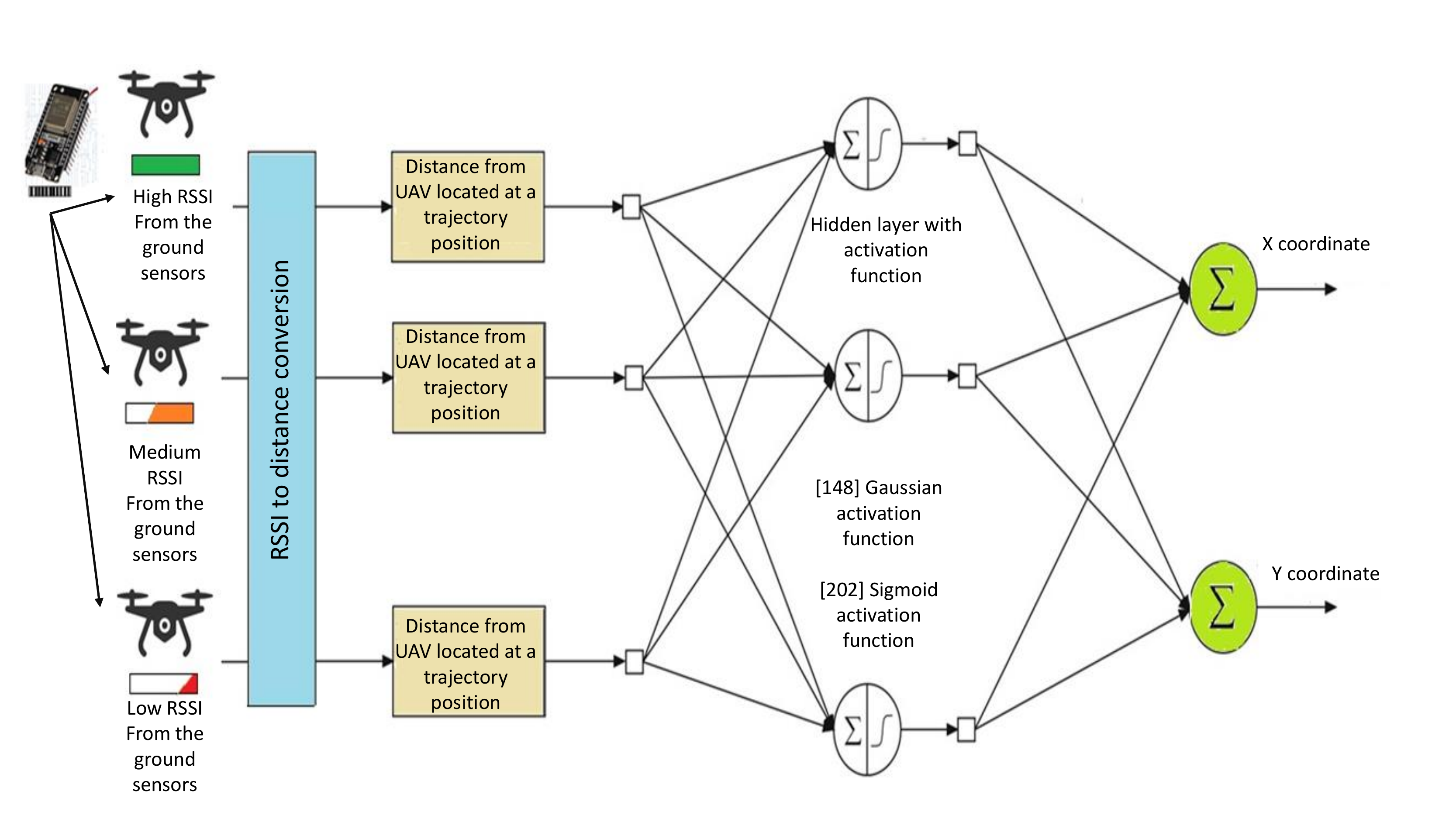}
\caption {The MLP for UAV-based localization of a WSN node using the Gaussian activation function \cite{annepu2020unmanned} and the RBF model using the Sigmoid activation function \cite{Annepu_MLP_Localization_2021}.}
\label{newmlp}
\end{figure*}

In \cite{annepu2020unmanned}, a UAV serves as a mobile aerial anchor node, which measures the received signal strengths from the ground sensors and locates the ground sensors. As compared to the deployment of terrestrial anchors, better localization accuracy is expected since line-of-sight (LOS) prevails between the ground sensors and UAVs. As shown in Fig.~\ref{newmlp}, an MLP model is created to estimate the locations of the sensors, which takes the RSSs as the input. The training of the MLP model is done by using backpropagation. The training data includes the positions of nodes randomly deployed in the given sensor field. By effectively capturing the non-linearity of the log-normal shadow fading, the nonlinear activation functions of the MLP model can improve the localization accuracy by up to 35\% over non-learning techniques. This technique is later extended by using radial basis functions in~\cite{Annepu_MLP_Localization_2021}.

\subsection{Unsupervised Learning for Trajectory Planning and Communications}

\begin{figure}[h!]
\centering
\includegraphics [width=3.5in]{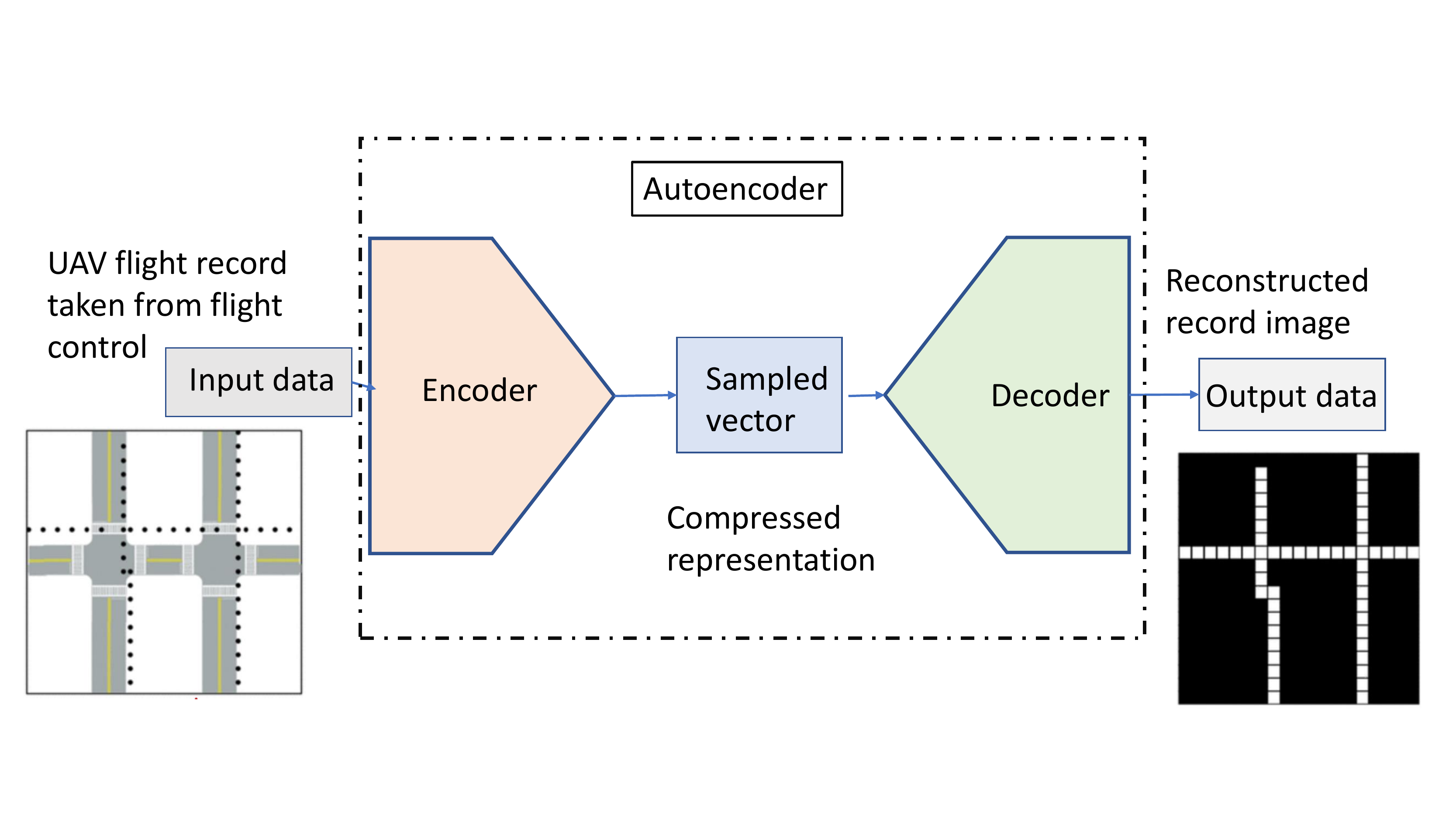}
\caption {Autoencoder used in trajectory planning by reconstructing the recorded images from flight control \cite{kwak2019autoencoder}}
\label{AEstruct}
\end{figure}

Unsupervised learning techniques, such as autoencoder, GAN, and GMM, have been applied to assist UAVs with trajectory planning and  communications.} Autoencoders can also be used in trajectory planning by generating the waypoints and suppressing such unintended flight records \cite{kwak2019autoencoder}. To do so, a three-step process is developed. First, the historical UAV's trajectories are utilized to generate a number of potential waypoints. Second, the images are generated based on the historical UAV's trajectories. Finally, those generated waypoints are determined as positions according to the repeating pixels of the images. By training the historical UAV's trajectories, the autoencoder accumulates and reconstructs the images. Moreover, the generated waypoints with the autoencoder are reduced by 84.21\% compared to those with the K-means, thereby improving the energy efficiency of the UAVs.

Autoencoders have also been used in movement prediction in dynamic environments \cite{sarkar2018sequential}. The autoencoder network is composed of a state and action-oriented decoder network, which is used to reconstruct the conditioned video according to the agent's actions. These predicted future frames can be used in trajectory planning in unknown terrain. The attitude of the flying UAV is commonly calculated by a data fusion algorithm combined with the data readings from the gyroscope, accelerometer and magnetometer, and adaptive Kalman filter. Attitude control is a control-based problem that ensures the smooth flight of a UAV. The authors of \cite{dai2018unsupervised} use deep autoencoders to fuse the features from the aforementioned sensors and define an optimal attitude estimation. {\color{black}In contrast to the classic autoencoders that have a single hidden layer, deep autoencoders will have multiple hidden layers depending on the neural network configuration. A deep autoencoder comprises of two symmetrical deep-belief networks that initially have four to five layers representing the encoding half of the net, and immediate hidden layers having the decoding half. These layers are based on restricted Boltzmann machines. Every hidden layer represents some form of fundamental features that are used in constructing the next layer of features. }

{\color{black}

For a UAV0-aided system, when analyzing a series of aerial images at various time points, there can be several issues, such as variations in camera pose, shadow, and illumination. Most of these issues are attributed to either noise or inadequate acquisition procedures. The authors of \cite{mesquita2019fully} use an autoencoder to cluster the features of the UAVs images in accordance with the reciprocal similarity. The features with more changes can be distinctively classified through training of the encoder network. The authors also confirm that an autoencoder can reduce the required training images.

\subsection{Semi-supervised Learning for UAV Trajectory Optimization}
A generative adversarial LSTM (GA-LSTM) network is developed to optimize the resource allocation in UAV-assisted machine-to-machine wireless communications in~\cite{xu2021generative}. 
The network joins the complementary strengths of GAN and LSTM for distributed optimization of the transmit power and mode, frequency channel, and the selection and trajectory of UAVs in a multi-agent environment with partial observability. LSTM is particularly selected to track and forecast the movement of the UAVs and facilitate reward evaluation under a partially observable situation. It is numerically demonstrated that GA-LSTM outperforms a direct use of LSTM or DQN in the sum rate. 


The authors of~\cite{Zhang_GLOBECOM_2018} consider a cellular network, where there are multiple UAV-based aerial BSs, ground BSs, as well as many ground terminals served by the UAVs and BSs. By using a weighted expectation-maximization algorithm, a GMM models the spatial distribution of radio traffic to assist with the deployment of BSs, including UAV-BSs. Traffic congestion is predicted accordingly and the optimal placement of the UAVs is derived to minimize the energy consumption of the UAVs on communication and relocation in case of  traffic distribution changes. Simulations show that the use of GMM helps save the energy of the UAVs by 20\% and 80\% on communication and relocation, respectively, as compared to a few heuristic-based alternatives.
}

\subsection{Reinforcement Learning for Joint Trajectory Planning and Mission Scheduling }
\textcolor{black}{DRL techniques can provide novel solutions for UAV trajectory planning in a dynamic environment.}


\subsubsection{Deep Q-Network with Trajectory Discretization}
Given partially observable network states in UAV-assisted communications and networking, reinforcement learning, e.g., Q-learning, can optimize UAVs' actions. However, due to curse-of-dimensionality, Q-learning is unable to be applied in the learning problems which contain a large state or action space~\cite{emami2021deepQ}. To circumvent the dimensionality issue of the learning problems, DQN is investigated to leverage neural networks to train the actions with an extended state and action spaces.

\begin{figure}[h!]
\includegraphics [width=3.3in]{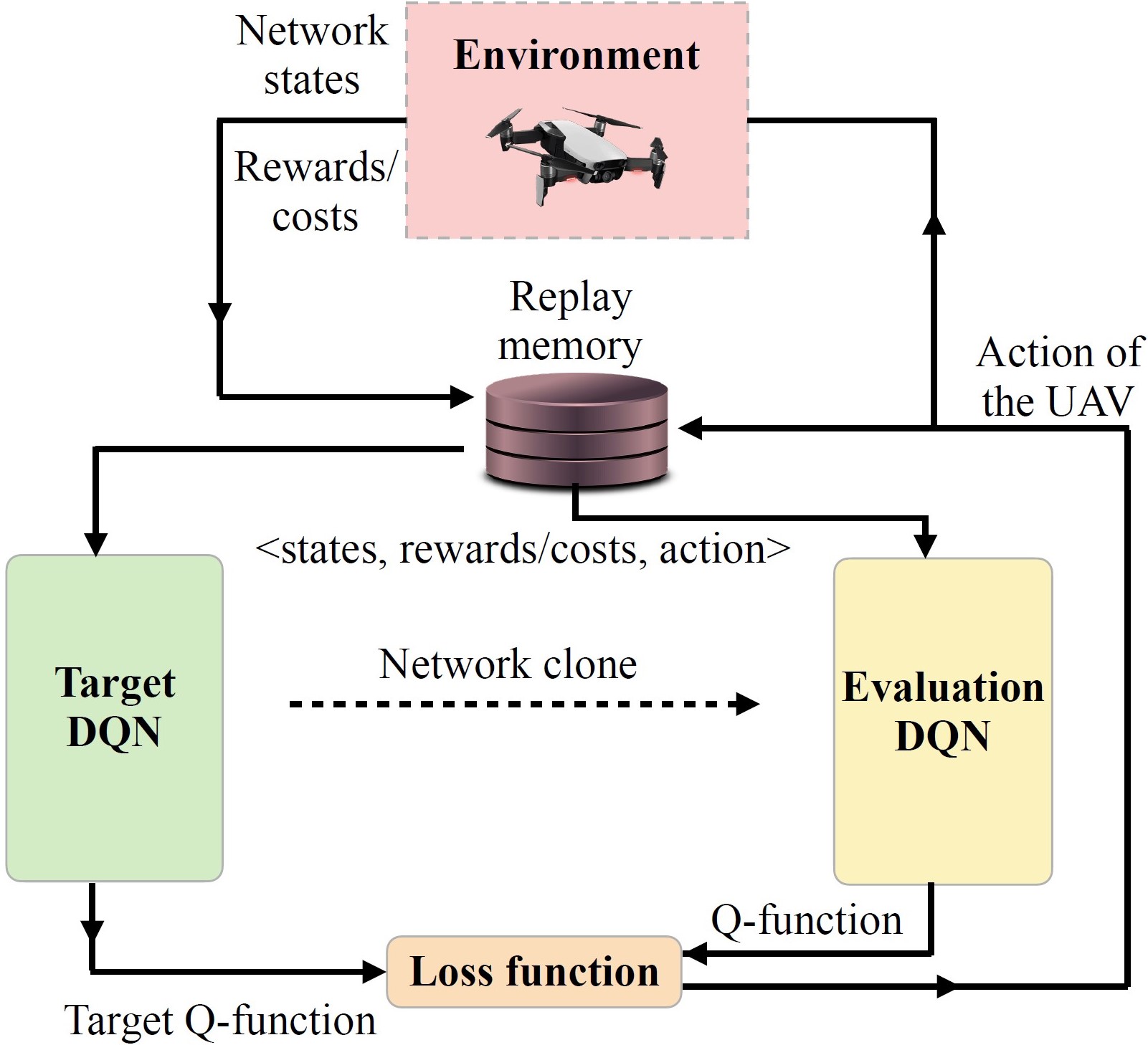}
\caption {Schematic illustration of the DQN model for trajectory and mission planning. The DQN model is designed to be trained onboard at the UAV to deliver the optimal policy of trajectory planning and radio resource allocation.}
\label{dqn1}
\end{figure}

\textbf{Age of information: }
DQN can be used to improve the energy efficiency of the trajectory planning while ensuring the data freshness, i.e., age-of-information (AoI), of ground nodes~\cite{abedin2020data}. 
In~\cite{ferdowsi2021neural}, a DRL algorithm using DQN is studied to obtain the optimal policy of trajectory planning and transmission scheduling to minimize AoI at the UAV. Since a large-scale network can result in an extremely high state space dimension, DQN requires a large replay memory. The authors of~\cite{ferdowsi2021neural} developed autoencoders with LSTM to capture spatio-temporal inter-dependencies between updated locations of the UAV and time instants. To enable efficient training in the large-scale network, the LSTM-based autoencoders extract features of the state space, which converts the states to a fixed-size vector. 

DQN~\cite{abedin2020data}, \cite{wang2020priority}, \cite{li2020online} can also be used to optimize the trajectory of the UAV and the bandwidth allocation of ground nodes, to maximize network throughput or minimize energy consumption. 
Wang~$et~al.$~\cite{wang2020priority} studied latency-prioritized trajectory planning in a time-sensitive UAV-enabled IoT network. DQN was adopted to optimize the cruise control of the UAV to improve the QoS. In~\cite{luo2020intelligent}, DQN-based trajectory planning considers the location of the UAV, the activation-sleep state of ground nodes, and the amount of transmitted data as network states. The DQN determines the heading direction of the UAV and the bandwidth allocation to maximize the data collection rate. 

\textbf{Packet loss: }
Li $et~al.$~\cite{li2019board} investigated onboard DQNs at the UAV to minimize buffer overflow and transmission failure of ground nodes. The DQN optimally controls the velocity of the UAV and communication schedule while learning the battery energy level, queue backlogs, and channel conditions of the ground nodes. In~\cite{li2020joint} and \cite{li2020online}, the trajectory planning and communication schedule are jointly optimized to reduce the data loss. The DQN is extended to optimize the discrete waypoints along the trajectory and select nodes for transmission. 

\textbf{Coverage: }
In~\cite{koushik2019deep}, the DQN is applied to control the UAVs' flight to ensure wireless connectivity and adequate coverage of a ground network. The DQN learns the network topology changes, and the UAV's trajectory is determined so that the network throughput and link condition can be guaranteed. 
DQN can also be trained at the UAVs to design their trajectories to cover the ground nodes fully ~\cite{liu2018energy}. The DQN learns the UAVs' coverage fairness and locations to minimize the UAVs' energy consumption while maintaining the network connectivity. 

\textbf{Energy efficiency: }
Since the ground nodes have limited battery energy, the flight trajectory is planned to maximize the uplink throughput during the flying time~\cite{zhang2021trajectory}. A DQN is studied with safe cruising policies to collect ground data while avoiding obstacles. 
A dueling DQN model is adopted to optimize the flight trajectory~\cite{liu2019green}, where the data packet is prioritized based on transmission latency constraints. The DQN is studied to learn the channel state and the priority of the data packet for minimizing the energy consumption of the ground nodes. UAVs can also be used for secure video streaming of ground nodes. A DQN ~\cite{zhang2020energy} is developed with safety policies to maximize the UAVs' energy utilization while ensuring the video quality.

\textbf{UAV-assisted MEC: }
Energy efficiency and security in UAV-enabled MEC networks have attracted attention. To reduce the computation burden, ground nodes can offload their computational tasks to edge devices. The task offloading optimization is studied in~\cite{zhao2021green} under attacks from the UAV eavesdropper. The network cost is formulated as a combined function of latency, energy, and price. A DQN-based resource allocation algorithm is studied to learn task offloading decisions to reduce the cost while ensuring communication security.
Double DQN is a variant of DQN, which can suppress the approximation errors in Q-learning and avoid overestimated rewards and the biased estimation of network state dynamics. As a result, double DQN can stabilize the learning process with fast convergence. In~\cite{wu2020deep}, cache-enabled UAVs are employed with MEC to assist content placement of the ground nodes. Given the limited battery power of the UAV, a double DQN model is developed for the UAV to maximize the network throughput. Since DQN may overestimate the action-value function, double DQN is utilized in~\cite{liu2020path} to maximize the long-term network throughput of MEC with the consideration of energy consumption of the UAV and QoS requirements.  UAVs can also be used to provide vehicular content caching in MEC-enabled autonomous driving~\cite{shi2020novel}. In this model, the UAV learns various vehicular content and available caching space to enhance the content response performance. 

\textbf{Others: }
In~\cite{jeong2017design}, UAV-assisted wireless power transfer (WPT) is studied with the DQN to design the flight trajectories and improve the energy harvesting efficiency. The WPT in~\cite{jeong2017design} can estimate the UAV's location, where Naive Bayes algorithms are used to train the flight data. The WPT efficiency can be enhanced by predicting the movement of the UAV. 
Li~\textit{et~al.}~\cite{li2020deepicc,li2020onboard} develop a DQN and a double DQN to optimally choose the ground node for data collection and WPT, as well as the optimal modulation of the selected ground nodes.
In~\cite{li2020deepiwcmc}, the DQN-based trajectory planning is further studied for UAVs, where the DQN determines the optimal position of the UAV to minimize buffer overflow of the ground nodes with sufficient harvested energy. 
In~\cite{kawamoto2018efficient}, a UAV-based network is developed based on Long-Term Evolution (LTE) sidelink physical channels. Q-learning is applied to schedule the UAV's transmissions and modulation allocation of the UAV.

\textit{Despite DQN can address many high-dimensional learning problems in UAV-assisted communications and networking, the action space of a DQN has to be discrete. In contrast, the action space is continuous for some online control problems of interest, especially the cruise control of the UAV.} 



\subsubsection{Online Trajectory Planning With Deep Deterministic Policy Gradient}
\textcolor{black}{DDPG integrates the value iteration and the policy iteration, which enables deep reinforcement learning with continuous state and action spaces~\cite{kurunathan2021deep}. The primary difference between DDPG and DQN is that the DQN predicts the Q values for each state-action pair. DDPG utilizes a critic network to determine the Q value,  and at the same time it utilizes an actor network to obtain the action~\cite{wang2019continuous}. 
}

\textbf{Cruise control: }
DDPG can be investigated to learn cruise control, e.g., headings and velocities of the UAV, to minimize network cost in continuous state and action spaces~\cite{li2021continuous,li2020onboardDDPG}. DDPG can conduct experience replay at the UAV to save the learning experience, which stabilizes its training. 
In~\cite{rodriguez2019deep}, DDPG is used to address the UAV autonomous landing on a moving platform, where the UAV learns the relative position to the ground platform and velocity difference. 
Moreover, a DDPG model is presented in~\cite{yang2019uav} to control the heading and velocity of the UAV under air combat situations. DDPG can continuously learn the air combat cruising policy by considering attacking zones and combat assessments while improving the maneuver decision. 
The authors of~\cite{bouhamed2020ddpg} present a DDPG model, which trains the navigation of the UAV to bypass obstacles in urban areas. Their DDPG-based trajectory planning maximizes the navigation reward to balance obstacle avoidance, the flight time to the destination, and the battery level of the UAV. 
DDPG can also be used to design the UAV's 3D movement to reduce the energy consumption and enhance the throughput fairness of the ground nodes since a battery-powered UAV has limited flight time~\cite{ding20203d}. 

\textbf{Age of information: }
In~\cite{samir2020age}, the trajectories of UAVs are designed to collect vehicular data while ensuring a minimized AoI to keep the information fresh. DDPG is used to learn time-varying traffic and road conditions, e.g., the number of ground vehicles, the instantaneous position of ground vehicles, and the AoI of ground vehicles. 
Based on the learning outcome of the DDPG, the AoI can be minimized by conducting the designed trajectories and scheduling policy. 
Sun $et~al.$~\cite{sun2021aoi} studied a twin delayed DDPG (TD3) model to minimize the AoI and energy consumption of the UAV-assisted IoT network. Their TD3 model learns AoI of all the ground IoT nodes and locations of neighboring UAVs while controlling the speed and trajectory of the UAV, as well as the bandwidth allocation. 
A multi-agent DRL is studied based on DDPG to minimize AoI by learning flight trajectories~\cite{wu2021uav}, where each of the UAVs decides either to transmit to the base station via cellular links or to the neighboring ground nodes. 
For minimizing the AoI, the authors of~\cite{samir2020online} presented a policy-based DRL algorithm to determine the flight altitude and the data transmission scheduling of the ground nodes. 

\textbf{UAV-assisted MEC: }
\textcolor{black}{UAVs can be employed to cache popular contents to release the pressure on wireless backhaul links while reducing content delivery delay~\cite{wang2020deep}. Since trajectory planning and resource allocation result in a large action space, the authors develop DDPG to optimize UAV's caching placement, schedule content delivery and specify the transmit power of the ground nodes. Given the predetermined flight trajectories, DDPG is used to optimize spectrum and computing resources in UAV-assisted MEC vehicular networks~\cite{peng2020ddpg}. The DDPG-based resource management is studied to enhance the number of offloaded tasks at the UAV while satisfying the required delay and QoS.}

\textbf{Others: }
A DRL-based trajectory control framework is studied to improve UAVs' coverage and resource allocation fairness while reducing energy consumption~\cite{liu2019distributed}. The critic neural network of the UAV is trained by environment state information, while the actor neural network applies observations at the UAV to determine its actions. 
{\color{black}
In~\cite{gao2021multi}, the allocation of UAVs to ground nodes is modeled as a potential game, where DDPG is used to optimize the trajectory of the UAVs for energy efficiency and obstacle avoidance. 
In~\cite{LinTCCN2020}, DRL and long short-term memory (LSTM) are integrated to derive the optimal strategy for individual UAVs in a formation flight to access the shared communication spectrum and achieve dynamic time slot allocation. The performance of the approach is gauged in terms of convergence speed and throughput.
In~\cite{GaoTGCN2021}, DRL, or more specifically, a dueling double deep Q-network, is applied to optimize the trajectory of a UAV subject to the initial energy, flight duration, initial and final positions of the UAV, so that the UAV experiences the least outage (i.e., being disconnected from the cellular networks) during the flight.
The dueling double deep Q-network is also applied to the simultaneous navigation and radio mapping of a UAV~\cite{ZengTWC2021}.}

\begin{figure}[h!]
\includegraphics [width=3.4in]{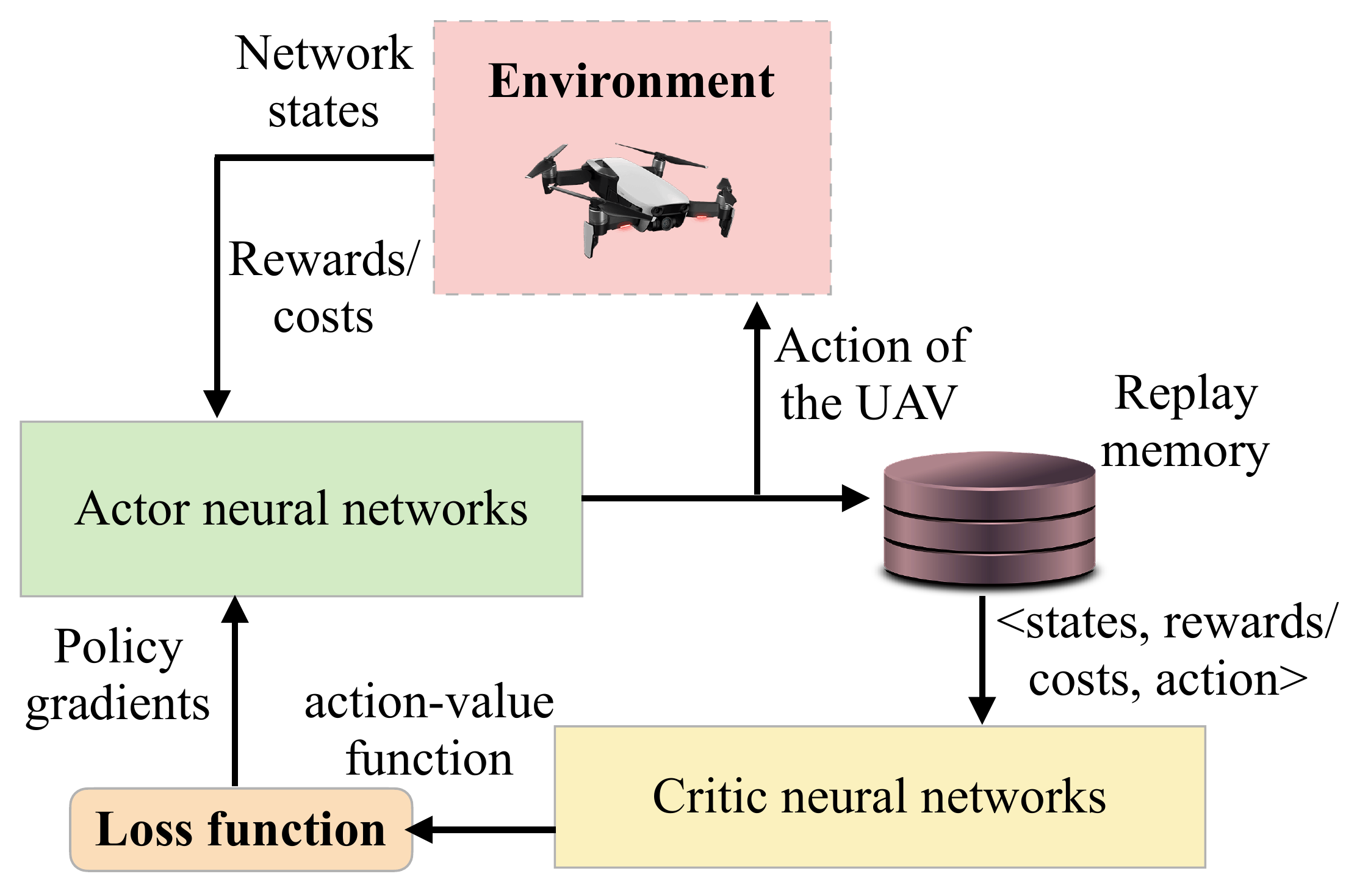}
\caption {An illustration of the DDPG architecture, where the actor and critic networks with an experience replay are designed to obtain the actions of the UAV.}
\label{ddpg1}
\end{figure}

\subsubsection{Multi-agent DRL for Multi-UAV Cooperation}
Multi-agent DQN is developed with multiple UAVs in~\cite{emami2021deepQ}, where the network state contains battery and data queue statuses of the ground nodes, as well as the waypoints of all the UAVs. The multi-agent DQN schedules the ground nodes' transmission while learning the data and energy arrivals. 
The authors of~\cite{jointYousef2021} extend their multi-agent DQN in~\cite{emami2021deepQ} to adjust the velocities of the UAV at waypoints while selecting the ground nodes for data transmissions. 
The authors of \cite{wu2020cellular} investigate the trajectory optimization of the UAVs with the communication scheduling of cellular networks. Due to the high complexity of the optimization, a multi-agent DQN is developed to optimize the UAVs' trajectories. The UAVs schedule the data transmission of the ground nodes of cellular towers according to the locations of the UAVs and the ground nodes. 
The authors of \cite{koulali2016green} focus on a non-cooperative game with periodic beaconing at the UAV to reduce network energy consumption. A multi-agent DRL algorithm is studied to determine the beaconing equilibrium durations with no observation of the other UAVs' transmission schedules. 
In~\cite{zhang2020uav} and \cite{zhang2020multi}, UAV jammers are deployed to improve channel secure capacity between ground nodes and legitimate UAVs. A multi-agent DDPG model is exploited to train the trajectory and jamming power of the UAV jammers and transmit power of the legitimate UAV. 

Wang $et~al.$~\cite{wang2020multi} present a multi-agent DDPG model to optimize the fairness of resource allocation in multi-UAV-enabled MEC, where the UAVs' trajectories and offloading decisions are trained to improve the energy efficiency of the MEC devices. 
\textcolor{black}{The authors of~\cite{peng2020multi} present a multi-agent DDPG-based resource allocation framework for MEC-based vehicular networks with UAVs. The MEC server is regarded as an agent, which trains the scheduling of the UAVs and the ground vehicles, as well as it performs resource allocation for vehicular computation. Using a federated learning framework~\cite{ShuyanCOMST2021}, the multi-agent DDPG-based resource management aims to enhance the number of the offloaded tasks from the ground vehicles to the UAVs.} 

\begin{figure}[h!]
\centering
\includegraphics [width=3.4 in]{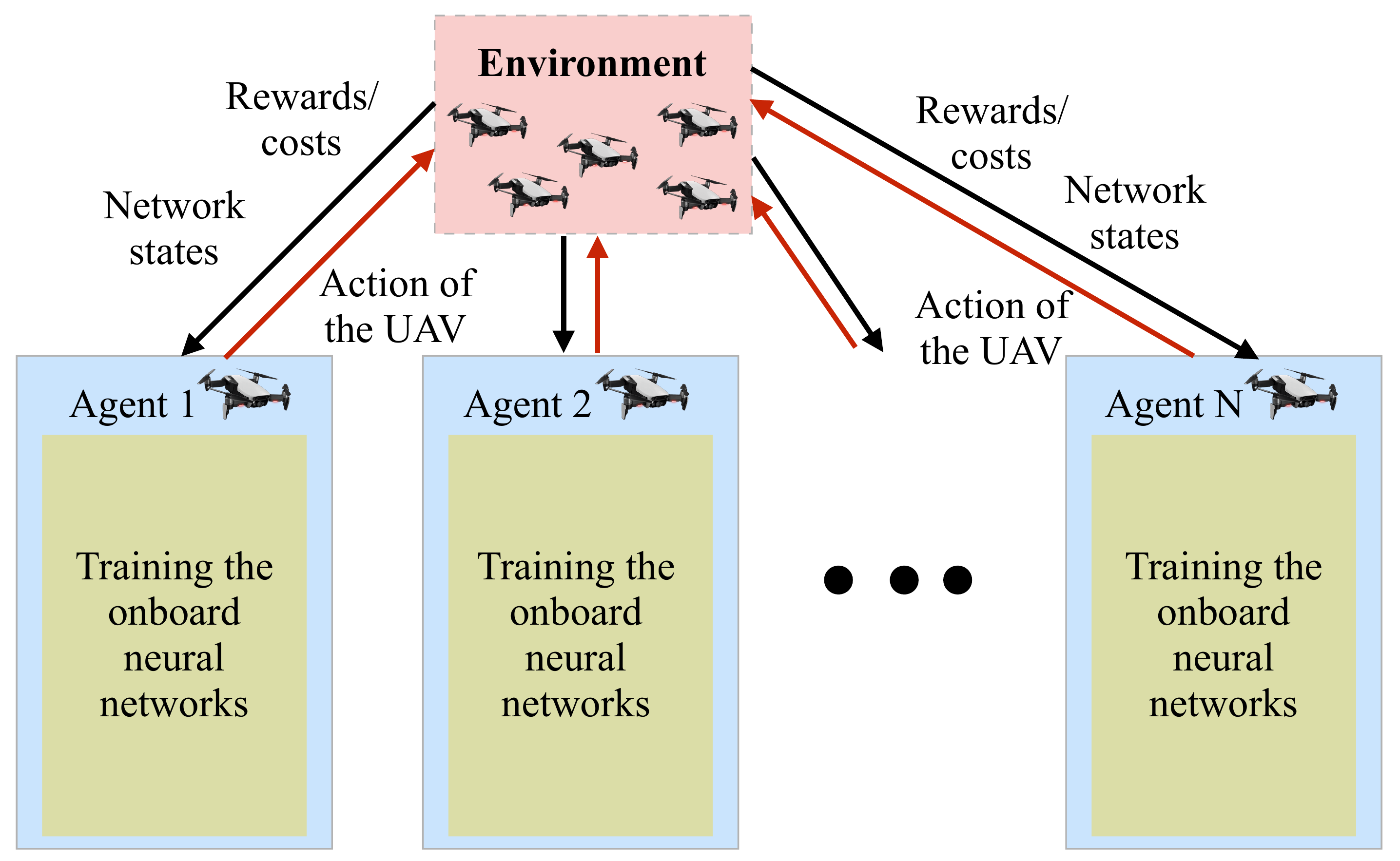}
\caption {A structure of multi-agent DRL. Each of the agents, i.e., a UAV, trains an onboard neural network to deliver the optimal joint actions.}
\label{MutiAgentDRL}
\end{figure}




 
  
   
  

\noindent \textbf{Remark:} Both DQN and DDPG have been utilized to support multi-agent DRL for online learning with experience replay in multi-UAV networks or UAV swarms. The choice of DQN or DDPG is determined by whether the problem space is discrete or continuous. DQN is  used for problems with discrete action space (e.g., clustering of UAVs), while DDPG is used for problems with continuous action space  (e.g., UAV trajectory planning)). Additionally, methods, such as autoencoding, have also been used for feature extraction and efficient supervised learning.



\section{ML for Aerodynamic Control and Operation of UAV}\label{sec: control and operation}
\textcolor{black}{ML has played a pivotal role in predicting the outcomes and taking appropriate decisions for UAVs based on different system parameters. Supervised learning techniques, such as CNN, RNN and MLP, have been utilized to facilitate path-finding and control of motion based on image and video input feed. Novel learning techniques, such as spiking neural network (SNN), have the capability to take control decisions for UAVs based on event-based input feeds. 
}

\subsection{Supervised Learning-based UAV Operations}

{\color{black}Supervised learning strategies, such as CNN and MLP, are predominantly used for image feature extraction and have the capability to train a machine based on video feeds. Recurrent neural networks (RNNs) possess a sequential solving structure, making them highly suitable for control decisions for a UAV. }

\subsubsection{Convolution Neural Network for Navigation}  
CNN can be used to develop autonomous navigation on the UAV. In \cite{bojarski2016end}, the UAV is equipped with a front-facing camera, where a CNN is trained using input images to control the steering or heading angles of the flight. To control the motion of the UAV, CNN has also been used for path finding, control, and maneuvering in an adaptive manner \cite{padhy2018deep}. The CNN in the aforementioned work uses the video feed from the front camera of the UAV and processes it through a deep neural network model to choose the next waypoints. 
A CNN model is studied in \cite{li2019intelligent}, where the situation data, maneuvering decision variables, and evaluation indices are used to learn intelligent maneuvering decisions.

With optimal UAV caching, a CNN-based deep supervised learning architecture is studied to make fast online flight control decisions. In \cite{9127423}, system parameters, such as network density and content request distribution with spatio-temporal dimensions, are labeled as images and used to train a CNN. A clustering-based two-layered algorithm is developed to provide online decisions based on the CNN model.

\subsubsection{Dynamics Tracking Recurrent Neural Network} 


 

In general, it is challenging to precisely control UAVs on the fly. The underlying reason is that getting an accurate mathematical model of UAVs is non-trivial since the fidelity is highly affected by various factors including but not limited to unmodeled dynamics, parametric uncertainties, and disturbances~\cite{gu2019survey}. RNN, a powerful data-driven method, has been used to model and control UAVs. 
An RNN is an ML technique that uses sequential data feeding~\cite{challita2019machine}. An RNN is used to address a time-driven problem of sequential input data~\cite{lei2020time}. The sequential problem-solving structure of RNN makes it more suitable for classification prediction problems. The input of an RNN consists of the current input (fresh data) and the previous data. A directed graph is formed based on the connections between nodes along a temporal sequence. The RNN has an internal memory where it stores the computation information from the previous samples to take future decisions. 
Long short-term memory (LSTM) is an important member of RNNs with feedback connections, which is designed for eradicating RNNs' vanishing gradient problem. It can process both single data points and sequences of data. As they can process several sequences of data, LSTM can make predictions based on time series data in spite of any dire time lags~\cite{wang2019data}. 

\begin{figure}[h!]
\centering
\includegraphics [width=3.5in]{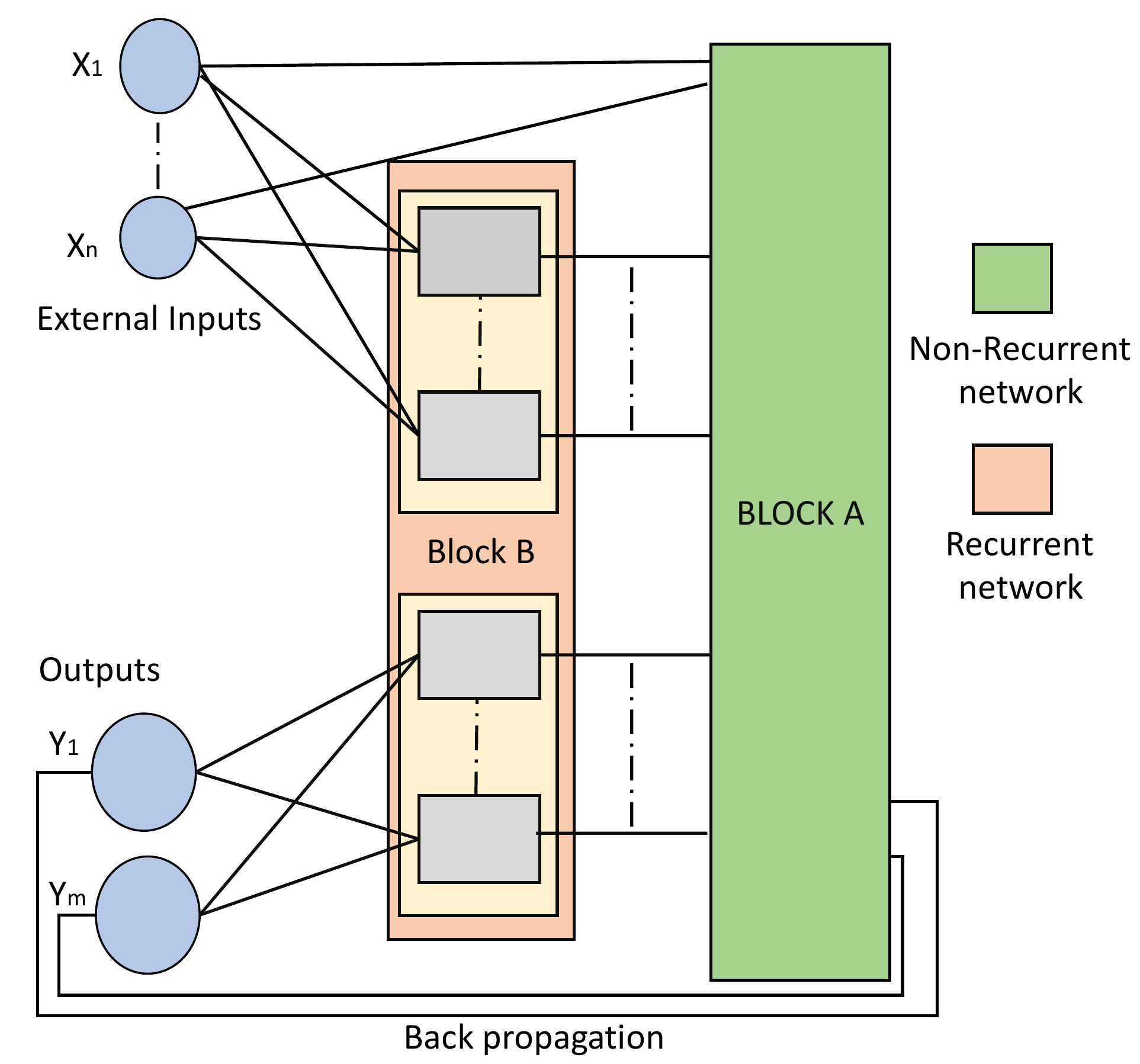}
\caption {A hybrid framework consisting of non-recurrent networks (Block A) from which the output is fed through backpropagation to the recurrent networks (Block B) to model the dynamics of a UAV like pitch, yaw, and roll \cite{san2006unmanned}.}
\label{FIG1_HH}
\end{figure}

\textit{Modeling the motion and dynamics of UAVs is critical in the control of UAVs. Emerging publications have reported RNN-based modeling methods.} \textcolor{black}{The paper~\cite{san2006unmanned} proposes a hybrid framework consisting of non-recurrent networks and recurrent networks to model the dynamics of a helicopter UAV; see Fig.~\ref{FIG1_HH}. Two sub-systems, each of which consists of a non-recurrent network (block A in Fig. \ref{FIG1_HH}) and a recurrent network (block B in Fig. \ref{FIG1_HH}), are connected in cascade to model the attitude (including roll, pitch and yaw angles) and the position of the UAV, respectively. The Elman contextual neurons are used in the recurrent network, and the MLP or the radial basis network is used in the non-recurrent network. In the figure, the external inputs are represented by the vector $[X_1,\cdots,X_n]$. These inputs generate contextual neurons. The outputs are represented by vector $[Y_1,\cdots,Y_m]$, which also generate contextual neurons. The contextual neurons define previous states to be memorized. These states will be constantly fed onto the recurrent neural networks. The knowledge from the memorized states aid in the decisions regarding modeling of the UAV dynamics.} 

Results show that MLP performs well in level flights, while the radial basis network performs well in take-off and landing. This leads to the method's main disadvantage, i.e., the need to have separate neural networks for different flight stages. 
The authors of \cite{mohajerin2014modular} propose a modular deep RNN framework to model the altitude dynamics of a quadrotor UAV. It shows that the capability of learning and modeling the high-order dynamics and non-linearity can be significantly improved by introducing feed-forward inter-layer connections in a multi-layer RNN, as these connections alleviate the vanishing/exploding gradient problem \cite{bengio1994learning}. Then, standard gradient descent-based training methods, such as the Levenberg-Marquardt (LM) algorithm can be used for training the model.

\begin{figure}[h!]
\includegraphics [width=3.7in]{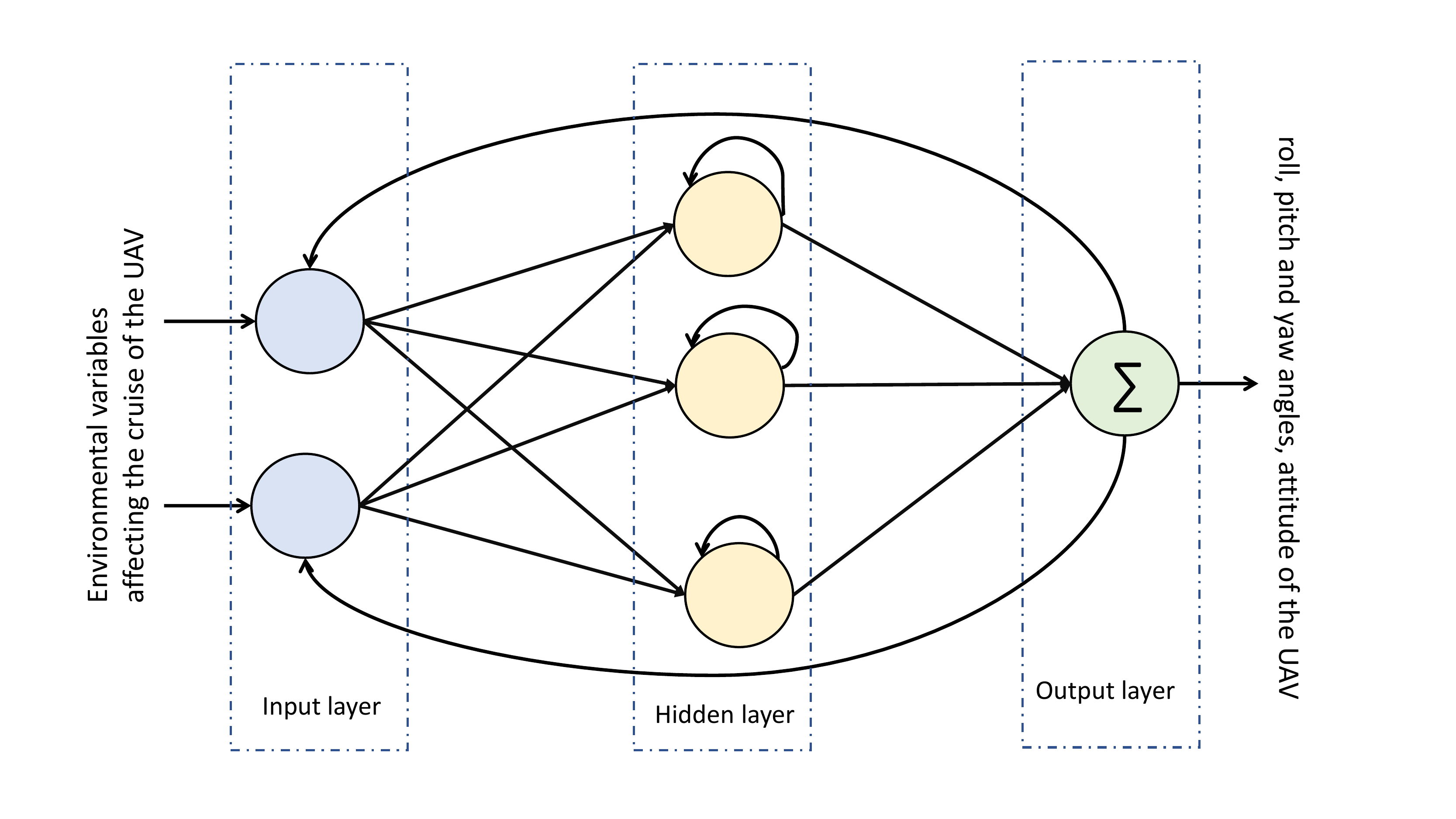}
\caption {The double loop RNN structure with two feedback loops by which both the output signal and the interior information can be stored making it suitable to estimate unknown dynamics~\cite{fei2017adaptive}.}
\label{FIG2_HH}
\end{figure}

\textit{RNN has also attracted increasing attention for the control of UAVs.} The authors of \cite{lin2014intelligent} propose a recurrent wavelet neural network (RWNN), comprising an input layer, a wavelet layer, a product layer, and an output layer, to mimic an ideal controller for trajectory tracking of a fixed-wing UAV. For online parameter training, a gradient descent method minimizing the sliding condition is chosen. Computer simulations were presented to demonstrate that favorable tracking performance can be achieved even with control effort deterioration and crosswind disturbance. The main limitation is that this study only tested the method on a linear motion model. The authors of \cite{fei2017adaptive} propose a double loop RNN structure for adaptive sliding mode control; see Fig. \ref{FIG2_HH}. Since this structure has two feedback loops, both the output signal and the interior information can be stored, making it capable of estimating unknown dynamics better. 

The authors of \cite{dadian2016recurrent} discuss the application of an echo state network (ESN), a class of RNN, to control a fixed-wing UAV. The ESN has been utilized in offline and online training. While the offline training achieves the inversion needed for the feedback linearization, the online training reduces the inversion error due to the modeling deficiencies. With the data collected from the FlightGear model, the authors show that the trained networks can significantly improve the open-loop and closed-loop responses in terms of roll rate and bank angle. As an extension, the controller's performance was evaluated in the terms of pitch and yaw angles. 

Similar work was presented in \cite{pugach2017nonlinear}, where the authors use the stochastic gradient descent (SGD) method for online learning. The authors of \cite{kelchtermans2017hard} study the control of a rotary-wing UAV with a forward-looking camera for safe flight in a cluttered indoor environment. The RNN is used to train an LSTM network for controlling the UAV. A window-wise truncated backpropagation through time (WW-TBPTT) sampling method is developed to address the highly correlated visual data. It shows that only retraining the fully connected layers achieves competitive performance with the end-to-end training. The main benefit is the reduction of the amount of training data and training time. Since the study is done in a simulator, real-world experiments can be considered for further validation. 

The authors of \cite{zhou2021control} study the usage of RNN for vertical take-off and landing (VTOL) of a UAV. The designed controller is composed of an outer-loop position controller and an inner-loop attitude controller. An RNN is used in the outer loop  to approximate a nonlinear solver since the latter suffers from high computational complexity. It is reported that the approximation errors of the proposed RNN are negligible. The RNN generates much smoother outputs than the nonlinear solver, and it is computationally efficient and can run in real-time (e.g., 50 Hz). The system robustness and trajectory tracking accuracy are verified in the presence of wind disturbance. 

The authors of \cite{hu2006application} apply an RNN to the control of a follower UAV in the tight formation flight. Regarding the pitch angle induced by the leader UAV as a seeking object, an annealing RNN is developed for extremum seeking to compute the minimum power demand of the wingman follower UAV. Computer-based simulations show that the developed approach solves the chatter problem observed in general algorithms for extremum seeking. The authors of~\cite{tsai2019vision} use RNN to process UAV images for collision avoidance. The images are firstly fused based on a deep CNN. Then, an RNN extracts image features for object tracking. These works are tested on experimentally collected datasets. Integrating the algorithms into hardware platforms is underway to evaluate their effectiveness in practice. In Table \ref{table6}, we tabulate some of the variants of RNN and the applications they support.

\begin{table}[]

\caption{The enhancements of RNN and the applications that they support }

\begin{tabular}{l|l|l}
Paper                      & Technique       & Application                   \\
\hline
\cite{san2006unmanned}      & Hybrid RNN       & \begin{tabular}[c]{@{}l@{}}Dynamics of a\\  helicopter UAV\end{tabular}                               \\
\\

\cite{mohajerin2014modular} & Deep RNN         & \begin{tabular}[c]{@{}l@{}}Altitude dynamics of a \\ quadrotor UAV\end{tabular}                       \\
\\

\cite{bengio1994learning}   & Multi-layer RNN  & \begin{tabular}[c]{@{}l@{}}Alleviate the vanishing/\\ exploding gradient problem\end{tabular}         \\
\\
\cite{lin2014intelligent}   & RWNN             & \begin{tabular}[c]{@{}l@{}}Mimicking an ideal controller\\  for trajectory tracking\end{tabular}      \\
\\
\cite{fei2017adaptive}     & Double loop RNN & Adaptive sliding mode control \\
\\
\cite{dadian2016recurrent} & ESN             & Control a fixed-wing UAV      \\
\\
\cite{kelchtermans2017hard} & RNN - LSTM       & \begin{tabular}[c]{@{}l@{}}Control of a rotary-wing UAV \\ with a forward-looking camera\end{tabular} \\
\\
\cite{zhou2021control}      & Outer loop - RNN & \begin{tabular}[c]{@{}l@{}}Vertical take-off and landing \\ (VTOL) UAV\end{tabular}                   \\
\\
\cite{hu2006application}    & Annealing RNN    & \begin{tabular}[c]{@{}l@{}}Control of a follower UAV \\ in the tight formation\end{tabular}          
\end{tabular}
\label{table6}
\end{table}

\subsubsection{Multilayer Perceptron for Aerodynamic Control}

Multilayer Perceptron (MLP) can be applied to the onboard flight control and management of UAVs, e.g., \cite{mansouri2017remaining,Gomez-Avila_PID_2019,Gomez-Avila_EKF_2020,Hernandez-Barragan_Tracking_2021}, and \cite{Bhandari2017NonlinearAC}. 
In \cite{mansouri2017remaining}, MLP is used to predict the remaining battery life of the UAV, where the lifetime prediction is formulated as a standard remaining useful life prognostic. The MLP is tested with a UAV prototype which is powered by four sets of 4.2 Volt lithium polymer batteries. The results show that the MLP-based flight control outperforms the linear models in terms of the battery life prediction. 
MLP can also be used to implement adaptive proportional-integrative-derivative (PID) controllers of the UAV, which provide continuously modulated control of the motions. The MLP configures the adaptive PID controllers, where one MLP per degree of freedom of the nonlinear dynamic control (e.g., motion and rotation). According to~\cite{Gomez-Avila_EKF_2020} and \cite{Gomez-Avila_PID_2019}, training the MLP with PID controllers can be modeled as an estimation of a nonlinear PID controlling, and can be solved by the Kalman filter. Specifically, the weights of the MLP are the states that the Kalman filter estimates. The output is the measurement used by the Kalman filter. The weights of the MLP are optimized to minimize the prediction error of the Kalman filter. 

\begin{figure}[h!]
\includegraphics [width=3.7in]{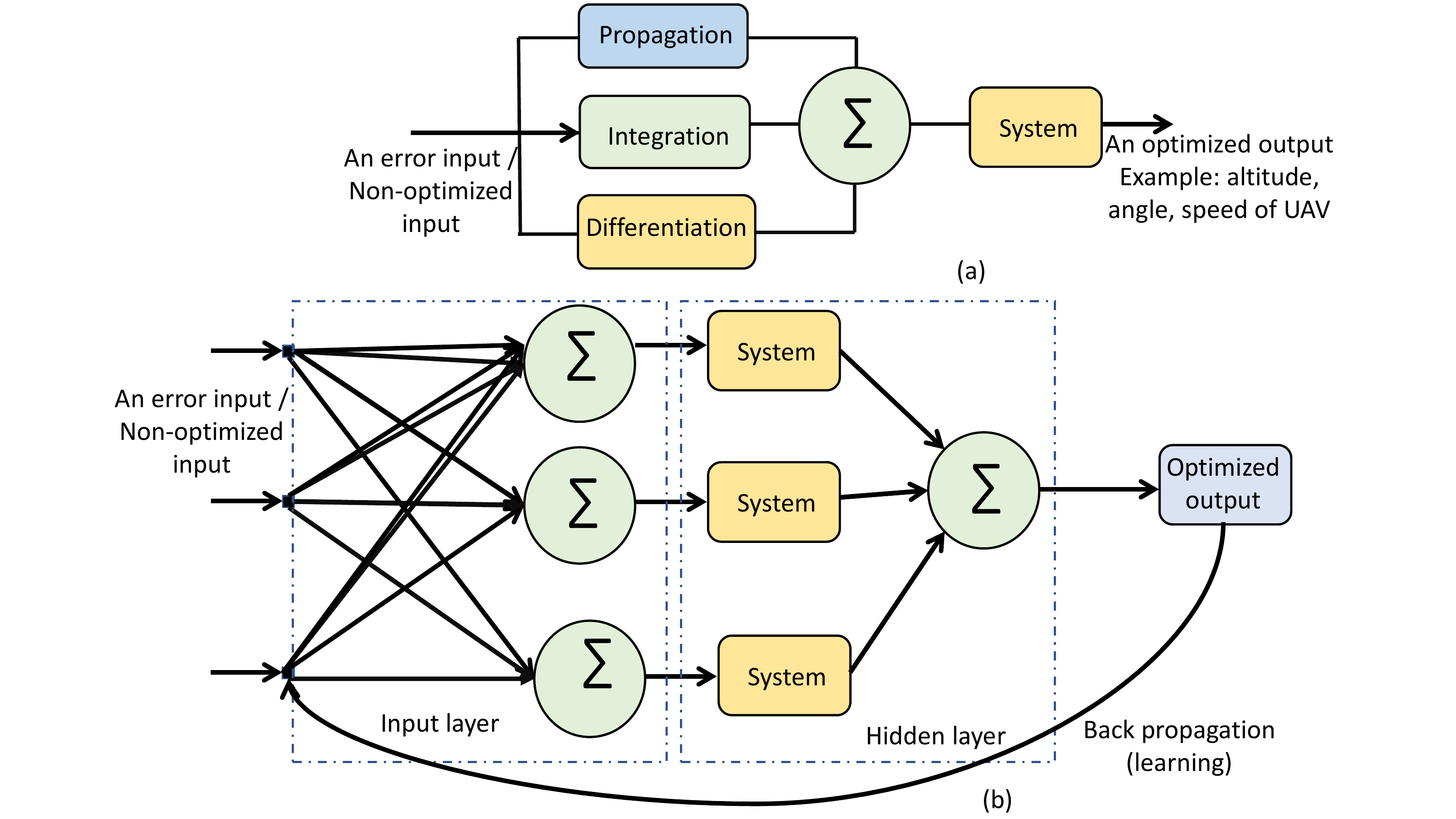}
\caption {Comparison between (a) conventional PID controller and (b) MLP-based adaptive PID controller - an MLP-based PID controller adjusts its gain adaptively an eliminates steady-state error and oscillations through training}
\label{Fig_MLP_PID_single_Neuron}
\end{figure}

Training the MLP with an extended Kalman filter~\cite{Hernandez-Barragan_Tracking_2021} is studied to track the UAV's trajectories. As shown in Fig.~\ref{Fig_MLP_PID_single_Neuron}, an MLP-based PID controller adjusts its gain adaptively, thereby suppressing the steady-state error and oscillations pertaining to the integration operation of the PID controller. Experiments are carried out using a KUKA Youbot mobile manipulator~\cite{KUKA_youbot_2011} to show that the neural networks with the extended Kalman filters lead to faster learning and convergence than the training based on backpropagation. 
The MLP is also used to implement a nonlinear adaptive controller for fixed-wing UAVs~\cite{Bhandari2017NonlinearAC}, where the networks can be trained online or offline. Synthetic data can be produced with an experimentally validated nonlinear flight dynamics model, e.g., FlightGear Flight Simulator~\cite{FlightGear_2017}, to train the MLP for reducing modeling errors, noise and disturbance.


\subsection{Unsupervised Learning-based Approaches}




{\color{black} Novel unsupervised learning techniques, such as spiking neural network (SNN), have energy-savvy and high processing capabilities, making them suitable to take faster and energy-efficient on-the-air control decisions and the dominating approaches to unsupervised learning-based UAV control and operations.}

{\color{black}Neuromorphic SNNs utilize the temporal difference learning for predicting both the rewards and the temporal sequence prediction in a physical time domain. Typically, temporal difference learning can be achieved by analyzing the temporal distance between neighboring events that can vary in a decay time constant. Neuromorphic SNNs replicate the functionalities of a central nervous system. The neuromorphic SNNs usually operate on orders of magnitude less power than traditional computing systems. This low-power capability is due to its  event-driven and massively parallel nature of operation, where typically only a small portion of the entire system is active at any given time while the other part is idle. This can aid in applications such as edge computing where there are strict energy constraints. 

To leverage the ultra-low-power of neuromorphic processors (in the order of several milliWatts), a neuromorphic SNN model is studied for onboard deployment at the UAV to control the UAV's movements for obstacle avoidance~\cite{salt2019parameter, salt2017obstacle}. Differential evolution and Bayesian optimization are used to obtain the optimal SNN configuration.}
In~\cite{stagsted2020towards}, an SNN-based proportional integral derivative (PID) controller is integrated with motor control of the UAV for ultra-low power consumption and high processing rate. An SNN-based control architecture is developed, where each spiking neuron carries sensor measurements and control information and fires a spike when they reach thresholds or biases. 

SNN is studied in~\cite{qiu2020evolving} to control a hexacopter UAV in six degrees of freedom, i.e., yaw, roll, pitch, height, position, and angular velocity. The researchers in this work propose a recurrent spiking controller that solves nonlinear control problems in continuous domains using a topology evolution algorithm as the learning mechanism. Their results suggest that the SNNs have the ability to solve ongoing control problems by maintaining sufficient spike activities and decoding from weighted spike frequencies.
In~\cite{kirkland2019uav}, an unsupervised spike time-dependent plasticity approach is developed, where SNNs are asynchronously trained to detect UAVs on the images. A new system is designed, which uses the features of an event-based camera to identify UAVs. An SNN is trained by using an unsupervised method of Spike Time Dependent Plasticity (STDP). The system is shown to be asynchronous and low in both power and computational overhead.

Zhao $et~al.$~\cite{zhao2018brain} study a decision-making model for UAV's flight control, where an SNN is used to simulate the function of brain zones. The SNN at the UAV determines the control actions to fly through a window or avoid obstacles according to their relative positions. 
The authors of~\cite{zhao2019lgmd} present lobula giant movement detectors to control the UAV indoor navigation for obstacle avoidance. By partitioning the image of the onboard camera, the spiking neurons are added to detect and locate obstacles in a reconstructed map, which is fed to the navigation model of the UAV. As one of the SNN models, a liquid state machine can track the network states over time while analyzing the behavioral information of the data to predict the data feature distribution.
In~\cite{chen2019liquid} and \cite{chen2017liquid}, liquid state machines are developed for the resource allocation of cache-enabled UAVs. The liquid state machines can learn the data request distribution of the ground nodes and determine the data caching policies for the UAV. 

\vspace{0.2cm}
\noindent \textbf{Remark:} Due to the capability to recurrently process data and constantly learn from the environment to take decisions, methods, such as RNN, rightly fit into the domain of controlling the UAVs. Novel techniques like SNN bring the advantages of being energy efficient into UAV control. The usage of ML has further extended into setting the waypoints of the UAV online and smart trajectory planning for applications like data collection and sensing.

\section{ML for UAV Perception and Feature Extraction}\label{sec: feature extraction}
{\color{black}Feature extraction is a form of dimensionality reduction. Feature extraction, pattern recognition, and image processing usually start from an actual set of measured data (taken through the camera of the UAV). It builds derived values (features such as edges, shapes, object recognition) that are informative. This derived learning is non-redundant and facilitates subsequent learning to obtain better feature extraction.} {\color{black}UAVs can provide an eagle-eyed view of the region of interest compared to their counterparts, i.e., the non-UAV imaging platforms. The mobility of UAVs can also provide the capability to cover a larger geographical area than their stationary counterparts~\cite{giordani2020non}. In what follows, we discuss important ML techniques used in UAV-assisted imaging.


\subsection{Supervised Learning-based UAV Perception and Feature Extraction}

Supervised ML strategies, such as MLP, can process information through multiple layers and help in the perception of images captured by the UAV. Methods, such as CNN, can segment and connect the layers of the image and aid in feature extraction.}

\subsubsection{Multilayer Perceptron for Image Processing}

%


UAVs have been increasingly utilized for precision agriculture, where MLP models demonstrate their applicability to the analysis of aerially captured images for crop disease and vegetation management~\cite{Abdulridha_tomato_2020,Kestur_tomato_2018}.  In~\cite{Abdulridha_tomato_2020}, a UAV equipped with hyperspectral cameras is used to take hyperspectral images of a tomato field for early diagnosis of spots resulting from fungus and bacteria. An MLP neural network is used as a classifier to analyze the hyperspectral images and detect and identify tomato diseases with an impressive accuracy of 99\%. In~\cite{Kestur_tomato_2018}, a quadcopter UAV equipped with a Raspberry Pi single-board computer with an onboard camera module is used for vegetation mapping of tomato crops. An MLP is used to segment the tomato crop images and demonstrated to provide better precision and recall performances than its alternative based on a support vector machine (SVM).

The capability of MLP on aerial image analysis has been applied to environmental management, e.g., weed eradication~\cite{tamouridou2017application} and flood management~\cite{rahnemoonfar2018flooded,Rahnemoonfar_Flood_2021}.
In \cite{tamouridou2017application}, a multi-spectral camera (green-red-near infrared) is installed on an eBee fixed-wing UAV to acquire high-resolution images. The UAV is remotely controlled and lifted to the altitude to acquire complete imaging coverage of the interesting field. An MLP with automatic relevance detection (MLP-ARD) is applied to detect a particular weed type, Silybum marianum, among other vegetation. A feed-forward MLP neural network with one hidden layer and one output unit is regulated by Bayesian regularization to avoid over-fitting, trained based on spectral and textural input data, and classifies the weed. 
In~\cite{rahnemoonfar2018flooded}, a densely connected CNN and an RNN are used to perform semantic analysis of the aerial images of flooded areas collected by UAVs in Houston, Texas. An MLP is used for each class at the output of the RNN. The CNN and the RNN are separately trained using Adam and Adagrad with the learning rates of 0.00001 and 0.01 and the batch sizes of 12 and 8, respectively. An accuracy of 96\% is achieved in terms of detecting flooded areas. The technique is recently applied to post-flood scene understanding~\cite{Rahnemoonfar_Flood_2021}.

MLP models have been used to assist UAVs with route planning~\cite{Gunchenko_harvesting_2017,annepu2020unmanned}. 
A UAV is utilized to assist  route planning and harvest volume measurement for unmanned agricultural harvesting equipment in~\cite{Gunchenko_harvesting_2017}. This is motivated by the fact that some branches of US agriculture lose 30\% of their harvest due to the inability to harvest on time. The UAV carries multi-spectral cameras. Different neural networks are tested to analyze the multi-spectral images, estimate harvest volume, and identify various obstacles in the field. Considering a linear network with three neurons in the input layer, the authors of~\cite{Gunchenko_harvesting_2017} test the MLP with three neurons in the hidden layer, generalized regression network with thousands of neurons in the hidden layer, or radial-basic function with hundreds of neurons in the hidden layers. The results indicate the radial-basic function with 154 neurons provides the best accuracy in testing data.

\subsubsection{Convolution Neural Network for Image Processing}
%

\begin{figure}[h!]
\centering
\includegraphics [width=4 in]{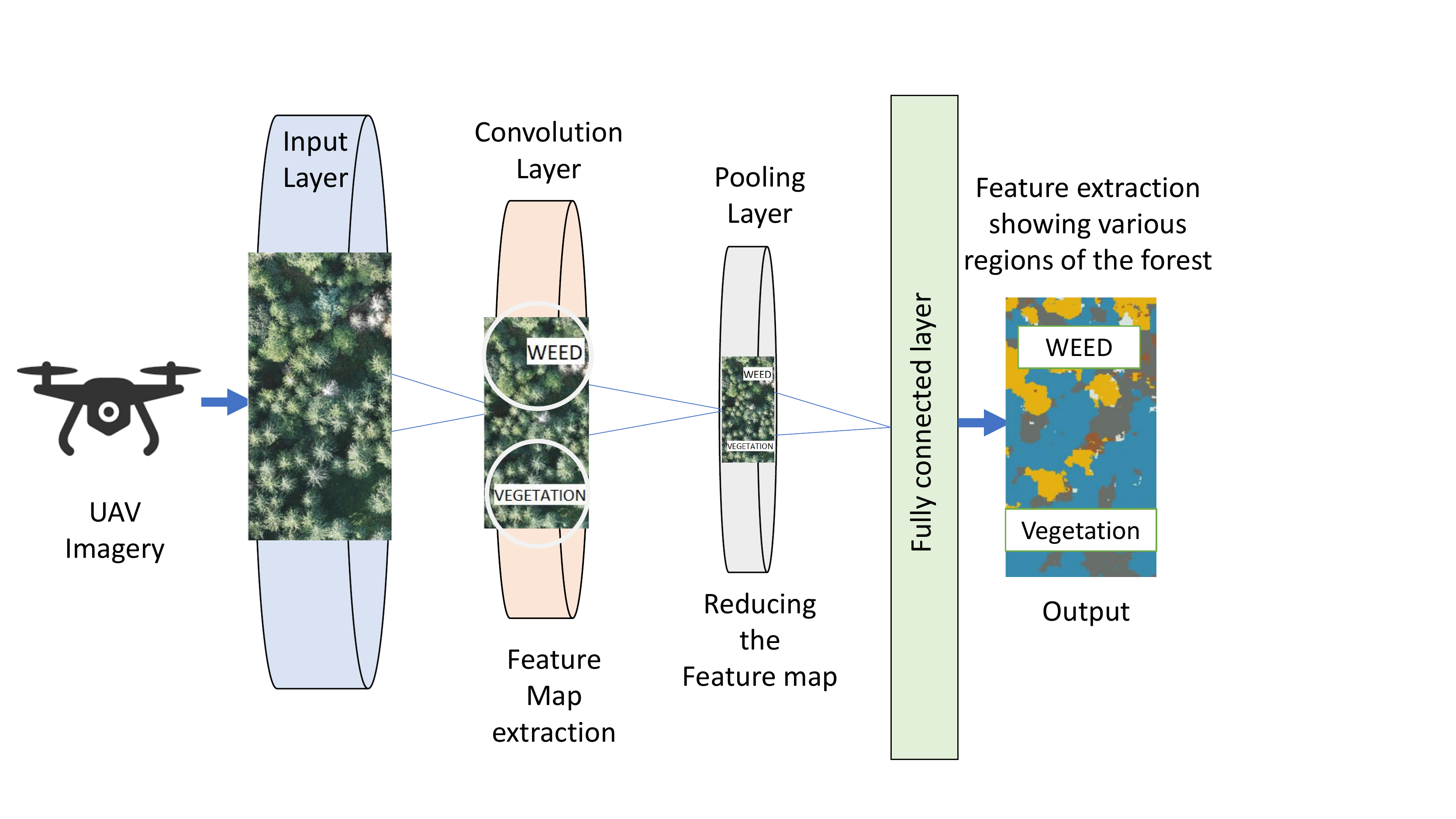}
\caption {Layout of CNN performing feature extraction and classification of drought areas and vegetation using UAV-captured forest imagery.}
\label{cnn1}
\end{figure}



CNN is commonly used to classify, and segment remotely sensed imagery due to its prowess in in-depth extraction features. In Figure \ref{cnn1}, we present an example of UAV forest imagery where CNN is employed to extract various features of the forest such as vegetation and dry areas. CNN has been used in several multi-object tracking methods \cite{bewley2016simple} for online and real-time applications to effectively associate objects. Ill-conditioned radio connections between the UAV and the base can degrade the resolution and precision of videos or images sent to the base, giving rise to difficulties in image analysis. These adverse environments can result in packet loss and wastage of bandwidth. The authors of \cite{yongqiang2020optimal} propose an Optimal Strategy Library (OSL) for video encoding, which can adapt to the packet loss rate and bandwidth of the radio connections between the UAV and the base. This method facilitates encoding video sequences and recovering partially corrupted videos. 

The authors of \cite{amorim2019semi} use CNN-based approaches for slope failure detection from UAV remote sensing imagery. The precision and accuracy assessment of the CNN approach in their experiment leveled to almost 90\% in the imagery of a moving terrain. Similar precision and accuracy assessments of over 90\% are achieved using CNN in several image classification-oriented applications \cite{ghorbanzadeh2019uav}. UAV-based high-throughput phenotyping using high-resolution multi-spectral imaging is enabled using CNN-based techniques. CNN with sufficient training is used in this application for classification and segmentation. Alongside a steady throughput, CNN is able to provide almost an accuracy of 99\%.
The authors of \cite{bejiga2017convolutional} apply UAVs with video cameras installed to carry out search and rescue avalanche survivors. The pictures presenting avalanche debris captured by the UAV are analyzed using a trained CNN to detect useful features and signs of survivors. A linear Support Vector Machine (SVM) is trained and concatenated to the CNN to help the object detection. CNN extracts the data from the image for prediction. Due to its efficiency in terms of accuracy and precision, CNN is used in extracting the features of the image regardless of spatial resolution, and spectral bands \cite{zhang2020transferability}.

CNN is predominantly used in imaging and related applications, due to its prowess in computer vision-based tasks, such as localization \cite{sermanet2013overfeat}, object detection \cite{amorim2019semi,xu2017car}, and image segmentation \cite{attari2017nazr}. Some application domains \cite{bejiga2017convolutional} use CNN to classify the UAV images to assist rescue operations.
One of the significant drawbacks of CNN is that the process of segmentation is very detailed, and it is time-, energy- and resource-consuming. Several techniques, such as recurrent CNN (R-CNN)  \cite{girshick2014rich} were developed to overcome the process of exhaustive processing. In \cite{zhang2020transferability}, UAV imagery applications using R-CNN are able to obtain better accuracy in detection with more acceptable image resolutions.
%
%
 The authors of \cite{kyrkou2018dronet} propose a lightweight CNN architecture that runs efﬁciently on embedded processors. The aforementioned lightweight network accelerates the execution of the model without any dire trade-off on the overall accuracy. An energy-aware design for Vision-Based Autonomous Tracking and Landing of a UAV was proposed by \cite{zamanakos2020energy}. They use a marker detection algorithm that runs with marginal energy overhead, simultaneously adapting the QoS level of CNN results for a considerable power saving. In Table \ref{table7}, we tabulate some of the variants of CNN and the applications they support.

\begin{table}[]
\caption{Applications realized though CNN-aided UAVs}
\begin{tabular}{l|l|l}
Paper                           & Technique used & Application              \\
\hline
\cite{bewley2016simple}        & Faster-region CNN & Multi-object tracking    \\
\\
\cite{amorim2019semi}          & Unlabeled CNN & Slope failure detection  \\
\\
\cite{ghorbanzadeh2019uav}  & Multi spectral CNN   & Multi-spectral imaging   \\
\\
\cite{bejiga2017convolutional} & SVM-CNN  & \begin{tabular}[c]{@{}l@{}}Avalanche search and \\ rescue applications\end{tabular} \\
\\
\cite{zhang2020transferability} & SVM-CNN & Image feature extraction \\
\\
\cite{sermanet2013overfeat}     & OverFeat & Image localization       \\
\\
\cite{attari2017nazr}           & Nazr-CNN & Image segmentation       \\
\begin{tabular}[c]{@{}l@{}}\cite{amorim2019semi}\\ \\
\\
\cite{xu2017car}\end{tabular}  & R-CNN & \begin{tabular}[c]{@{}l@{}}Image detection and \\ localization\end{tabular}         \\
\\
\cite{kyrkou2018dronet}       & Lightweight CNN  & Autonomous tracking     
\end{tabular}
\label{table7}
\end{table}

\textit{CNN does not encode the position and orientation of an object. CNN can sometimes be time-consuming as the classification and segmentation are performed in detail. The layers that are closer to the CNN input help in classifying simple features, such as edges, corners, endpoints, etc. When CNN has more layers, the training process takes longer. This drawback can be alleviated by the usage of several lightweight CNN models that do not demand more potent GPUs for computing.}

\textit{Recurrent Neural Networks (RNN) have been considered to enhance the CNN in the processing of images taken by UAVs.} In \cite{rahnemoonfar2018flooded}, an integration of densely connected CNN and RNN networks is proposed. The dense connection helps improve the information flow and gradients across the network, which further helps in the training process of deeper networks while still reducing over-fitting issues. An accuracy of 96\% is reported on a real-world dataset. 

\subsection{Unsupervised Learning for UAV Feature Extraction}

{\color{black}Similar to image feature extraction, unsupervised ML strategies can also enable radio feature extraction and can be used to extract features, such as received signal strength and channel strength. GAN with the ability to discriminate local datasets has been heavily featured for radio feature extraction.}

\subsubsection{Generative Adversarial Network for Image Extraction}


\begin{figure}[h!]
\includegraphics [width=3.3in]{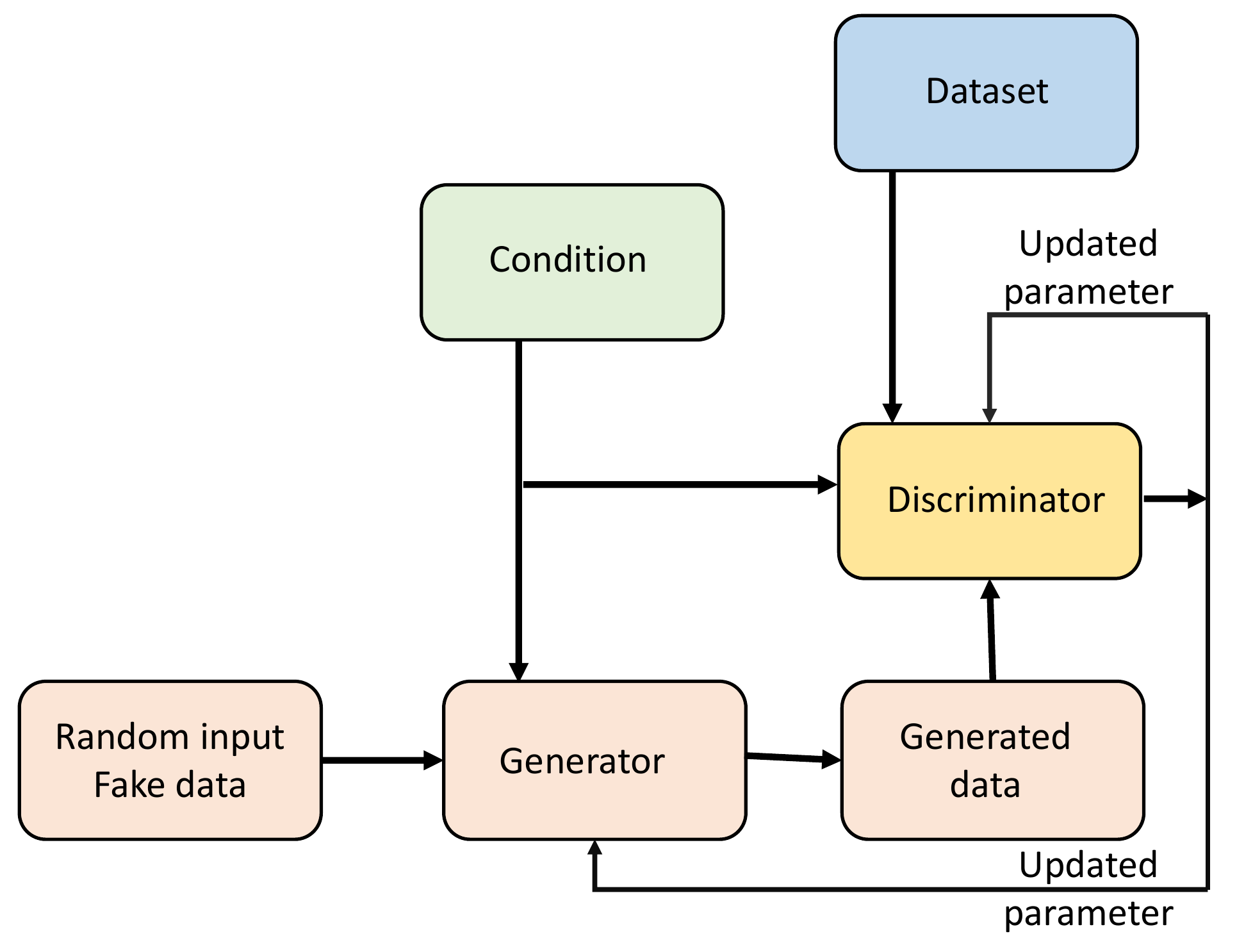}
\caption {Generative adversarial network framework shown in \cite{zhang2021distributed}, where each UAV has a condition sampler, a generator, a discriminator and a local data set.}
\label{GANstruct}
\end{figure}


The authors of~\cite{kerdegari2019smart} study a GAN-based pixel-wise image classification in UAV-assisted crop monitoring, where a generator is formulated to create real images, making a discriminator extract features and improve its learning accuracy on the pixel classification.  In~\cite{xi2020drl}, GAN is studied with a dual-stream representation learning model to identify small objects from low-resolution UAV images. In coupling with an autoencoder, a GAN can decompose a low-resolution image into low-frequency and high-frequency components. The missing information in the decomposed components can be recovered by training the GAN. Moreover, GAN-based remote sensing and image processing have also been studied extensively; see~\cite{becker2021generating,hu2019uav,wen2019single,costea2017creating,pacot2020cloud,shashank2020identifying,wang2018effective}.

\subsubsection{Generative Adversarial Network for Radio Feature Extraction}
Other than imagery features, radio propagation, e.g., received signal strength (RSS)~\cite{Li2018RecentAO}, is another important feature, which can be used to design the flight trajectory of UAVs, maintain their connectivity, and schedule radio communication resources. 
%
%
%
%
In~\cite{zhang2021distributed} and \cite{zhang2021distributed2}, the UAV trains a local GAN for mmWave channel distribution estimation according to the captured air-to-ground and air-to-air channel information. A distributed cooperative learning framework based on the GAN allows the UAV to learn the channel distribution from other agents while avoiding revealing the real measured data or the trained channel model to the other agents. 
Moreover, GAN is integrated with a long short-term memory (LSTM) to maximize the sum rate of UAV-assisted wireless communications~\cite{xu2021generative}. LSTM is an artificial RNN architecture used in deep learning. LSTM utilizes feedback connections for learning. The GAN-LSTM framework is trained at the UAVs to learn the optimal resource allocation, e.g., transmission power, spectrum allocation, communication schedule, and trajectories.

\vspace{0.2cm}
\noindent \textbf{Remark:} With their capability of processing data and efficiently extracting features, Deep Neural Networks (DNN) methods, such as CNN and MLP, have been at the forefront of enabling UAV-aided imagery applications. Adversarial networks can also aid in both imagery and radio feature extraction, and meet the demands of accuracy and image precision. However, some methods require larger computational resources. The layer of AI added to these applications can also be used to interpret the aforementioned features and model the features. 






\section{ML for Feature Interpretation and Regeneration}\label{sec: feature interpretation}

{\color{black}ML has the great potential to improve processes and aid in decisions in various application domains. The concepts of interpretability and regeneration in ML are possible through decision trees,  clustering of data, and regression models. Supervised learning strategies, such as LR, use regressive stochastic configurations to interpret the features captured by UAVs. Feature interpretation has also been used to assist navigation through semi-supervised clustering models. The regenerative capabilities of the ML algorithm help model the environment and, in turn, assist the safe cruise of the UAVs based on the probabilistic knowledge of the environment. GMM has been extensively used to model the environments and assist in flight path decisions. }


\subsection{Supervised Learning-based Feature Interpretation}


\subsubsection{Feature interpretation by Linear Regression}

There has been an increasing usage of UAVs in environmental monitoring and crop surveillance.  UAVs collect sensing information via onboard sensors such as cameras, infrared sensors, etc. 
The commonly used tool for processing the sensory data is linear regression (LR) and some variants. {\color{black}Regression analysis is a domain under supervised machine learning. This strategy aims at modeling the relationship between a certain number of features and a continuous target variable. This results in a quantitative result to define and interpret the underlying features.}
The authors of \cite{LR1} develop and validate a UAV-based air pollution measurement system. 
An LR model is adopted to estimate how the sensor position influences the measurement of pollutant concentration. Guidelines are provided on the development of a UAV system to detect the point source emissions. 
The authors of \cite{LR2} aim at developing the relationship between the crop coefficient and the normalized difference vegetation index for evapotranspiration estimation. Besides the LR model, the authors of \cite{LR2} also use a deep stochastic configuration networks model to build the relationship. 

The authors of \cite{LR3} use a UAV with a camera to calculate visible band vegetation indices and plant height to estimate biomass. The multiple LR model is used to combine the plant height information and the vegetation indices. 
In \cite{LR4}, to evaluate the health condition of wetland ecosystems, the  structure from motion (SfM) approach was adopted to map a field with overlapping photos captured by a UAV. 
The vegetation indices and SfM  cloud points can potentially describe the aquatic plants' growth conditions, which can be utilized for designing an LR model. 

Some other regression models have been utilized to extract models from the sensory data. 
The study conducted in \cite{LR5} focuses on the monitoring of water quality conditions and analyzes the near-infrared (NIR) data captured by a UAV  using a fuzzy regression model.
The authors of \cite{LR6} consider the issue of bathymetric mapping. With the collected RGB images, the authors use a geographically weighted regression (GWR) model and show that the developed GWR model successfully alleviates the biases of the multiple LR model. 
The authors of \cite{LR7} investigate on the quantitative estimation of soil salinity. A piece of electromagnetic induction equipment and a hyperspectral camera is used to collect data, and a random forest regression model is developed. 

\textit{LR is a supervised MP method that is easy to implement.} However, its major shortcoming lies in the assumption of the linearity between dependent and independent variables.  Assuming the existence of a straight-line relationship often leads to incorrect models. In addition, this method is prone to noise and over-fitting. In particular, it cannot be used in cases where the number of features is larger than that of the observations as a result of which an over-fitting model is built.


\subsection{Semi-supervised Learning for UAV-based Feature Regeneration}

\textcolor{black}{Semi-supervised learning has been extensively used in UAV-based feature extraction and prediction. These strategies have been able to provide forecasts to enable IoT applications with non-trivial QoS requirements. Feature interpretation is vital to avoid embedded biases in a learning model. Interpretations help to determine how an ML algorithm arrives at its predictions. The usage of interpretation can be extrapolated to measure the effects and trade-offs in an ML model.}

\subsubsection{Classification by K-Means Clustering}  


\begin{figure}[h!]
\centering
\includegraphics [width=3.5in]{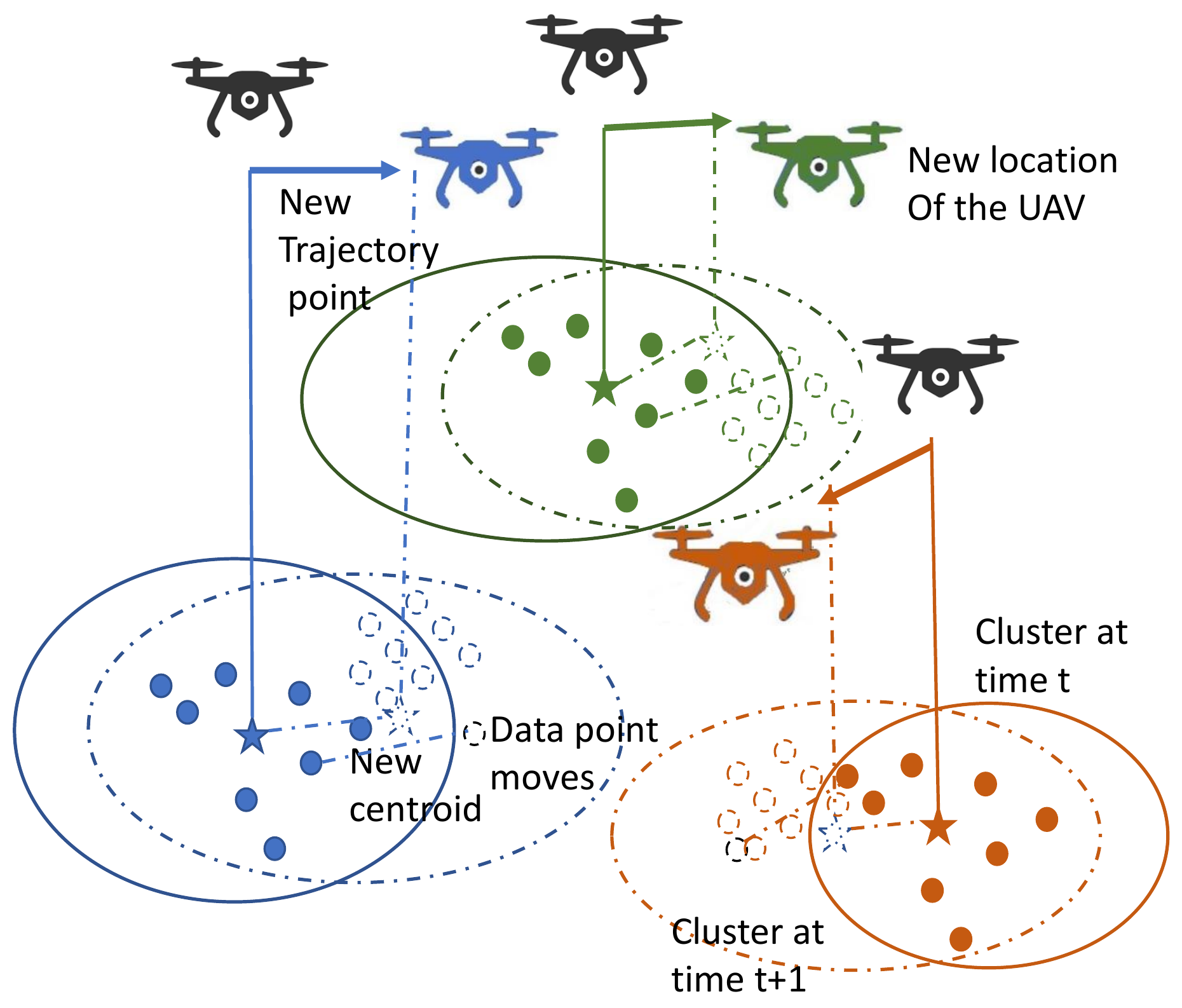}
\caption {K-means clustering-enabled multi-UAV surveillance systems, where the UAVs are driven towards the updated centroids \cite{kmeans6}.}
\label{kmeansstruct}
\end{figure}



K-means clustering has been used in planning paths for multi-UAV systems. A well-studied problem is using a fleet of UAVs to conduct multiple tasks in a particular area \cite{kmeans2, kmeans3}. With the given locations of the tasks, K-means is used to cluster the tasks into several subsets first. Within each subset, existing optimization methods, including but not limited to the simulated annealing (SA) algorithm \cite{kmeans3} and the genetic algorithm (GA)~\cite{kmeans2}, can then be applied to plan each UAV's flight route. 
K-means has also been used to navigate UAVs' movement for coverage control. The authors of \cite{kmeans4} consider using a multi-UAV system to provide cellular  services to users in an area of interest. To achieve a good enough quality of service, the problem of optimally deploying UAVs is investigated. The developed method interactively groups the users given their locations and UAVs and then drives the UAVs towards the corresponding centroids. The algorithm is proved to achieve local optimum. A similar idea has also been used for aerial surveillance by multi-UAV systems \cite{kmeans5, kmeans6}.

\subsubsection{Gaussian Mixture Model for Environment Modelling} 

  
\begin{figure}[h!]
\includegraphics [width=3.6in]{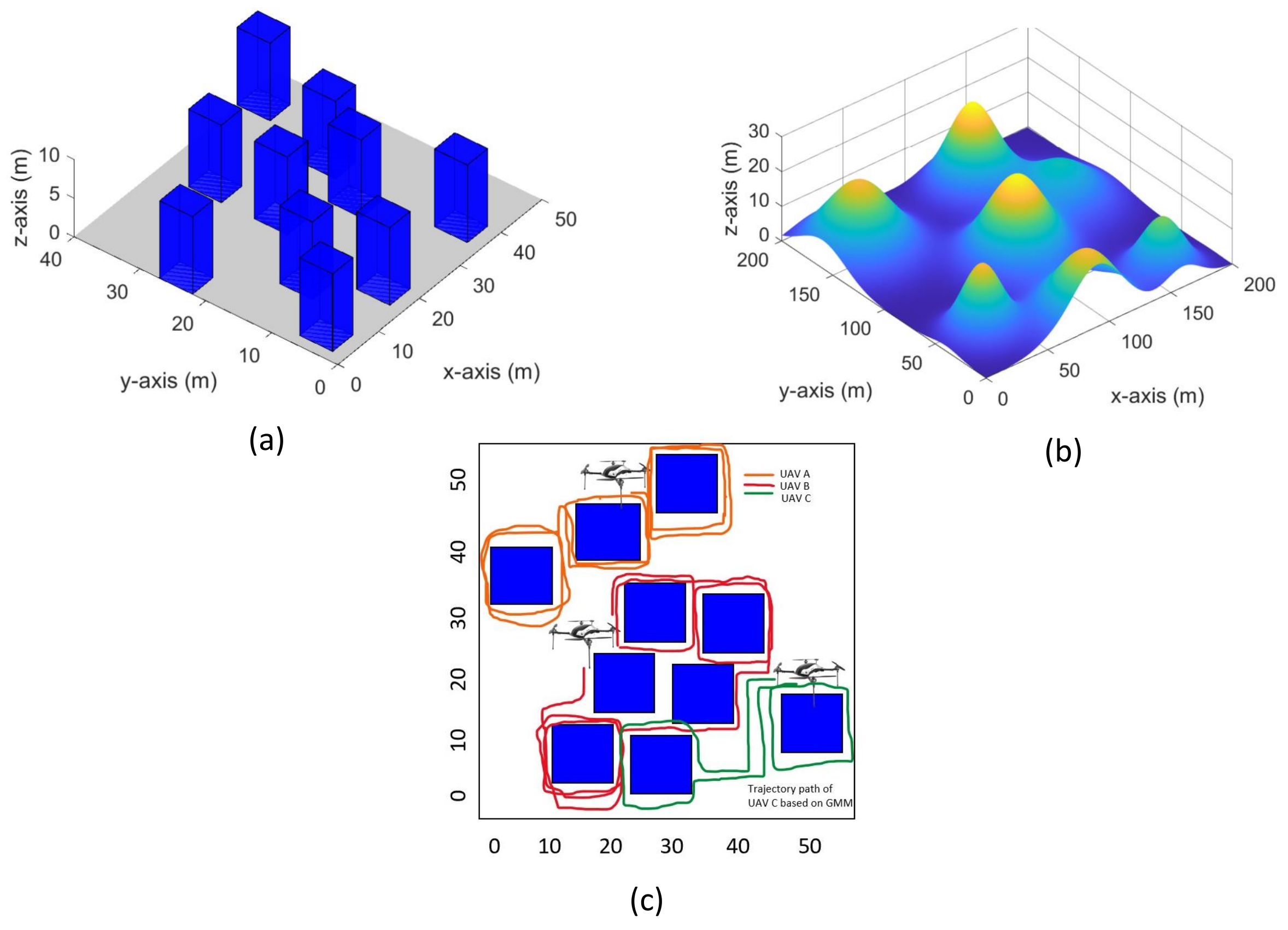}
\caption {Environment modeling and regeneration for trajectory planning using GMM: (a) the modeling of the locations of buildings across axes; (b) GMM with the approximate spatial distribution; and (c) the routes of the deployed UAVs through the considered area to maximize the detection probability based on the GMM. }
\label{gmmstruct}
\end{figure}


GMM is used to model two-dimensional complex-shaped, static obstacles and help prevent UAVs from collisions. In~\cite{Mok_GMM_2017}, given the prior probabilistic knowledge of the obstacles, a GMM is generated to construct the potential field of the area of interest. By following the standard GMM approximation steps, the EM method is used to iteratively estimate the parameters of GMM and allow the GMM to approach the known distribution of the obstacles. The potential field can be generated by taking derivatives over the GMM. The flight paths of UAVs can be obtained by following the field arrows. 
Qiao \emph{et al.}~\cite{Qiao_GMM_2015} propose a trajectory prediction model, named GMTP, which models the complex motion patterns based on GMMs and clusters the trajectory data into distinct components. As a result, the possible trajectories can be inferred by carrying out Gaussian process regression in TensorFlow probablity. 

GMM is also applied to model the heatmap of the probabilities of finding an object in an area.
A UAV is employed to execute a search mission in~\cite{Lin_GMM_2014}, where the probability of finding an anticipated object is maximized by producing an efficient flight path. Different probabilities are modeled to detect the object in different parts of the considered area, depending on the environmental parameters, e.g., foliage coverage, shadowing, and illumination conditions. GMM is employed to approximate the spatial distribution of the probabilities over the considered area by using the ``Accord.Machine Learning'' library in the ``Accord.NET'' framework \cite{lin2014hierarchical} to estimate model parameters. The GMM model provides a probabilistic mission difficulty map for the search mission and allows the different parts of the area to be prioritized hierarchically for the search. A few heuristics, namely, Top2 and TopN, are designed to hierarchically route the UAV through the considered area to maximize the detection probability. 

GMM is further integrated with the celebrated horizon control to plan the trajectories of multiple UAVs dispatched to search a complex environment~\cite{Yao_GMM_2017}. As done in~\cite{newaz2016fast}, GMM is employed to approximate the {\em a priori} known probability of finding the object. The searching area is accordingly divided and prioritized. 
The receding horizon control, also known as the model predictive control (MPC), is deployed at each of the UAVs to plan their flight paths on the fly for target search, collision avoidance, and simultaneous arrival at a destination. To maximize the predicted mission payoff, cooperation among the UAVs is needed, where the UAVs notify each other of their flight paths by regular broadcast.

Additionally, GMM can  model the spatial distribution of radio traffic to assist with the deployment of BSs, including UAV-BSs. In~\cite{Zhang_GLOBECOM_2018}, a cellular network is considered, which consists of multiple UAV-based aerial BSs, ground BSs, and a set of user terminals on the ground served by the UAVs and BSs. It is crucial to predict traffic congestion for optimal placement of the UAVs, e.g., to minimize UAVs' energy consumption on communication as well as relocation. By using a weighted expectation-maximization algorithm, a GMM is generated to model the traffic distribution. Simulations show that this method can reduce UAVs' energy consumption on  communication by 20\% and on mobility by 80\%, as compared to heuristic-based alternatives.

\vspace{0.2cm}
\noindent \textbf{Remark:} K-means clustering and linear regression with their function of clustering and regression of the data points helps in interpreting the features from the environment. Probabilistic models, such as GMM, also have been used in modeling the spatial distribution and classifying the features, and predicting them. These methods are very accurate when there is enough prior data in the environment to process. With the feature extraction and interpretation, the advancement of ML techniques opens the doors to complex applications, such as the control of a UAV.

\section{Challenges, Open Issues, and Discussion}\label{sec: open issues}

As revealed in this survey, considerable effort has been devoted to ML-based designs of the four key elements of UAV operations and communications, i.e., joint trajectory and mission planning, aerodynamic control and operation, perception and feature extraction, and feature interpretation and regeneration. However, little progress has been witnessed to jointly and holistically design an ML-based, end-to-end solution to closely integrate the four elements of efficiency, reliability, and quality assurance~\cite{doi:10.1142/S0218194020400227}. 

Such a holistic, end-to-end ML design of the four elements is important, due to the fact that UAVs are increasingly equipped with intelligence and autonomy and deployed in teams for sophisticated operations, such as safety and security surveillance~\cite{HailongTASE2021,Dilshad2020ICTC}, environmental survey, and objective detection~\cite{MITTAL2020104046}, disaster rescue~\cite{RAJAN2021129}, and animal herding \cite{sun2020quantifying}. Moreover, there is a growing demand for having UAVs work collaboratively with humans to form human-UAV teams~\cite{ZDSLZC21}. 

When designing holistically the end-to-end ML solution for sophisticated operations and collaborations, the following challenges arise.


\subsection{Support for IoT With Minimal Prior Data} With the growth of the UAV market speculated within the next decade and the increasing number of IoT applications supported by UAVs, there will be an exponential increase in air traffic. UAVs with the aid of ML algorithms must identify the authorized airspace restrictions, synchronize with other nearby aircraft paths, and plan their trajectory to ensure the safety of the UAV, other flying objects in the environment, ground pedestrians and properties~\cite{kakaletsis2021computer}. It must also aim at task completion while providing an equal priority to meet the aforementioned demands~\cite{tian2013drone}.
Since data are essential for data-driven ML algorithms, a typical issue is the lack of prior data about the environment changes or the unexpected events around the operation (e.g., other UAVs' cruise paths)~\cite{flint2002cooperative}. There is a need for some offline training data that can be used for operations and communications of UAVs.


\subsection{Increasing Energy Requirement vs. Finite Battery}

ML operations can be computationally expensive and energy-hungry. Lower levels of feature extraction and the online training of DRL modules demand significant computational resources~\cite{zhang2022mobile}.

Limited battery and onboard processing capabilities of the UAVs restrict the applications of ML-based techniques to on-board object detection, depth prediction, target tracking, and localization~\cite{abeywickrama2018empirical}. Practical constraints in accordance with the computational power and real-time parallel data processing heavily impact the design and implementation of ML solutions for UAV-aided applications.
    
There is a need for investigation and verification of energy-efficient AI/ML-aided aerial systems, especially in line with the computation efficiency and hardware design. Some recent advancements propose a combination of a variety of ML techniques to predict cooperatively the outputs and thus improve computational efficiency. There is also a need for lightweight ML techniques, e.g., R-CNN and SNN, that do not heavily impose demands on the underlying system.
This gives rise to a challenge for developing suitable embedded hardware and software, and the need for more efficient ML architectures.
    
Most commercially available UAVs are powered by onboard batteries or fuel. Due to the payload limitation, many UAVs can fly for a short time. To enable a long-distance flight, a novel idea of UAV collaborating with public transport systems is proposed \cite{HSH21ITS,BinLiuTITS2021}. The UAVs can rest on the roof of public transport vehicles, and turn off their motors for energy saving. If charging facilities are installed, the UAVs' batteries can also be recharged. This would be a solution to the long-time operation of UAVs in a smart city. The UAVs need to be embedded with advanced decision-making, planning, navigation, and control systems to conduct various actions, such as deciding which vehicle to travel with, predicting the vehicle's arrival time, etc. 

Additionally, there are inherent trade-offs between the computing demand and energy budget in UAV platforms~\cite{rao2010network}. For instance, when using feature extraction methods to achieve higher accuracy in a model, a UAV may suffer from a higher requirement of computational resources, more significant latency to reach the convergence of its ML model, faster depletion of energy, and hence a much shorter UAV mission time. 
A quick depletion of energy reserve could limit the maneuverability of the UAVs at a later stage, compromising mission quality and completion~\cite{gupta2015survey}. 
Additionally, there are also complex UAV applications with multiple conflicting objectives. ML algorithms, such as reinforcement learning, can suffer from difficulties to converge, because of many different objectives and penalties~\cite{shamsoshoara2019distributed}.

\subsection{UAV Cooperation Without Persistent Connectivity}
By employing UAV teams or swarms, the actions of the UAVs are individually trained at each UAV with the independent state observations,  e.g., to achieve fast object detection or environment mapping~\cite{spurny2019cooperative}. The action of a UAV at the current network state not only determines the next network state but also influences the actions of all other UAVs in the future. As a result, the network state observed by a UAV can be quickly outdated, since the network state has been transferred due to an action of another UAV~\cite{li2021research}. In this sense, multi-agent DRL would undergo a substantially long convergence time in multi-UAV networks, or even divergence. A potential solution can be sharing online the action and state observations among the UAVs so that joint action can be trained for all the agents. However, this requires all the UAVs to maintain consistent and reliable wireless connections, which could be challenging in practice.~\cite{SongTIFS2020}.
  
\subsection{Privacy and Security of UAV Communications} To coordinate the training of multiple collaborative UAVs’ actions, some private information, such as the network states and rewards, needs to be shared among the UAVs. Concerns arise from privacy and data security. Due to the broadcast nature of wireless channels, the UAVs’ transmissions for updating the training environment of the agents are vulnerable to eavesdropping and message modification attacks~\cite{xiao2017user}. An adversary can potentially maliciously manipulate the action training of the UAVs, which destructs the applicability of multi-agent ML to real-world UAV networks. Although distributed training can address the private information leakage issue, each UAV has to conduct supervised learning to pre-process the prior knowledge of network states in the environment~\cite{munawar2021uavs}. It is noted that this environmental sensitive information requires a considerable effort to obtain, e.g., recording the network state values of every movement of the UAVs along the trajectories in advance.
    
\subsection{Support for Heterogeneous UAV Swarms} 
Cooperative UAVs are playing increasingly important roles in precision agriculture~\cite{radoglou2020compilation} and disaster management~\cite{terzi2019swifters}. A UAV swarm can be heterogeneous, and consists of UAVs of different types, sizes, features, and functionalities with a diverse variety of processing capabilities and GPUs. These differences will have a dire impact on their maneuverability, computing capability, communication range, and response delays. 
When we take an ML algorithm, e.g., DRL, to learn and predict the environment, the actions and the environment are expected to be updated synchronously (e.g., per episode). The delays pertaining to the heterogeneous nature of the UAVs may considerably slow down the convergence~\cite{XinchenJSAC2019}. Efficient offloading techniques and methods must be developed to improve real-time synchronization despite the diverse features of heterogeneous UAVs. 
   
\subsection{Responsible ML for UAVs Interacting With Reactive Objects}
There is a growing acknowledgement  that the best results ensue when humans work collaboratively with machines (e.g., BMW reports human/robot teams were about 85\% more productive than the old assembly lines~\cite{calzavara2020ageing}). Humans and UAVs can team up and cooperate in rescue, firefight, and public safety and security. A human-UAV team must have a shared understanding of the physical world and must ensure the safety and security of its members. The team also requires its members to understand each other’s capabilities and roles and identify intent (which is consistent with the idea of human-machine shared control \cite{huang2021human}). These requirements pose significant scientific challenges (e.g., how to develop situational awareness in UAVs and enable them to act cooperatively, recognize humans' intent, and distribute decision-making processes). The science of harnessing complementary human and machine intelligence represents a significant knowledge and capability gap. 
   
Higher-level abstractions, such as UAV supervision and planning systems, have so far garnered little attention from the research community. Most ML methods designed for UAVs are for sensing tasks, such as traffic detection \cite{khan2020smart} and classification of data.  Little investigation has been conducted on the interaction between UAVs and reactive objects, such as humans and animals. Complex behaviors feature the systems involving such interactions. One example is the interaction of UAVs and animals, such as sheep for herding purposes. To achieve the goal, it is important to understand how the sheep would react to the presence of UAVs.  Another example is human-UAV teams, where human participants could react differently to the same action of UAVs under different contexts. The use of supervised learning in these systems could be inappropriate, as most datasets are collected in the absence of UAVs.
   
\subsection{Experimental Prototyping and Validation} The ML tools, such as CNN, RNN, K-means, and GMM, are often implemented and tested on PyTorch or Google TensorFlow (i.e., the two most widely used ML platforms) for feature extraction and analysis of UAV control and communications. 
    To train the ML models, a real-world testbed with multiple UAVs needs to be built to collect large amounts of data. Such a system requires the UAVs to be highly cooperative for autonomous flight and minimize human intervention. A non-trivial effort would be required to deliver a prototype of the system. 
    The DRL tools, e.g., single/multi-agent DQN~\cite{jointYousef2021} or DDPG~\cite{li2021continuous}, are often designed for trajectory planning, flight control, and mission schedule of UAV-assisted systems. Unfortunately, the use of real-world datasets and testbeds to validate DRL techniques is challenging. The reason is that DRL interacts with the environment and makes decisions that can lead to further changes in the environment. Particularly, the decision of a UAV on its flight control and communication schedule can affect the statuses of not only the scheduled ground nodes in the training environment but also all the unscheduled nodes as well~\cite{li2021lstm}. To this end, a static real-world dataset, which does not interact with the UAVs and respond to the UAV's decisions, would be inadequate to evaluate the DRL techniques. 
   
 \subsection{Theory and Implementation of RL/DRL for UAV Attitude Control} Existing autopilot systems of UAVs are based primarily on the PID control systems. This type of control system has demonstrated excellent performance in stable environments~\cite{chao2010autopilots}. In unpredictable and harsh environments, however, more sophisticated control is needed. Intelligent flight control systems are a new option to address the shortcomings of the PID control systems by incorporating RL/DRL techniques. Recent publications have demonstrated the effectiveness of RL/DRL on auto-piloting and navigation~\cite{RL_control2, RL_control3}. An emerging direction is to use RL/DRL for attitude control \cite{RL_control4}. However, the theoretical aspects of how accurate RL/DRL approaches can achieve and how well they can tolerate uncertainties are unclear. Moreover, since a reward is required in RL, some general guidelines on the design of the reward need to be holistically investigated to achieve satisfactory attitude control.
    
\subsection{Meta-learning and Transfer Learning for UAV Operations} 
Over the recent years, new RL techniques, such as meta-learning and transfer learning, have been used to enable UAV-aided applications. 
Meta-learning uses meta-data that includes the properties of the algorithm used and even the learning tasks to define the output \cite{hu2020meta}. As most of the classic ML techniques require higher operational costs and strain heavily on larger data sets, methods such as meta-learning can fill in to meet the rising demands of ML-aided UAV applications \cite{li2020uav}. 
    On the other hand, transfer learning aims at eliminating the need of processing a large chunk of data to reach a decisive output \cite{chew2020deep}. It shortens the training time by encompassing a pre-trained learning model with much less training data~\cite{wu2020transfer}. Transfer learning uses the stored knowledge obtained from solving a problem and then reuses the knowledge for a similar problem to obtain an optimal solution. This can be useful to meet the challenges, where less prior data is available in a UAV control and communication system~\cite{swinney2020unmanned}.

\section{Conclusion}\label{sec: conclusion}


The amalgamation of UAV and ML techniques 
adds a new layer of artificial intelligence to the existing UAV-aided applications, by improving communications, feature extraction, prediction, planning, control, and operations. 
This survey presented an extensive overview of critical ML techniques used in UAV operations and UAV-aided communications and IoT applications. 
We first provided an in-depth review of the existing surveys and tutorials on UAV communications. Then, we discussed the key performance indicators and ML tools used in UAV operations and communications. 
Specifically, we classified different ML techniques based on their applications to feature extraction, environment interpretation, planning and scheduling, and control and operation in UAV operations and communications.  
The survey revealed that different ML techniques dominate the applications of ML to the four key modules of UAV operations and communications, namely, feature extraction, environment modeling, planning and scheduling, and control and operations. For instance, CNN has been predominately applied to UAV image processing. DRL is increasingly demonstrating its potential for online UAV control and communication scheduling. The survey also showed that there is an increasing trend to integrate different ML modules closely to tightly couple some of the UAV control modules. For example, RNN has been utilized to enhance the feature extraction and provide enhanced inputs to DRL for fast exploration and exploitation of UAV actions. However, little to no effort has been devoted to an ML-based end-to-end solution to UAV operations and communications, from feature extraction to control and operation. 
Last but not the least, the security, reliability, and trustworthiness of ML in UAV operations and applications is a white space and deserves significant attention before the full automation of UAVs comes to fruition.


%


\bibliographystyle{IEEEtran}
\bibliography{main.bib}

%








\end{document}